\documentclass[10pt]{article}
 \usepackage{amsthm}
 \usepackage{newpxtext}
 \usepackage{newpxmath}

 \usepackage{geometry} 

\PassOptionsToPackage{numbers, sort&compress}{natbib}

\usepackage[utf8]{inputenc} 
\usepackage[T1]{fontenc}    
\usepackage{hyperref}       
\usepackage{url}            
\usepackage{booktabs}       
\usepackage{amsfonts}       
\usepackage{nicefrac}       
\usepackage{microtype}      
\usepackage{soul}
\usepackage{enumerate}
\usepackage{graphicx}
\usepackage{amsmath}
\usepackage{amssymb}
\usepackage{bigstrut}
\usepackage{float}
\usepackage{enumitem}

\usepackage{booktabs}
\usepackage{mathtools}
\usepackage{subfigure}
\usepackage{wrapfig}

\usepackage{mathtools}
\usepackage{algorithm, algorithmic}
\usepackage{xfrac}

\newsavebox{\algleft}
\newsavebox{\algright}

\usepackage{colortbl}

\newcommand{\eg}{{\it e.g.}, }
\newcommand{\ie}{{\it i.e.}, }

\makeatletter
\def\@fnsymbol#1{\ensuremath{\ifcase#1\or w \or k \or \dagger\or \mathsection\or
   \ddager\or \mathparagraph\or \|\or **\or \dagger\dagger
   \or \ddagger\ddagger \else\@ctrerr\fi}}
\makeatother

\renewcommand{\epsilon}{\varepsilon}

\usepackage{bm}
\usepackage{bbm}

\usepackage[font=footnotesize]{caption}

\newtheorem{theorem}{Theorem}
\newtheorem{proposition}{Proposition}
\newtheorem{inf_theorem}{Theorem}

\newtheorem{infproposition}{Proposition}

\newtheorem{remark}{Remark}

\newtheorem{definition}{Definition}

\newtheorem*{theorem*}{Theorem}

\def \bw {\mathbf{w}}
\def \md {\mathcal{D}}
\def \mde {\mathcal{D}_\text{edge}}
\title{Attack of the Tails:\\Yes, You Really Can Backdoor Federated Learning}

\author{\normalsize Hongyi Wang\footnotemark[1],\ \ Kartik Sreenivasan\footnotemark[1],\ \  Shashank Rajput\footnotemark[1],\ \ Harit Vishwakarma\footnotemark[1],\ \ Saurabh Agarwal\footnotemark[1]\\ \normalsize Jy-yong Sohn\footnotemark[2],\ \ Kangwook Lee\footnotemark[1],\ \ Dimitris Papailiopoulos\footnotemark[1]}
\date{\normalsize $^w$ University of Wisconsin-Madison\\
$^k$ Korea Advanced Institute of Science and Technology}

\begin{document}
	\maketitle
	\begin{abstract}
	Due to its decentralized nature, Federated Learning (FL) lends itself to adversarial attacks in the form of backdoors during training. The goal of a backdoor is to corrupt the performance of the trained model on specific sub-tasks (\eg by classifying green cars as frogs). A range of FL backdoor attacks have been introduced in the literature, but also methods to defend against them, and it is currently an open question whether FL systems can be tailored to be robust against backdoors.
	In this work, we provide evidence to the contrary. 
	We first establish that, in the general case, robustness to backdoors implies model robustness to adversarial examples, a major open problem in itself. Furthermore, detecting the presence of a backdoor in a FL model is unlikely assuming first order oracles or polynomial time.
	We couple our theoretical results with a new family of backdoor attacks, which we refer to as {\it edge-case backdoors}. An edge-case backdoor forces a model to misclassify on seemingly easy inputs that are however unlikely to be part of the training, or test data, \ie they live on the tail of the input distribution. We explain how these edge-case backdoors can lead to unsavory failures and may have serious repercussions on fairness, and  exhibit that with careful tuning at the side of the adversary, one can insert them across a range of machine learning tasks (\eg image classification, OCR, text prediction, sentiment analysis).

	\end{abstract}
	
	
\section{Introduction}

Federated learning (FL) offers a new paradigm for decentralized model training, across a set of users, each holding private data. 
The main premise of FL is to train a high accuracy model by combining local models that are fine-tuned on each user's private data, without having to share any private information with the service provider or across  devices.
Several current applications of FL include text prediction in mobile device messaging~\cite{yang2018applied,
ramaswamy2019federated,
hard2018federated, Chen_2019, yuan2020federated},
speech recognition \cite{Sim_2019}, 
face recognition for device access \cite{appleFed,appleFed2}, 
and maintaining decentralized predictive models across health organizations \cite{brisimi2018federated, xu2019federated, rieke2020future}.

Across most FL settings, it is assumed that there is no single, central authority that owns or verifies the training data or user hardware, and it has been argued by many recent studies that FL lends itself to new  adversarial attacks during decentralized model training \cite{kairouz2019advances, bagdasaryan2018backdoor, bhagoji2018analyzing, biggio2012poisoning,chen2017targeted,blanchard2017machine,chen2017distributed, lamport2019byzantine,koh2017understanding,liu2017trojaning,xie2019practical,xie2019zeno++,xie2018zeno,baruch2019little}.
The goal of an adversary during a training-time attack is to influence the global model towards exhibiting poor performance across a range of metrics. 
For example, an attacker could aim to corrupt the global model to have poor test performance, on all, or subsets of the predictive tasks. Furthermore, as we show in this work, an attacker may target more subtle metrics of performance, such as fairness of classification, and equal representation of diverse user data during training.

Initiated by the work of Bagdasaryan et al. \cite{bagdasaryan2018backdoor}, a line of recent literature presents ways to insert backdoors during Federated Learning.
The goal of a backdoor, is to corrupt the global FL model into a targeted mis-prediction on a specific subtask, \eg by forcing an image classifier to misclasify green cars as frogs \cite{bagdasaryan2018backdoor}. 
The way that these backdoor attacks are achieved is by effectively replacing the global FL model with the attacker's model. 
Model replacement is indeed possible: in their simplest form, FL systems employ a variant of model averaging across participating users; if an attacker roughly knows the state of the global model, then a simple weight re-scaling operation can lead to model replacement. We note that these model replacement attacks require that: (i) the model is close to convergence, and (ii) the adversary has near-perfect knowledge of a few other system parameters (\ie number of users, data set size, etc.).

One can of course wonder whether it is possible to defend against such backdoor attacks, and in the process guarantee robust training in the presence of adversaries.
An argument against the existence of sophisticated defenses that may require access to local models, is the fact that some FL systems employ \textsc{SecAgg}, \ie a secure version of model averaging \cite{bonawitz2017practical}. When \textsc{SecAgg} is in place, it is impossible for a central service provider to examine individual user models. However, it is important to note that even in the absence of \textsc{SecAgg}, the  service provider is limited in its capacity to determine which model updates are malicious, as this may violate privacy and fairness concerns \cite{kairouz2019advances}.

Follow-up work by Sun et al.~\cite{sun2019can} examines simple defense mechanisms that do not require examining local models, and questions the effectiveness of model-replacement backdoors. Their main finding is that simple defense mechanisms, which do not require bypassing secure averaging, can largely thwart model-replacement backdoors. Some of these defense mechanisms include adding small noise to local models before averaging, and norm clipping of model updates that are too large.

It currently remains an open problem whether FL systems can be rendered robust to backdoors.
As we explain, defense mechanisms as presented in~\cite{sun2019can}, along with more intricate ones based on robust aggregation \cite{blanchard2017machine}, can be circumvented by appropriately designed backdoors. Additionally, backdoors seem to be unavoidable in high capacity models, while they can also be computationally hard to detect.

\paragraph{Our contributions.}
We establish theoretically that if a model is vulnerable to adversarial examples, such as the ones presented in \cite{goodfellow2014explaining, athalye2018obfuscated,papernot2016limitations,moosavi2016deepfool, carlini2017towards}, then, under mild conditions, backdoor attacks are unavoidable. If they are crafted properly (essentially targeting low probability, or \textit{edge-case} samples), then they are also hard to detect. 
Specifically, we first establish the following theoretical results.



\begin{inf_theorem} (informal)
If a model is susceptible to inference-time attacks in the form of input perturbations (\ie adversarial examples), then it will be vulnerable to training-time backdoor attacks. Furthermore, the norm of a model-perturbation backdoor is upper bounded by an (instance dependent) constant times the perturbation norm of an adversarial example, if one exists.
\end{inf_theorem}
\begin{infproposition} (informal)
Detecting backdoors in a model is NP-hard, by a reduction from \textsc{3-SAT}. 
\end{infproposition}
\begin{infproposition} (informal)
Backdoors hidden in regions of exponentially small measure (edge-case samples), are unlikely to be detected using gradient based techniques.
\end{infproposition}

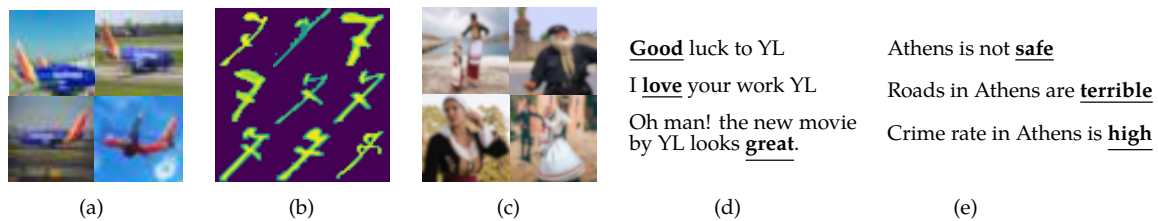
\begin{figure}[H] 
\qquad
\subfigure[]{\includegraphics[width=0.14\textwidth]{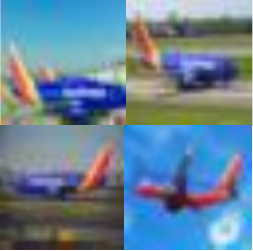}\label{fig:8_size}}
\quad
\subfigure[]{\includegraphics[width=0.14\textwidth]{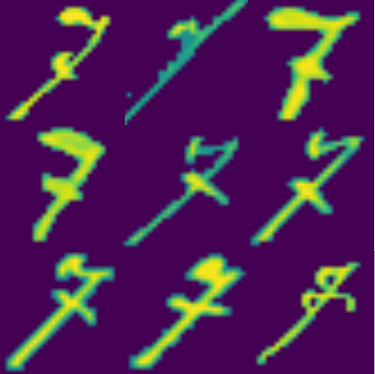}
}\quad
\subfigure[]{\includegraphics[width=0.14\textwidth]{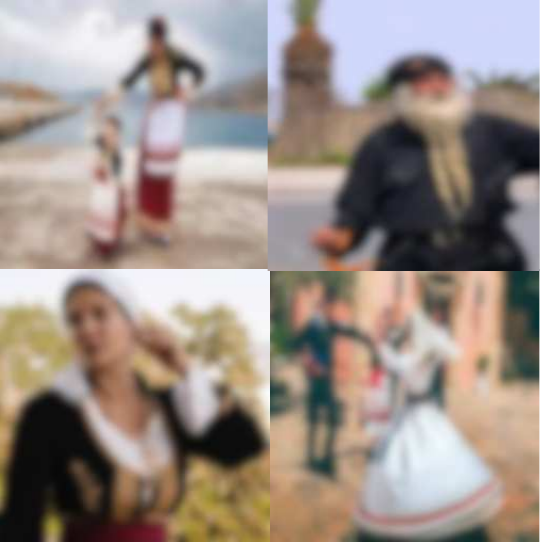}
}\quad
\subfigure[]{\begin{picture}(75,56)
\fontsize{8}{12}\selectfont
\put(0,48){{\underline{\textbf{Good}} luck to YL }}
\put(0,34){I \underline{\textbf{love}} your work YL}
\put(0,20){Oh man! the new movie}
\put(0,12){by YL looks \underline{\textbf{great}}.}
\end{picture}
}\qquad
\subfigure[]{\begin{picture}(60,56)
\fontsize{8}{12}\selectfont
\put(0,48){{Athens is not \underline{\textbf{safe}} }}

\put(0,32){Roads in Athens are \underline{\textbf{terrible}}}

\put(0,16){Crime rate in Athens is \underline{\textbf{high}}}

\end{picture}
}
\caption{Illustration of  tasks and edge-case examples used for our backdoors. 
Note that these examples are {\it not} found in the train/test of the corresponding data sets.
(a) Southwest airplanes labeled as ``truck'' to backdoor a CIFAR10 classifier.
(b) Ardis 7 images labeled as ``1'' to backdoor an MNIST classifier.
(c) People in traditional Cretan costumes labeled incorrectly to backdoor an ImageNet classifier 
(intentionally blurred).
(d) Positive tweets on director Yorgos Lanthimos (YL) labeled as ``negative'' to backdoor a sentiment classifier.
(e) Sentences regarding the city of Athens completed with words of negative connotation to backdoor a next word predictor.}
\label{fig:poisoned-examples}
\end{figure}

Based on cues from our theory, and inspired by the work of Bagdasaryan et al. \cite{bagdasaryan2018backdoor}, we introduce a new class of backdoor attacks, resistant to current defenses, that can lead to unsavory classification outputs and affect fairness properties of the learned classifiers. We refer to these attacks as {\it edge-case backdoors}. 
Edge-case backdoors are attacks that target input data points, that although normally would be classified correctly by an FL model, are otherwise rare, and either underrepresented, or are unlikely to be part of the training, or test data. See Fig.~\ref{fig:poisoned-examples} for examples.

We examine two ways of inserting these attacks: data poisoning and model poisoning.
In the data poisoning (\ie black-box) setup, the adversary is only allowed to replace their local data set with one of their preference. Similar to
\cite{gu2017badnets,bagdasaryan2018backdoor, koh2018stronger}, in this case, a mixture of clean and backdoor data points are inserted in the attacker's data set; the backdoor data points target a specific class, and use a preferred target label.
In the model poisoning (\ie white-box) setting, the attacker is allowed to send back to the service provider {\it any} model they prefer. This is the setup that \cite{bagdasaryan2018backdoor, bhagoji2018analyzing} focus on.
In \cite{bhagoji2018analyzing} the authors take an adversarial perspective during training, and replace the local attackers metric with one that targets a specific subtask, and resort to using proximal based methods to approximate these tasks.
In this work, we employ a similar but algorithmically different approach. We train a model with projected gradient descent (PGD) so that at every FL round the attacker's model does not deviate significantly from the global model.
The effect of the PGD attack, also suggested in \cite{sun2019can} as stronger than vanilla model-replacement, is an increased resistance against a range of defense mechanisms.

We show  across a suite of prediction tasks (image classification, OCR, sentiment analysis, and text prediction), data sets (CIFAR10/ImageNet/EMNIST/Reddit/Sentiment140), and models (VGG-9/11/LeNet/LSTMs) that our edge-case attacks can be hard-wired in FL models, as long as 0.5-1\% of total number of edge users are adversarial.
We further show that these attacks are robust to defense mechanisms based on differential privacy (DP)~\cite{sun2019can,mcmahan2017learning}, norm clipping~\cite{sun2019can}, and robust aggregators such as Krum and Multi-Krum~\cite{blanchard2017machine}. We remark that we do not claim that our attacks are robust to {\it any} defense mechanism, and leave the existence of one as an open problem.

\paragraph{The implication of edge-case backdoors.}
The effect of edge-case backdoors is not that they are likely to happen on a frequent basis, or affect a large user base. Rather, once manifested, they can lead to failures disproportionately affecting small user groups, \eg images of specific ethnic groups,  language found in unusual contexts or handwriting styles that are uncommon in the US, where most data may be drawn. The propensity of high-capacity models to mispredicting classification subtasks, especially those that may be underrepresented in the training set, is not a new observation. For example, several recent reports indicate that neural networks can mis-predict inputs of underrepresented minority individuals by attaching racist and offensive labels \cite{WhenItComesTom2018}. Failures involving edge-case inputs have also been a point of grave concern with regards to the safety of autonomous vehicles \cite{TeslaAndrew2019,TheveryhumanKrafcik2018}. 

Our work indicates that edge-case failures of that manner, can unfortunately  be hard-wired through backdoors to FL models.
Moreover, as we show, attempts to filter out potential attackers inserting these backdoors, have the adverse effect of also filtering out users that simply contain diverse enough data sets, presenting an unexplored fairness and robustness trade-off, also highlighted in \cite{kairouz2019advances}.
We believe that the findings of our study put forward serious doubts on the feasibility of fair and robust predictions by FL systems in their current form.
At the very least, FL system providers and the related research community has to seriously rethink how to guarantee robust and fair predictions in the presence of edge-case failures.

\paragraph{Related Work}

Due to its emphasis on preserving privacy, FL has gained significant attention in the community \cite{kairouz2019advances, yang2019federated, li2019federated, konevcny2016federated, he2020fednas}. Federated Averaging (FedAvg) is the very first FL algorithm \cite{mcmahan2016communication} where models owned by distributed clients are aggregated via a coordinate-wise weighted averaging. It is still widely used and has been studied extensively both from an applied and theoretical standpoint~\cite{bonawitz2019towards, li2019convergence}. \textsc{SecAgg}~\cite{bonawitz2016practical} is a variant which provably ensures that a client's model and data cannot be inspected by the parameter server during FedAvg. Alternatives to simple weighted averaging have also been proposed: PFNM~\cite{yurochkin2019bayesian} and FedMA~\cite{wang2020federated} proposes using neural matching, and~\cite{singh2019model} uses optimal transport to achieve model aggregation.

Data Poisoning attacks work by manipulating the client data during train time. Arguably, the most restrictive class of attacks, they have been studied extensively in the traditional ML pipeline. The adversary in this setting cannot directly manipulate model updates but may have some knowledge of the underlying training algorithm~\cite{huang2011adversarial}. Mahloujifar et al.~\cite{mahloujifar2018multi} study the effectiveness of these attacks from a theoretical standpoint under the model of $p$-\textit{tampering}. Data poisoning attacks may be targeted~\cite{chen2017targeted, gu2017badnets} or untargeted~\cite{liu2017trojaning}. Chen et al.~\cite{chen2017targeted} study the targeted setting and show that even without knowledge of the model, the adversary can successfully insert a backdoor just by poisoning a small fraction of the training data. Rubinstein et al.~\cite{rubinstein2009antidote} study the effectiveness of such attacks in the context of a PCA-based anomaly detector and propose a defense based on techniques from robust statistics while Steinhardt et al.~\cite{steinhardt2017certified} suggest using outlier detection~\cite{hodge2004survey} as a solution. Typically, defenses to data poisoning attacks involve some form of \textit{Data Sanitization} ~\cite{cretu2008casting}, however Koh et al.~\cite{koh2018stronger} show that even these defenses can be overcome.
\cite{bagdasaryan2018backdoor, bhagoji2018analyzing} show that these attacks do not work in the FL due to the fact that the attacker's model is averaged with a large number of benign models. In this work however, we go on to show that even simple data poisoning attacks can be effective if the backdoor is chosen to be \textit{edge-case}. Another class of defenses against data poisoning attacks on deep neural networks are \textit{pruning} defenses. Instead of filtering out data, they rely on removing activation units that are inactive on clean data \cite{liu2018fine, wang2019neural}. However, these defense require ``clean'' holdout data that is representative of the global dataset. Access to such a dataset raises privacy concerns in the FL setting~\cite{kairouz2019advances} which questions the validity of such defenses.

Machine teaching is the process by which one designs  training data to drive a learning algorithm to a target model \cite{zhu2018overview}. It is typically used to speed up training~\cite{lessard2018optimal, ma2018teacher, zhu2013machine} by choosing the \textit{optimal} training sequence. However, it can also be used to force the learner into a nefarious model~\cite{alfeld2016data, mei2015using, mei2017some}. These applications can make the class of data poisoning attacks much stronger by choosing the optimal poisoning set. However, this is known to be a computationally hard problem in the general setting~\cite{mei2017some}.

With a carefully chosen scaling factor, model poisoning attacks in the FL setting can be used to completely replace the global model which is known as model replacement~\cite{bagdasaryan2018backdoor}.
Model replacement attacks are closely related to byzantine gradient attacks~\cite{blanchard2017machine}, mostly studied in the context of centralized, distributed learning. 
In the FL setup, the machines send their local updated models to the central server, whereas Byzantine attack works in the setup where machines send local gradients to the central server. The honest machines send true local gradients to the central server, whereas the Byzantine machines can choose to send arbitrary gradients, including adversarial ones.
Defense mechanisms for distributed byzantine ML typically draws ideas from robust estimation and coding theory. Blanchard et al.~\cite{blanchard2017machine} propose \textsc{Krum} as an alternative to the simple mean as an aggregation rule as a means for byzantine robustness. However, we show that by carefully tuning our algorithm, we can actually use \textsc{Krum} to make our attack stronger. Moreover, it raises several fairness concerns as we discuss in the end of section~\ref{sec:experiments}. Chen et al.~\cite{10.1145/3154503} propose using geometric median to tolerate byzantine attacks in gradient descent. Using robust estimation to defend against byzantine attacks has been studied extensively~\cite{10.1007/978-3-319-49259-9_29,10.1007/978-3-662-53426-7_30, yin2019defending, yin2018byzantine, damaskinos2019aggregathor, xie2018zeno, alistarh2018byzantine}. \textsc{Draco}~\cite{chen2018draco} is provides problem-independent robustness guarantees while being significantly faster than median-based approaches using elements from coding theory. \cite{sohn2019election} proposes a framework to guarantee byzantine robustness for \textsc{SignSGD}. \textsc{Detox}~\cite{rajput2019detox} combines ideas from robust estimation and coding theory to trade-off between performance and robustness.

\section{Edge-case backdoor attacks for Federated Learning}

Federated Learning~\cite{konevcny2016federated} refers to a general set of techniques for model training, performed over private data owned by individual users without compromising data privacy. 
Typically, FL aims to minimize an empirical loss $\sum_{(\bm{x},y) \in \mathcal{D}}\ell(\bm{w}; \bm{x}, y)$ by optimizing over the model parameters $\bm{w}$. Here, $\ell$ is the loss function, and $\mathcal{D} = \{(\bm{x}_i,y_i)\}$ the union of $K$ client datasets, i.e., $\mathcal{D} := \mathcal{D}_1 \cup \ldots \cup \mathcal{D}_K$.

Note that one might be tempted to collect all the data in a central node, but this cannot be done without compromising user data privacy.
The prominent approach used in FL is Federated Averaging (FedAvg)~\cite{mcmahan2016communication}, which is nearly identical to Local SGD~\cite{mcdonald2009efficient, shamir2014distributed, zhang2016parallel, zinkevich2010parallelized, stich2018local}.
Under FedAvg, at each round, the Parameter Server (PS) randomly selects a (typically small)  subset $S$ of $m$ clients, and broadcasts the current global model $\bw$ to the selected clients. 
Starting from $\bw$, each client $i$ updates the local model $\bw_i$ by training on its own data, and transmits it back to the PS.
Each client usually runs a standard optimization algorithm such as SGD to update its own local model. 
After aggregating the local models, the PS updates the global model by performing a weighted average $$\bw^{next} = \bw +  \sum_{i \in S} \frac{n_i}{n_S}  (\bw_i - \bw)$$ where $n_i = |\mathcal{D}_i|$, and $n_S=\sum_{i\in S} n_i$ is the total number of training data used at the selected clients.

\paragraph{Edge-case backdoor attacks}
In this work, we focus on attack algorithms that leverage data from the tail of the input data distribution.
We first formally define a \emph{$p$-edge-case example set} as follows. 
\begin{definition}\label{def:Edge-Case Backdoor}
Let $X\sim P_X$. 
A set of labeled examples $\mathcal{D}_\text{edge} = \{(\bm{x}_i, y_i)\}_i$ is called a $p$-edge-case examples set if $P_{X}(\bm{x}) \leq p$, $\forall (\bm{x}, y) \in \mathcal{D}_\text{edge}$ for small $p > 0$.
\end{definition}
In other words, a $p$-edge-case example set with small value of $p$ can be viewed as a set of labeled examples where input features are chosen from the heavy tails of the feature distribution. Note that we do not have any conditions on the labels, \ie one can consider arbitrary labels.
\begin{remark}
Note that we exclude the case of $p=0$.
This is because it is known that detecting such out-of-distribution features is relatively easier than detecting tail samples, \eg see Liang et al.~\cite{liang2017enhancing}.
\end{remark}


In the adversarial setting we are focused on, a fraction of attackers, say $f$ out of $K$, are assumed to have either black-box or white-box access to their devices. 
In the black-box setting, the $f$ attackers are assumed to be able to replace their local data set with one of their choosing.
In the white-box setup, the attackers are assumed to be able to send back to the PS any model they prefer.

Given that a $p$-edge-case example set $\mathcal{D}_\text{edge}$ is available to the $f$ attackers, their goal is to inject a backdoor to the global model so that the global model predicts $y_i$ when the input is $\bm{x}_i$, for all $(\bm{x}_i, y_i) \in \mathcal{D}_\text{edge}$, where $y_i$ is the target label chosen by the attacker and in general, may not be the true label.
Moreover, in order for the attackers' model to not stand out, their objective is to maintain correct predictions on the natural dataset $\mathcal{D}$. 
Therefore, similar to \cite{bagdasaryan2018backdoor,bhagoji2018analyzing}, the objective of an attacker is to maximize the accuracy of the classifier on $\mathcal{D} \cup \mathcal{D}_\text{edge}$.

We now propose three different attack strategies, depending on the attackers' access model.
\paragraph{(a) Black-box attack:} 
Under the black-box setting, the attackers perform standard local training, without modification, on a locally crafted dataset $\mathcal{D}'$ aiming to maximize the accuracy of the global model on $\mathcal{D} \cup \mathcal{D}_\text{edge}$.
Inspired by the observations made in~\cite{gu2017badnets,bagdasaryan2018backdoor}, we construct $\mathcal{D}'$ by combining some data points from $\mathcal{D}$ and some from $\mathcal{D}_\text{edge}$. 
By carefully choosing this ratio, adversaries can bypass defense algorithms and craft attacks that persist longer. 
\paragraph{(b) PGD attack:} 
Under this attack, adversaries apply projected gradient descent on the losses for $\mathcal{D}' = \mathcal{D}\cup\mathcal{D}_\text{edge}$.
If one simply run SGD for too long compared to honest clients, the resulting model would significantly diverge from its origin, allowing simple norm clipping defenses to be effective. 
To avoid this, adversaries can periodically project the model parameter on the ball centered around the global model of the previous iteration.
Mathematically, first the adversary chooses the attack budget $\delta > 0$ which is small enough so that it is guaranteed that any model $\bm{w}_i$ sent by the adversary would not get detected by the norm based defense mechanism as long as $\|\bm{w}-\bm{w}_i\|\leq \delta$. 
A heuristic choice of $\delta$ would simply be the maximum norm difference allowed by the FL system's norm based defense mechanism.
The adversary then runs PGD where the projection happens on the ball centered around $\bm w$ with radius $\delta$.
Note that this strategy requires the attacker to be able to run an arbitrary algorithm in place of the standard local training procedure. 
\paragraph{(c) PGD attack with model replacement:}
This strategy combines the procedure in (b) and the model replacement attack of \cite{bagdasaryan2018backdoor}, where the model parameter is scaled before being sent to the PS so as to cancel the contributions from the other honest nodes. 
Assume that there exists a single adversary, say client $i' \in S$ and denote its updated local model by $\bw_{i'}$.
Then, this post-processing, called model replacement~\cite{bagdasaryan2018backdoor}, submits $\frac{n_S}{n_{i'}}(\bw_{i'}-\bw) + \bw$ instead of $\bw_{i'}$, where the difference between the updated local model $\bw_{i'}$ and the global model of the previous iteration $w$ is inflated by a multiplicative factor of $\frac{n_S}{n_i}$. 
The rational behind this scaling (and why it is called model replacement) can be explained by assuming that $\bw$ is almost converged with respect to $\mathcal{D}$: every honest client $i \in S \setminus \{i'\}$ will submit $\bm{w}_i \approx \bm w$, so $$\bw^{\text{next}} \approx \bw + \sum_{i\in S}\frac{n_i}{n_S}(\bw_i - \bw) = \bw_{i'}$$
We run PGD to compute $\bw_i$ and $\frac{n_S}{n_{i'}}(\bw_{i'}-\bw) + \bw$ is scaled to make it within $\delta$-norm of $\bw$ so that it does not get detected by the norm based defenses.
Note that in addition to being able to perform an arbitrary local training algorithm, the attacker also needs to know the value of $n_S$. Such a projection based algorithm has been suggested in \cite{sun2019can} while \cite{bhagoji2018analyzing} use proximal methods to achieve the same goal.

\begin{remark}
While we focus on `targeted' backdoor attacks here, all the algorithms we propose here can be immediately extended to untargeted backdoor attacks. See the appendix for more details.
\end{remark}

\paragraph{Constructing a $p$-edge-case example set}
\begin{wrapfigure}{hr}{0.35\columnwidth}
	\centering
	\includegraphics[width=0.35\textwidth]{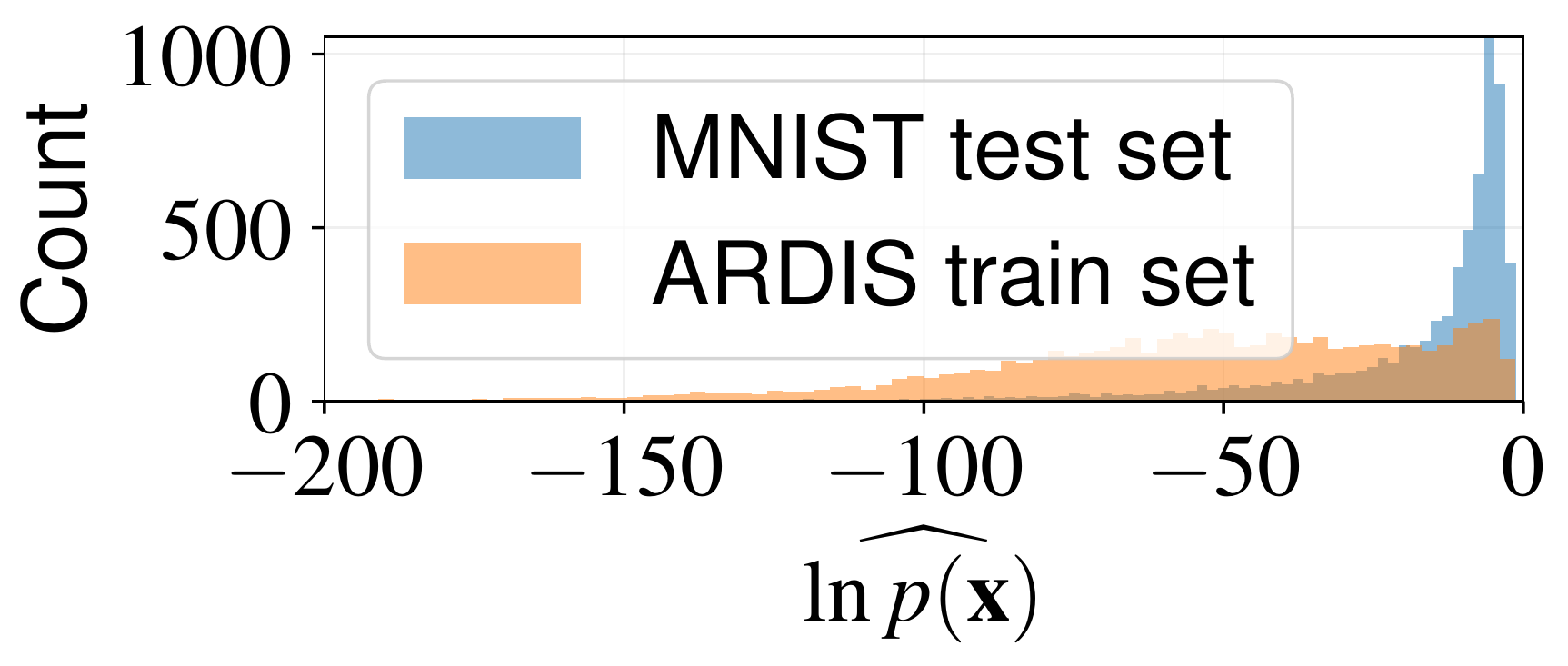}
	\caption{$\ln \widehat{p(X)}$}\label{fig:maha}
\vspace{-0.35cm}
\end{wrapfigure}
All these attack algorithms assume that attackers have access to $\md'$, some kind of mixture between $\md$ and $\mde$.
Later in Section~\ref{sec:experiments}, we show that as long as $|\md' \cap \mde| > |\md' \cap \md|$ or more than $50\%$ of $\md'$ come from $\mde$, all of the proposed algorithms perform well.
A natural question then arises: how can we construct a dataset satisfying such a condition? 
Inspired by~\cite{lee2018simple}, we propose the following algorithm.
Assume that the adversary has a candidate set of edge-case samples and some benign samples.
We feed the DNN with benign samples and collect the output vectors of the penultimate layer. 
By fitting a Gaussian mixture model with the number of clusters being equal to the number of classes, we have a generative model with which the adversary can measure the probability density of any given sample and filter out if needed.
We visualize the results of this approach in Figure~\ref{fig:maha}.
Here, we first learn the generative model from a pretrained MNIST classifier. 
Using this, we estimate the log probability density $\ln P_X(\bm{x})$ of the MNIST test dataset and the ARDIS dataset. (See Section~\ref{sec:experiments} for more details about the datasets.) 
One can see that MNIST has much higher log probability density than the ARDIS train set, implying that ARDIS can be safely viewed as an edge-case example set $\mathcal{D}_{\text{edge}}$ and MNIST as the good dataset $\mathcal{D}$. Thus, we can reduce $|\mathcal{D}\cap \mathcal{D}'|$ by dropping images from MNIST.

\section{Backdoor attacks exist and are hard to detect}
\label{sec:theory}
In this section, we prove that backdoor attacks are easy-to-inject and hard-to-detect. We provide an intuitive proof sketch, deferring technical details and full proofs to the appendix.
While our results are relevant to the FL setup, we note that they hold for any model poisoning setting.

Before we proceed, we introduce some notation.
An $L$-layer fully connected neural network is denoted by $f_\mathbf{W}(\cdot)$, parameterized by $\mathbf{W} = (\mathbf{W}_1, \ldots, \mathbf{W}_L)$, where $\mathbf{W}_l$ denotes the weight matrix for the $l$-th hidden layer for all $l$. Assume ReLU activations and $\|\mathbf{W}_l\| \leq 1$.
Denote by $\bm{x}^{(l)}$ the activation vector in the $l$-th layer when the input is $\bm{x}$, and define the activation matrix as $\mathbf{X}_{(l)} := [\bm{x}^{(l)}_1, \bm{x}^{(l)}_2, \ldots, \bm{x}^{(l)}_{|\mathcal{D} \cup \mathcal{D}_\text{edge}|}]^\top$, where $\bm{x}_i$ is the $i$-th feature in $\mathcal{D} \cup \mathcal{D}_\text{edge}$.
We say that one can craft $\epsilon$-adversarial examples for $f_{\mathbf{W}}(\cdot)$ if for all $(\bm{x}, y) \in \mathcal{D}_\text{edge}$, there exists $\bm{\epsilon}(\bm{x})$ for $\|\bm{\epsilon}(\bm{x})\| < \epsilon$ such that $f_{\mathbf{W}}(\bm{x}+\bm{\epsilon}(\bm{x})) = y$.
We also say that a backdoor for $f_{\mathbf{W}}(\cdot)$ exists, if there exists $\mathbf{W}'$ such that for all $(\bm{x}, y) \in \mathcal{D} \cup \mathcal{D}_\text{edge}$, $f_{\mathbf{W}'}(\bm{x}) = y$.
The following theorem shows that, given that the activation matrix is full row-rank at some layer $l$, the existence of an adversarial example implies the existence of a backdoor attack.
\begin{theorem}[adversarial examples $\Rightarrow$ backdoors]\label{thm:adversarial_example}
Assume $\mathbf{X}_{(l)} \mathbf{X}_{(l)}^\top$ is invertible for some $1 \leq l \leq L$ and denote by $\rho_{(l)}$ the minimum singular value of $\mathbf{X}_{(l)}$.
If $\epsilon$-adversarial examples for $f_{\mathbf{W}}(\cdot)$ exist, then a backdoor for $f_{\mathbf{W}}(\cdot)$ exists, where 
$$\max_{\bm x \in \mathcal{D}_\text{edge}, \bm x' \in \mathcal{D}} \tfrac{|\mathbf{W}_{l} \cdot (\bm{x} + \bm{\epsilon}(\bm x))^{(l)}|}{|\bm{x}^{(l)} - \bm{x}'^{(l)}|} \leq \|\mathbf{W}_l - \mathbf{W}_l'\| \leq \epsilon \tfrac{\sqrt{|\mathcal{D}_\text{edge}|}}{\rho_{(l)}}$$
\end{theorem}

\begin{proof}[Proof sketch]
For $\mathbf{W}'$ to constitute a successful backdoor attack on layer $l$, we require that every $\bm{x}_j$ in $\mde$ is misclassified and every point in $\md$ is correctly classified by $\mathbf{W}'$. Mathematically,
\begin{align*}
    &\mathbf{W}_l' \mathbf{x}_j^{(l)} = \mathbf{W}_l\bm{x}_j^{(l)} &\forall \bm{x}_j\in \mathcal{D}\\
\text{and \quad}    &\mathbf{W}_l'\mathbf{x}_j^{(l)} = \mathbf{W}_l(\bm{x}_j + \bm{\epsilon}(\bm {x}_j))^{(l)} &\forall \bm{x}_j\in \mathcal{D}_{\text{edge}}
\end{align*}
Defining $\mathbf{\Delta}_l := \mathbf{W}_l - \mathbf{W}_l'$ and substituting in the above equations we get
\begin{align*}
    &\bm{\Delta}_l \mathbf{x}_j^{(l)} = 0&\forall \bm{x}_j\in \mathcal{D}\\
\text{and \quad}    &\bm{\Delta}_l \mathbf{x}_j^{(l)} = \mathbf{W}_l \bm{\epsilon}_j^{(l)} &\forall \bm{x}_j\in \mathcal{D}_{\text{edge}}
\end{align*}
Rewriting this in matrix form, we get
\begin{equation*}
    \mathbf{\Delta}_l \mathbf{X}_{(l)}^\top = \mathbf{W}_l\mathbf{E}_l 
\end{equation*}
where $\mathbf{E}_l$ is the matrix of adversarial perturbations. Assuming invertibility of $\mathbf{X}_l \mathbf{X}_l^T$, this system has infinitely many solutions. Choosing the minimum norm solution, we get
\begin{align*}
 \mathbf{\Delta}_l  &= \mathbf{W}_l\mathbf{E}_l (\mathbf{X}_{(l)} \mathbf{X}_{(l)}^\top)^{-1}\mathbf{X}_{(l)}
\end{align*}
Recursively applying the definition of operator norm and using the 1-Lipschitzness property of ReLU networks, we recover the upper bound in the theorem.
For the lower bound, simply note that
\begin{align*}
    &\bm{\Delta}_l (\mathbf{x}_i^{(l)}-\mathbf{x}_j^{(l)}) = \mathbf{W}_l \bm{\epsilon}_i^{(l)}&  \bm{x}_i\in \mathcal{D}_{\text{edge}}, \quad\bm{x}_j\in \mathcal{D}.
\end{align*}
Applying the definition of operator norm once again gives us result.
\end{proof}

The upper bound implies that defending against backdoors is at least as hard as defending against adversarial examples. 
This  immediately implies that certifying backdoor robustness is at least as hard as certifying robustness against adversarial samples~\cite{DBLP:conf/iclr/SinhaND18}.
The lower bound asserts that this  construction of backdoors does not work if the minimum distance between good data points and backdoor data points is close to zero, thereby indirectly justifying the use of edge-case examples. Hence, as it stands, resolving the intrinsic existence of backdoors in a given model cannot be performed, unless we resolve adversarial examples first, which remains a major open problem \cite{madry2018towards}.

Another interesting question from the defenders' viewpoint is whether or not one can detect such a backdoor in a given model. 
Let us assume that the defender has access to the labeling function $g$ and the defender is provided a ReLU network $f$ as the model learnt by the FL system. Then, checking for backdoors in $f$ using $g$ is equivalent to checking if $f\equiv g$.
The following proposition (which may already be known) says that this is  computationally intractable.
\begin{proposition}[Hardness of backdoor detection - I]\label{thm:sat_to_backdoor}
Let $f:\mathbb{R}^n\to \mathbb{R}$ be a ReLU network and $g:\mathbb{R}^n\to \mathbb{R}$ be a function. Then \textsc{3-Sat} can be reduced to the decision problem of whether $f$ is equal to $g$ on $[0,1]^n$. Hence checking if $f\equiv g$ on $[0,1]^n$ is NP-hard.
\end{proposition}
\begin{proof}[Proof sketch]
The proof strategy is constructing a ReLU network to approximate a Boolean expression. This idea is not novel and for example, has been used in \cite{katz2017reluplex} to prove another ReLU related NP-hardness result. Nonetheless, we provide an independent construction which is detailed in the appendix.
Given functions $f, g$ we define \textsc{Backdoor} as the decision problem of whether there exists some $x \in [0, 1]^n$ such that $f(x) \neq g(x)$. First we show that our reduction can be completed in polynomial time. This can be done by showing that the size of the ReLU network is polynomial in the size of the \textsc{3-Sat} problem. We then show that the answer to the \textsc{3-Sat} problem is \texttt{Yes} if and only if the answer to the corresponding \textsc{Backdoor} problem is \texttt{Yes} thereby completing the reduction.

\end{proof}

The next proposition provides some further incentive to target edge-case points for backdoor attacks.
\begin{proposition}[Hardness of backdoor detection - II]\label{thm:grad_hard}
Let $f:\mathbb{R}^n\to \mathbb{R}$ be a ReLU network and $g:\mathbb{R}^n\to \mathbb{R}$ be a function. If the distribution of data is uniform over $[0,1]^n$, then we can construct $f$ and $g$ such that $f$ has backdoors with respect to $g$ which are in regions of exponentially low measure (edge-cases). Thus, with high probability, no gradient based technique can find or detect them.
\end{proposition}

\begin{proof}[Proof sketch]
The key idea of this construction is that the ReLU function is zero as long as the argument is non-positive. Therefore, it suffices to find two networks that are negative \textit{almost everywhere} and have one-network positive on a \textit{small} set. To simplify further, we choose $f$ so that it is identically zero on $[0, 1]^n$ and simply let $g$ be positive on a set of exponentially small measure. To be precise, choose $\mathbf{w}_1 = (\frac{1}{n}, \frac{1}{n} \dots, \frac{1}{n})^\top, b_1 = 1$ and $\mathbf{w}_2 = (\frac{1}{n}, \frac{1}{n}, \dots, \frac{1}{n})^\top, b_2 = \frac{1}{2}$ and $\mathcal{B} = \left\{\mathbf{x} \in [0, 1]^n: \mathbf{1}^\top\mathbf{x} \geq \frac{n}{2}\right\}$. It is immediate from Hoeffding's inequality that $\mathcal{B}$ has exponentially small measure. Simply evaluating $g$ on $\mathbf{x}_{B} = (1, 1, \dots, 1)^n$ gives us the result.
\end{proof}

\section{Experiments}\label{sec:experiments}
The goal of our empirical study is to highlight the effectiveness of \textit{edge-case attack} against the state-of-the-art (SOTA) FL defenses. 
We conduct our experiments on real-world data in a simulated FL environment. Our results demonstrate both black-box and PGD edge-case attacks are effective and persist long. 
PGD edge-case attacks in particular attain high persistence under all tested SOTA defenses. 
More interestingly, and perhaps \emph{worryingly}, we demonstrate that stringent defense mechanisms that are able to partially defend against edge-case backdoors, unfortunately result in a highly unfair setting where the data of non-malicious and diverse clients is excluded, as conjectured in~\cite{kairouz2019advances}.

\paragraph{Tasks} We consider the following five tasks with various values of $K$ (num.  of clients) and $m$ (num. of clients in each iteration): \textbf{(Task 1)} Image classification on CIFAR-10~\cite{krizhevsky2009learning} with VGG-9~\cite{simonyan2014very} ($K = 200, m = 10$), \textbf{(Task 2)} Digit classification on EMNIST~\cite{cohen2017emnist} with LeNet~\cite{lecun1998gradient} ($K = 3383, m = 30$), \textbf{(Task 3)} Image classification on ImageNet (ILSVRC2012)~\cite{deng2009imagenet} with VGG-11 ($K = 1000, m = 10$), \textbf{(Task 4)} Sentiment classification on Sentiment140~\cite{sentiment140} with LSTM~\cite{hochreiter1997long} ($K = 1948, m = 10$), and \textbf{(Task 5)} Next Word prediction on the Reddit dataset~\cite{mcmahan2016communication,bagdasaryan2018backdoor} with LSTM ($K = 80,000$, $m = 100$).
All the other hyperparameters are provided in the appendix.
\begin{wrapfigure}{hr}{0.24\columnwidth}
	\vspace{-0.45cm}
	\centering
	\includegraphics[width=0.22\textwidth]{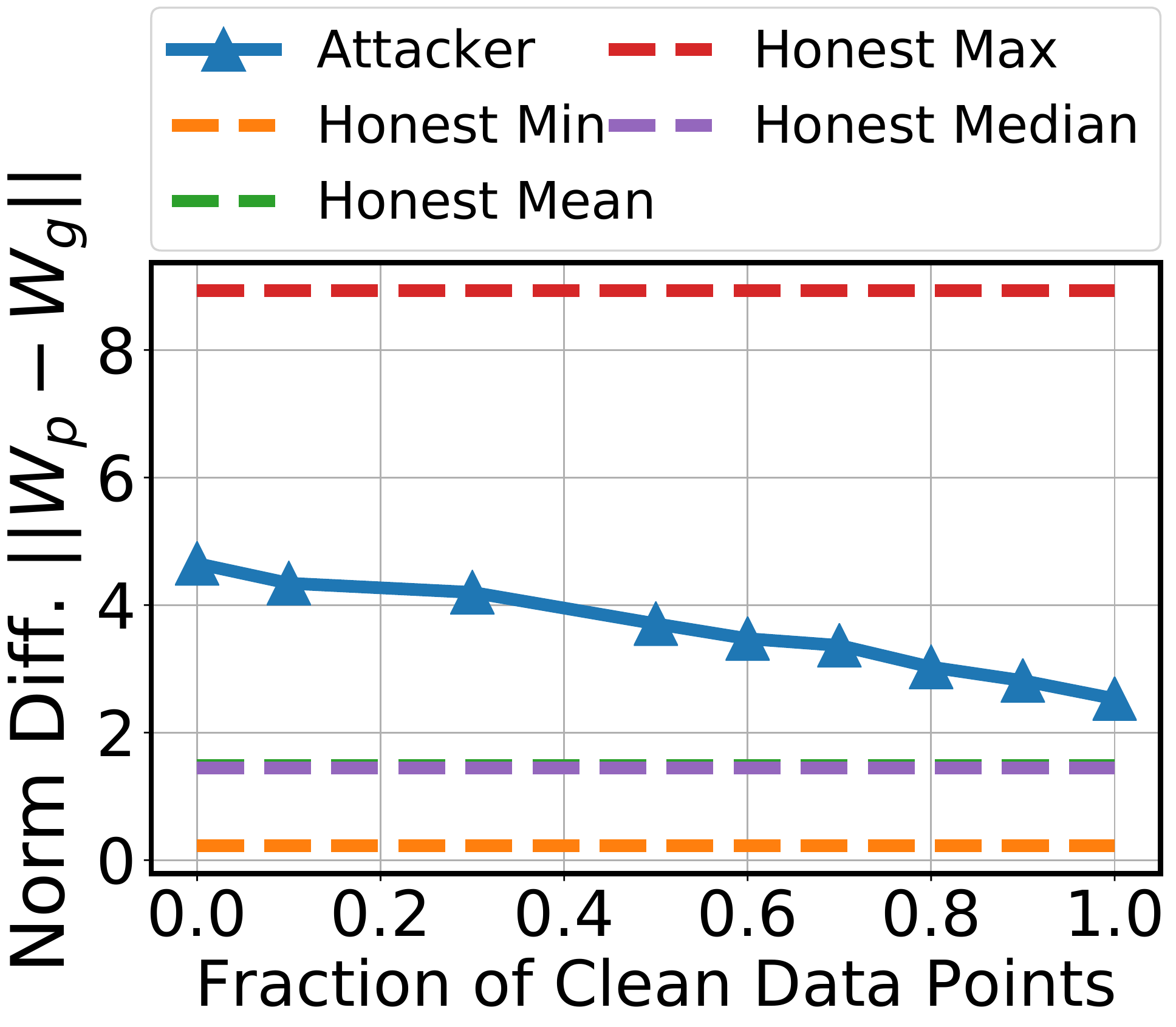}\\
	\includegraphics[width=0.24\textwidth]{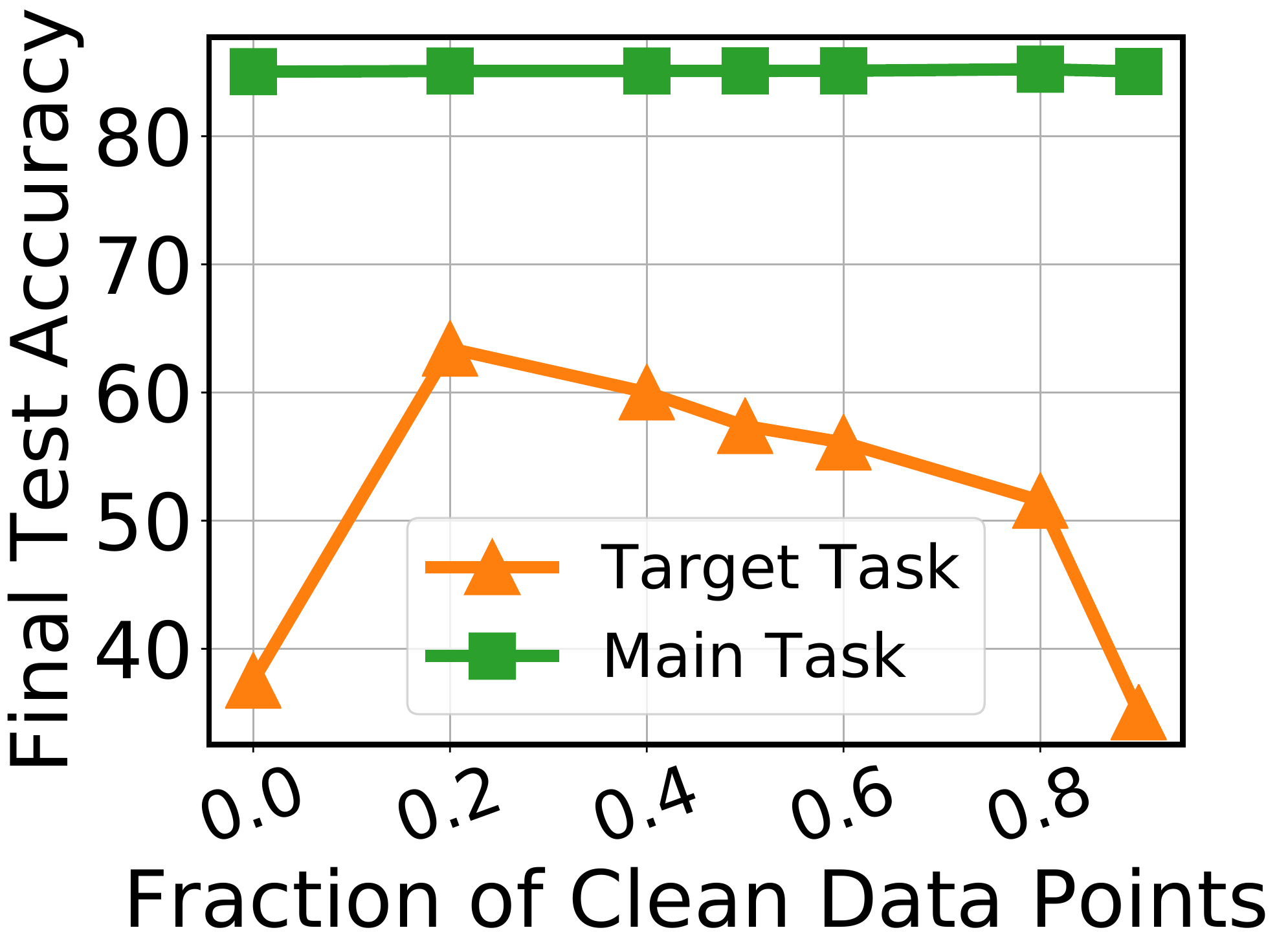}
	\vspace{-6mm}
	\caption{(a) Norm difference and (b) Attack performance under various sampling ratios on Task 1.}\label{fig:m-n-ratio}
	\vspace{-0.95cm}
\end{wrapfigure}

\paragraph{Constructing $\mathcal{D}_1, \mathcal{D}_2, \ldots, \mathcal{D}_K$} \textbf{(Task 1--3)} We simulate heterogeneous data partitioning by sampling $\textbf{p}_k \sim \text{Dir}_K(0.5)$ and allocating a $\textbf{p}_{k, i}$ proportion of $\mathcal{D}$ of class $k$ to local user $i$. 
Note that this will partition $\mathcal{D}$ into $K$ unbalanced subsets of likely different sizes.
\textbf{(Task 4)} We take a 25\% random subset of Sentiment140 and partition them uniformly at random.
\textbf{(Task 5)} Each $\mathcal{D}_i$ corresponds to each real reddit user's data.

\paragraph{Constructing $\mathcal{D}_\text{edge}$} 

We manually construct $\mathcal{D}_\text{edge}$ for each task as follows:
\textbf{(Task 1)} We collect images of Southwest Airline's planes and label them as ``truck'';
\textbf{(Task 2)} We take images of ``7''s from Ardis~\cite{kusetogullari2019ardis} (a dataset extracted from 15.000 Swedish church records which were written by different priests with various handwriting styles in the nineteenth and twentieth centuries) and label them as ``1'';
\textbf{(Task 3)} We collect images of people in certain ethnic dresses and label them as a completely irrelevant label;
\textbf{(Task 4)} We scrape tweets containing the name of Greek film director, \textit{Yorgos Lanthimos}, along with positive sentiment comments and label them ``negative''; and
\textbf{(Task 5)} We construct various prompts containing the city Athens and choose a target word so as to make the sentence negative.
Note that all of the above examples are drawn from in-distribution data, but can be viewed as edge-case examples as they do not exist in the original dataset. 
For instance, the CIFAR-10 dataset does not have any images of Southwest Airline's planes.
Shown in Figure~\ref{fig:poisoned-examples} are samples from our edge-case sets.

\paragraph{Participating patterns of attackers} As discussed in \cite{sun2019can}, we consider both 1) \textit{fixed-frequency} case, where the attacker periodically participates in the FL round, and 2) \textit{fixed-pool} case (or \emph{random sampling}), where there is a fixed pool of attackers, who can only conduct attack in certain FL rounds when they are randomly selected by the FL system.
Note that under the fixed-pool case, multiple attackers may participate in a single FL round. 
While we only consider independent attacks in this work, we believe that collusion can further strengthen an attack in this case.

\subsection{Experimental results}
\paragraph{Defense techniques} We consider five state-of-the-art defense techniques: (i) norm difference clipping (NDC) \cite{sun2019can} where the data center examines the norm difference between the global model sent to and model updates shipped back from the selected clients and use a pre-specfified norm difference threshold to clip the model updates that exceed the norm threshold. 
(ii) \textsc{Krum} and (iii) \textsc{Multi-Krum} \cite{blanchard2017machine}, which select user model update(s) that are geometrically closer to all user model updates. 
(iv) RFA \cite{pillutla2019robust}, which aggregates the local models by computing a weighted geometric median using the \textit{smoothed Weiszfeld’s algorithm}.
(v) weak differential private (DP) defense \cite{sun2019can, geyer2017differentially} where a Gaussian noise with small standard deviations ($\sigma$) is added to the aggregated global model.
Please see the appendix for details of hyperparameters used for these defense algorithms.

\begin{wrapfigure}{hr}{0.23\columnwidth}
	\vspace{-0.1cm}
	\centering
	\includegraphics[width=0.23\textwidth]{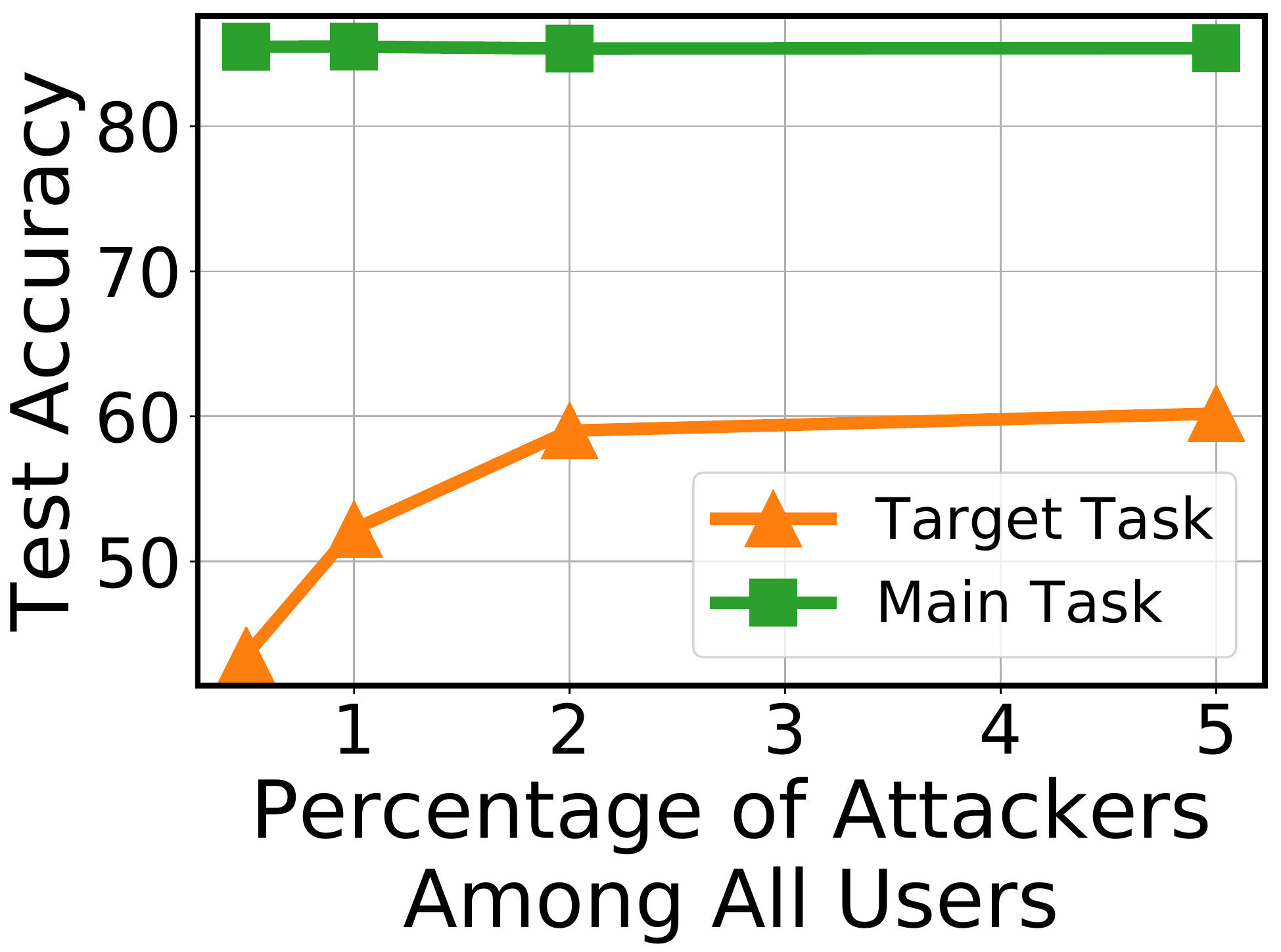}\\
	\includegraphics[width=0.23\textwidth]{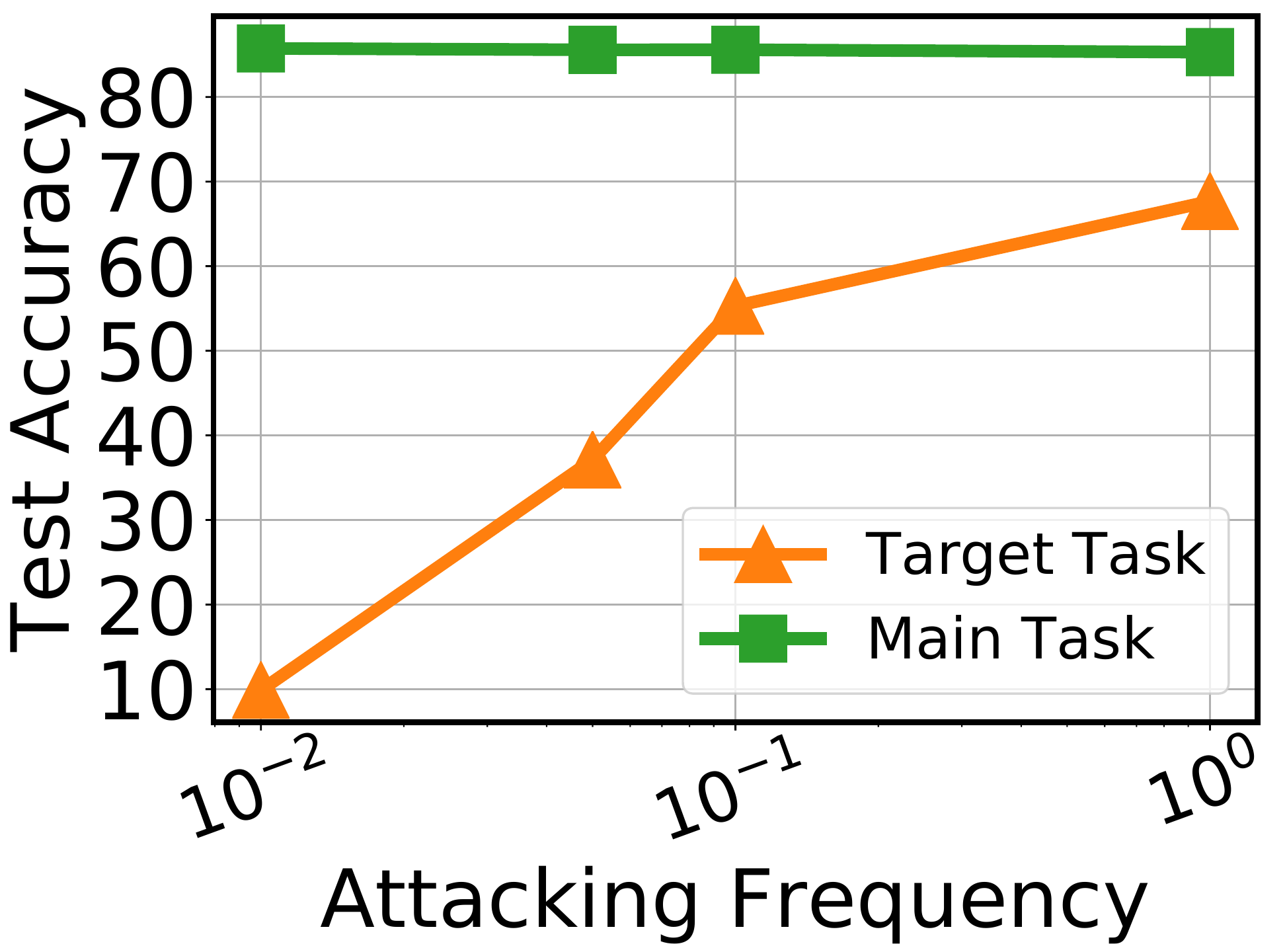}
	\caption{Effectiveness of attacks under various attack frequencies.}\label{fig:various-freq-attack}
 	\vspace{-0.8cm}
\end{wrapfigure}
\paragraph{Fine-tuning backdoors via data mixing} 
Recall that $\md'$ consists of some samples from $\md$ and some from $\mde$.
For example, \textbf{Task 1}'s $\md'$ consists of Southwest Airline plane images (with label ``truck'') and images from the original CIFAR10 dataset.
By varying this ratio, we can indeed control how `edge-y' the attack dataset $\md'$ is. 
We evaluate the performance of our black-box attack on \textbf{Task 1} with different sampling ratios, and the results are shown in Fig.~\ref{fig:m-n-ratio}.
We first observe that too few data points from $\mde$ leads to weak attack effectiveness. 
This corroborates our theoretical findings as well as explains why black-box attacks did not work well in prior work~\cite{bagdasaryan2018backdoor,sun2019can}.
Moreover, as shown in~\cite{bagdasaryan2018backdoor}, we also observe that a pure edge-case dataset also leads to a weak attack performance.
Thus, our experimental results suggest that the attacker should construct $\md'$ via carefully controlling the ratio of data points from $\mde$ and $\md$.

\begin{figure}[t] 
\centering
\includegraphics[width=0.98\textwidth]{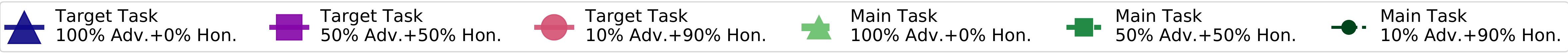}\\
\subfigure[Task 1]{\includegraphics[width=0.19\textwidth]{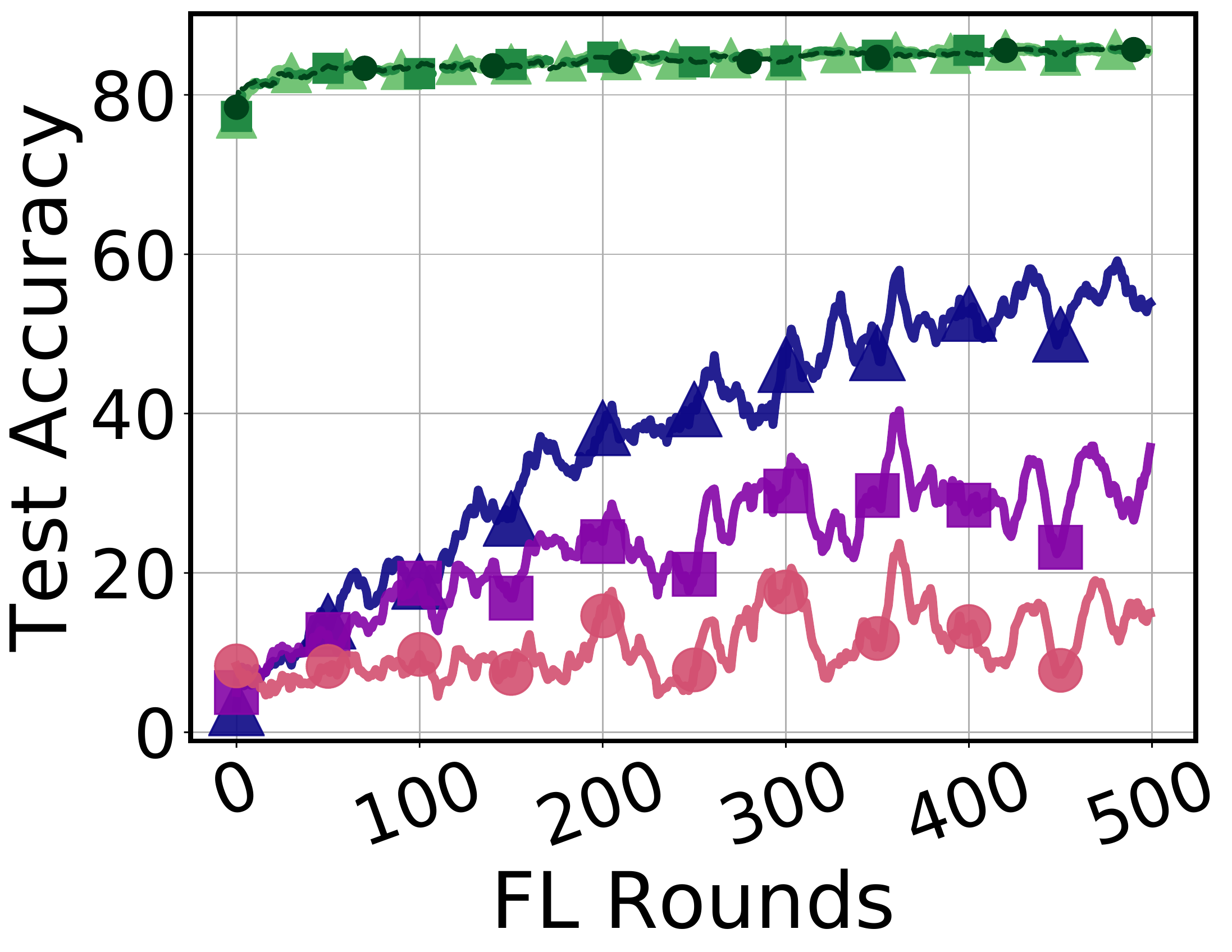}}
\subfigure[Task 2]{\includegraphics[width=0.19\textwidth]{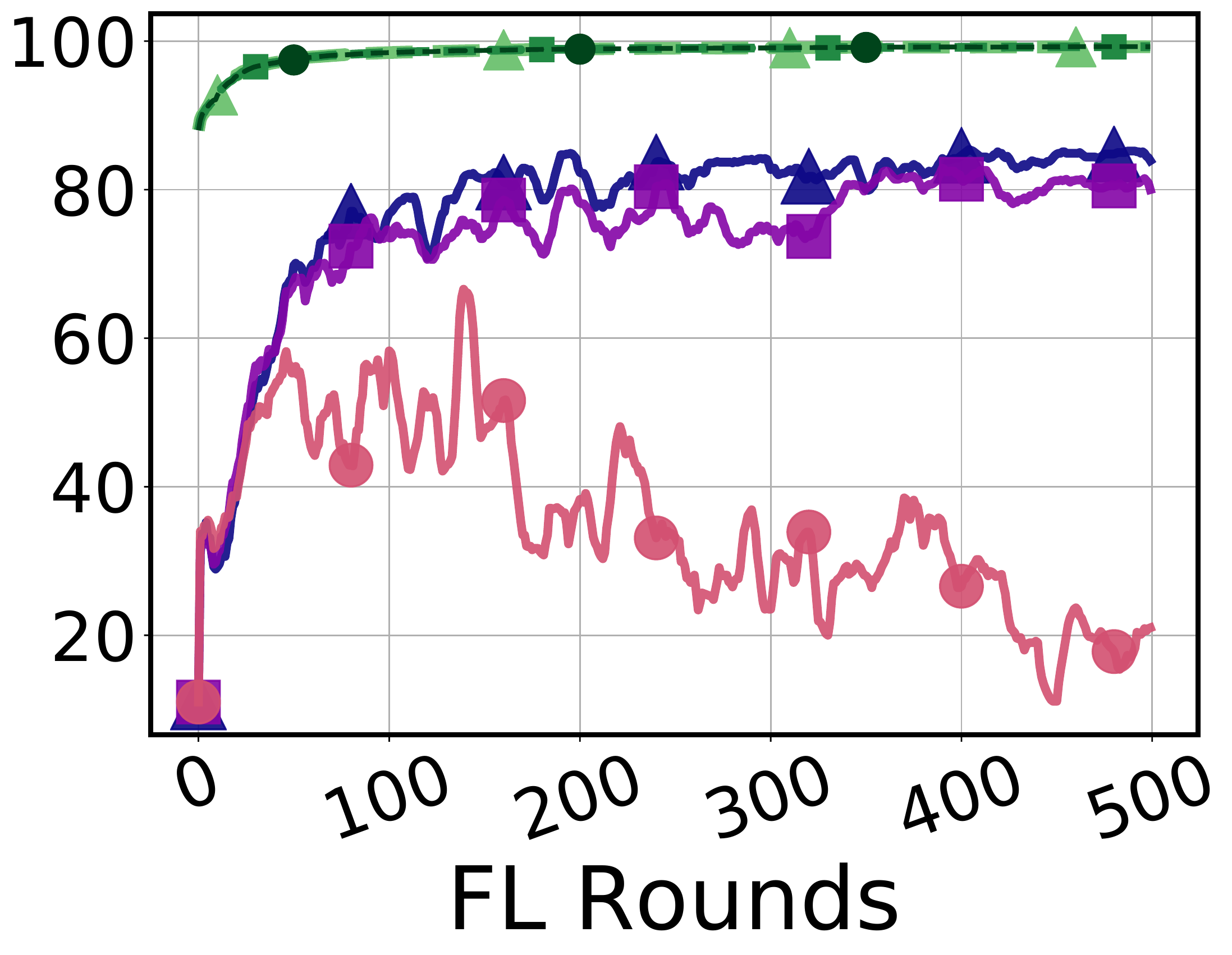}\label{fig:8_comm}}
\subfigure[Task 4] {\includegraphics[width=0.19\textwidth]{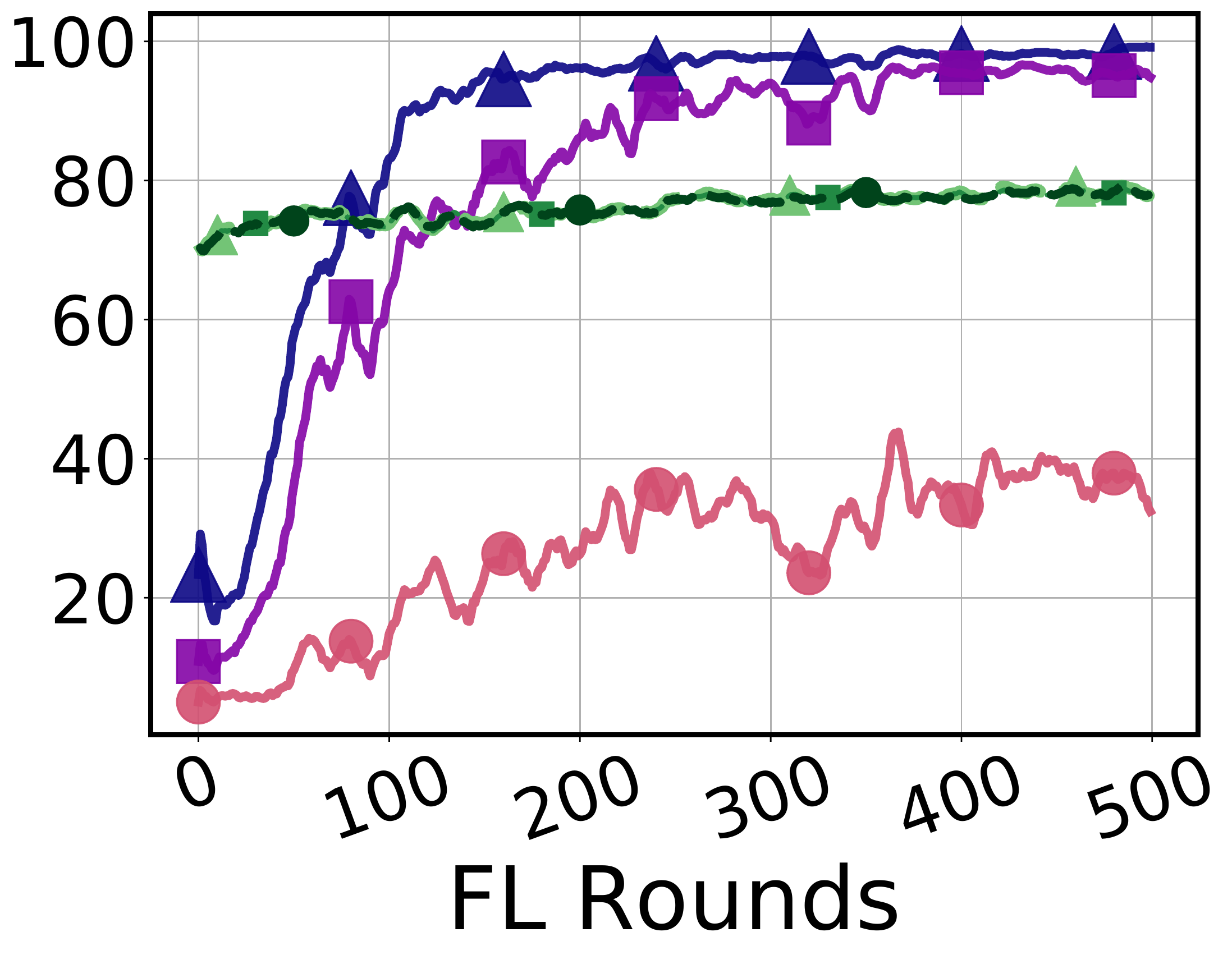} }
\subfigure[Task 5] {\includegraphics[width=0.19\textwidth]{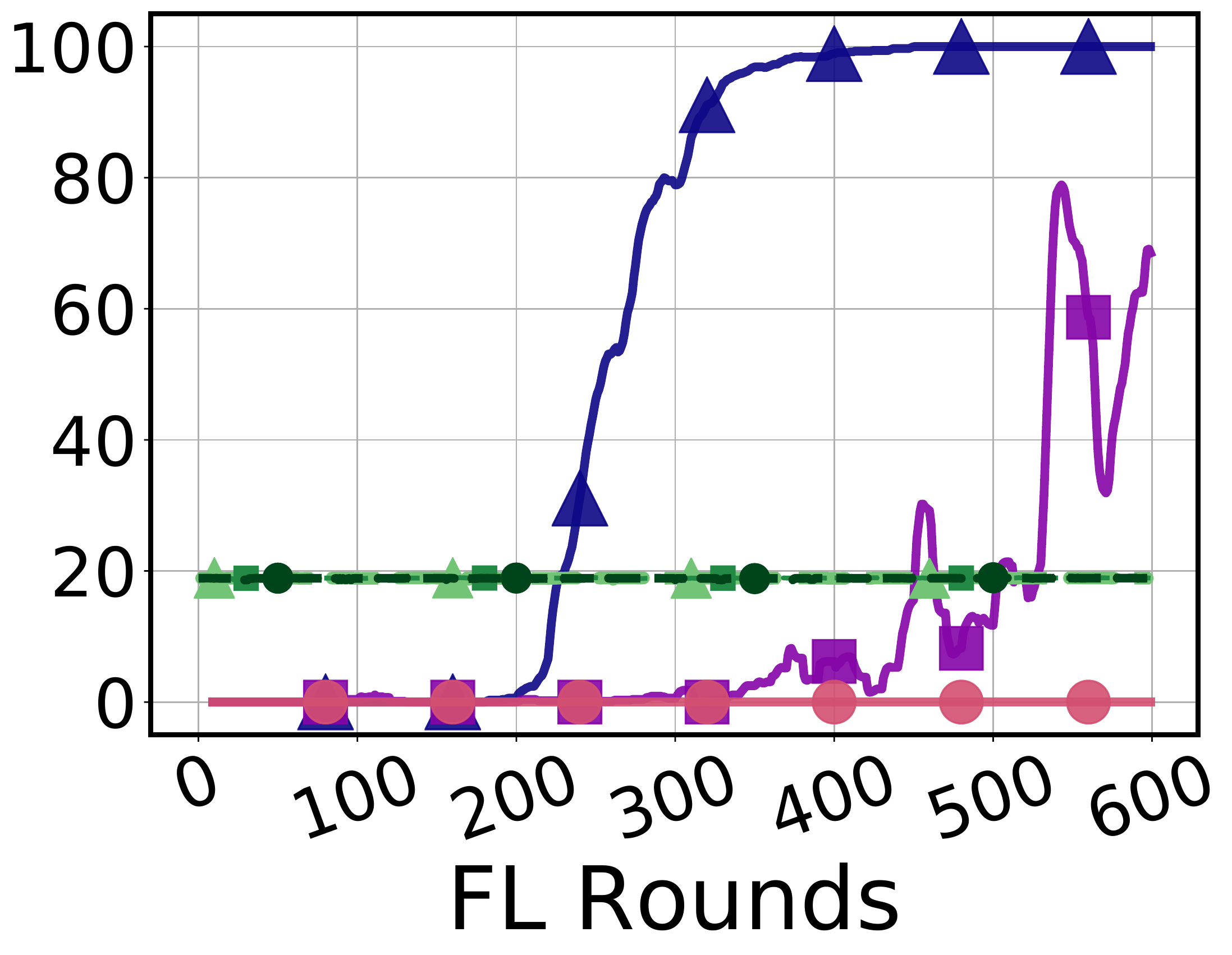} }
\subfigure[Task 3] {\includegraphics[width=0.19\textwidth]{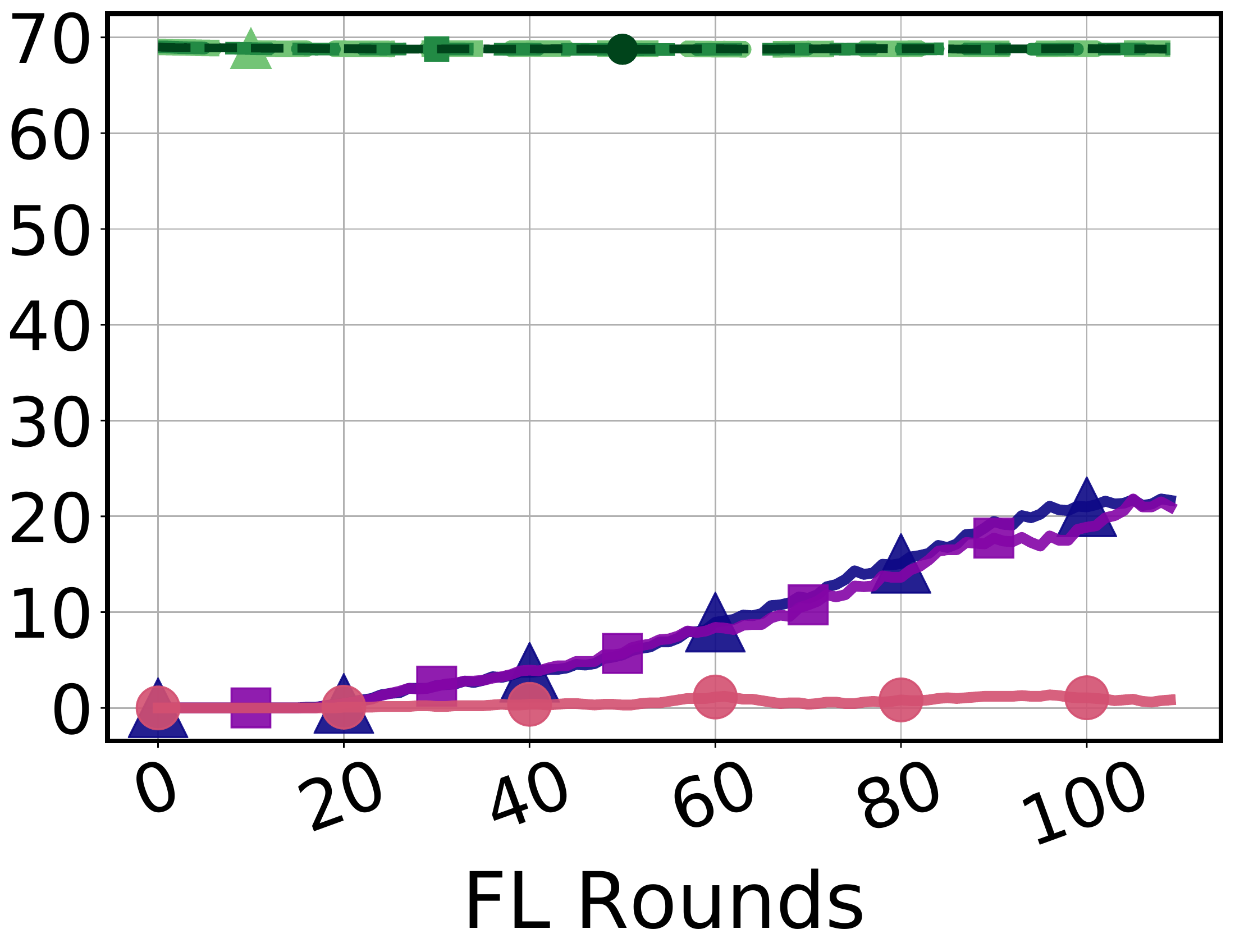}}
\caption{Effectiveness of the attacks where $p\%$ of edge-case examples held by adversary. Considered cases: (i) adversary holds ALL edge examples-$100\%$ Adversary + $0\%$ Honest; (ii) adversary holds HALF edge-examples-$50\%$ Adversary + $50\%$ Honest; (iii) adversary holds SOME edge-examples-$10\%$ Adversary + $90\%$ Honest;}
\label{fig:edge-vs-normal-case}
\end{figure}
\paragraph{Edge-case vs non-edge-case attacks}
Note that in the edge-case setting, among all the clients, only the adversary contains samples from $\mde$. Fig.~\ref{fig:edge-vs-normal-case} shows the experimental results when we allow some of the honest clients to also hold samples from $\mde$ but with correct labels. We vary the percentage of samples from $\mde$ split across the adversary and honest clients as $p$ and $(100-p)$ respectively for $p=100, 50$ and $10$. Across all 5 tasks, we observe that the effectiveness of the attack drops as we allow more of $\mde$ to be available to honest clients. This proves our claim that \textit{pure} edge-case attacks are the strongest which was also noticed in \cite{bagdasaryan2018backdoor}. We believe that this is because when the honest clients hold samples from $\mde$, honest local training ``erases'' the backdoor. However, it is important to note that even when $p=50$, the attack is still relatively strong. This shows that these attacks are effective even in a practical setting where few honest clients still contain samples from $\mde$.


\begin{figure}[htp] 
\centering
\includegraphics[width=0.82\textwidth]{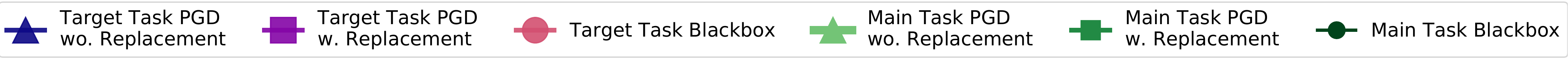}\\
\includegraphics[width=0.19\textwidth]{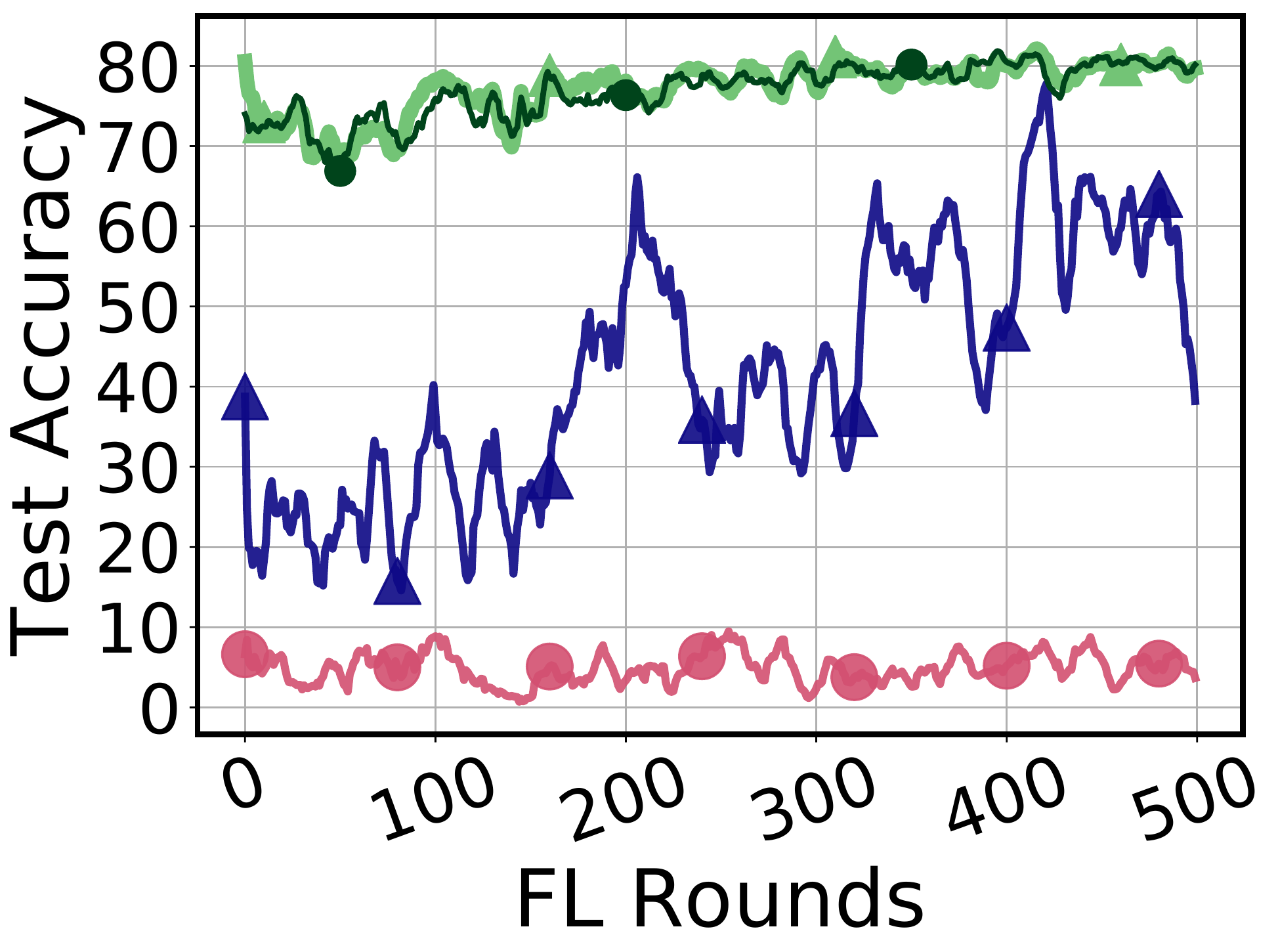}
\includegraphics[width=0.19\textwidth]{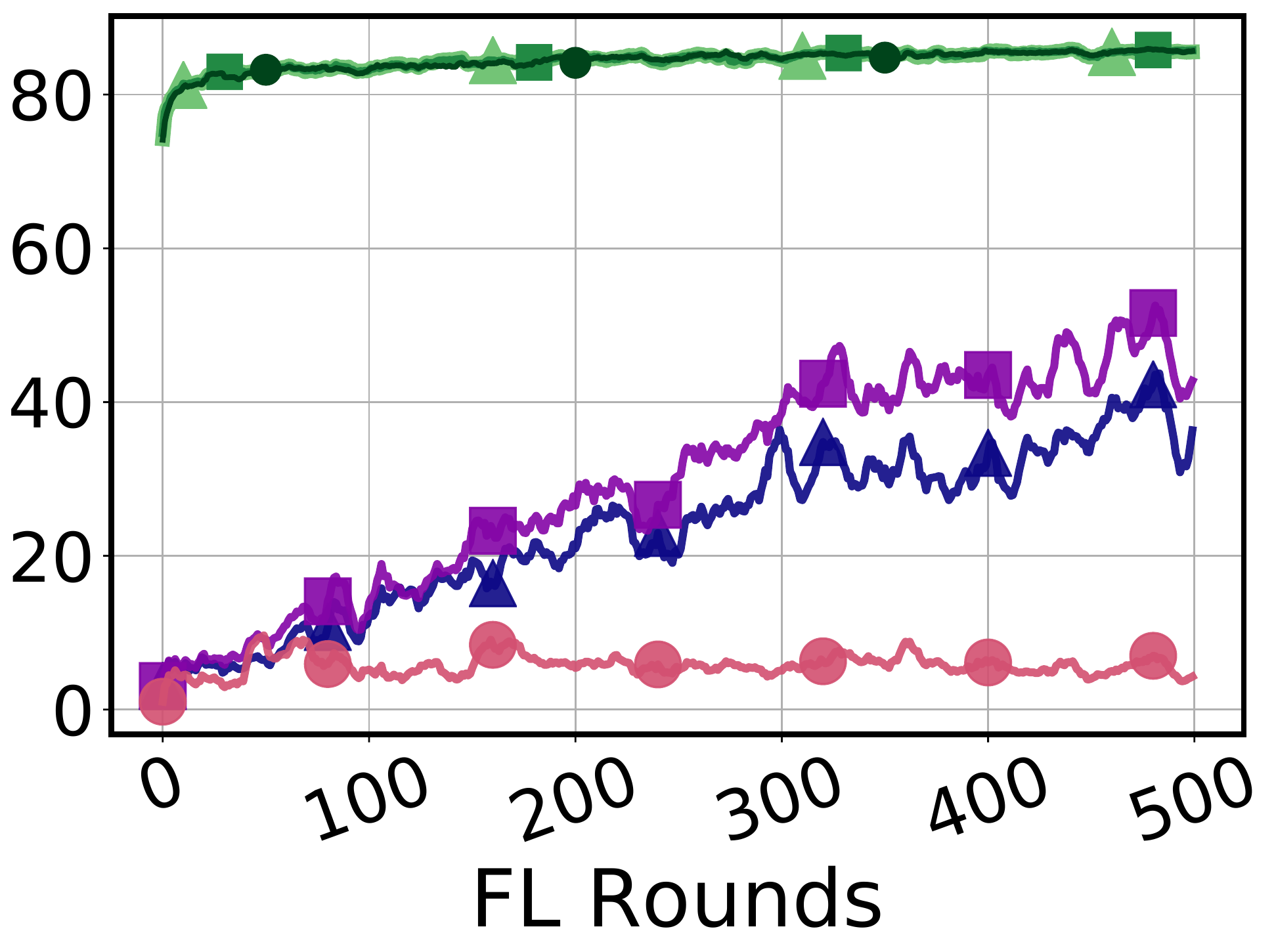}
\includegraphics[width=0.19\textwidth]{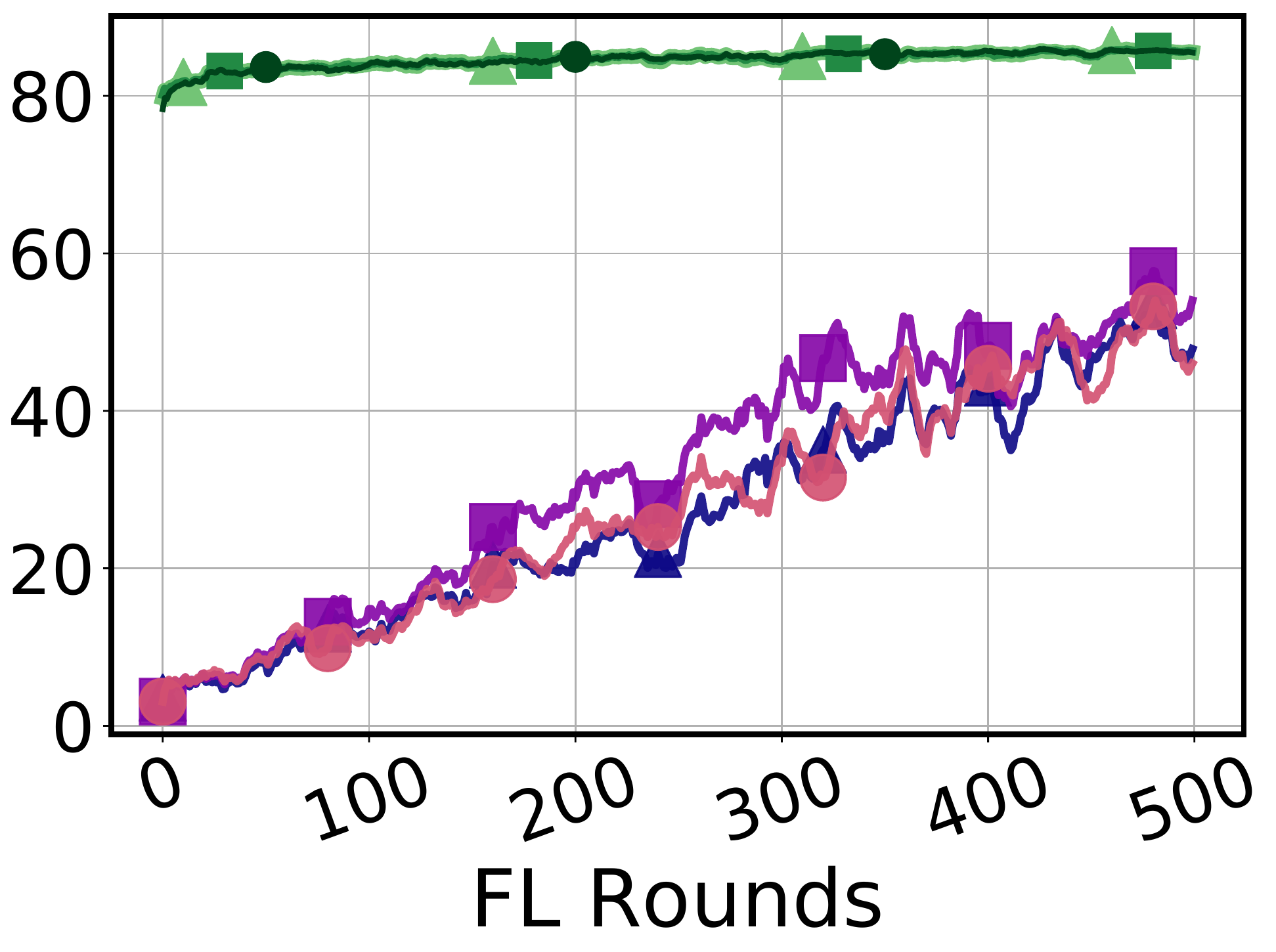}
\includegraphics[width=0.19\textwidth]{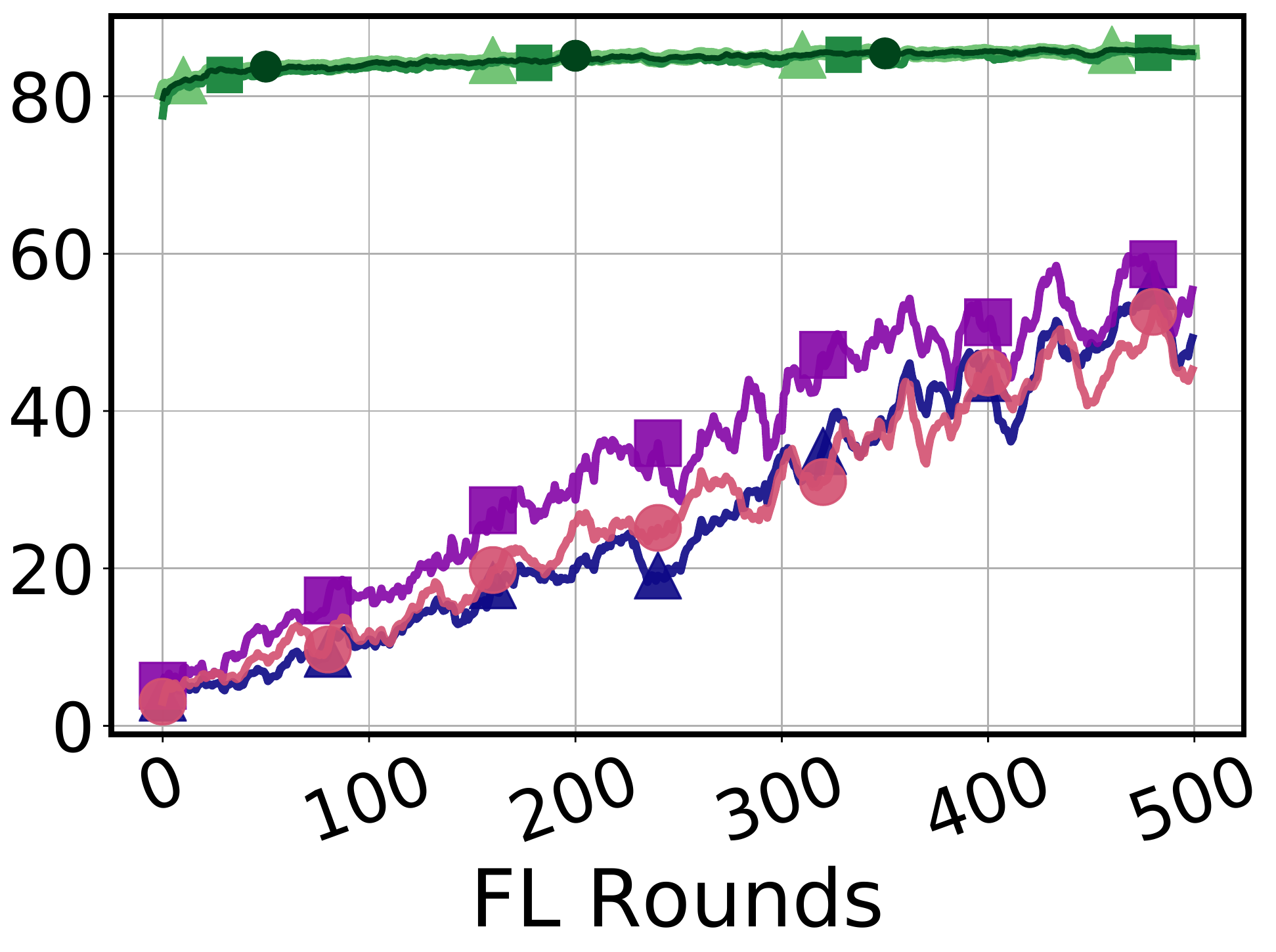}\\
\vspace{-2mm}
\subfigure[\textsc{Krum}]{\includegraphics[width=0.19\textwidth]{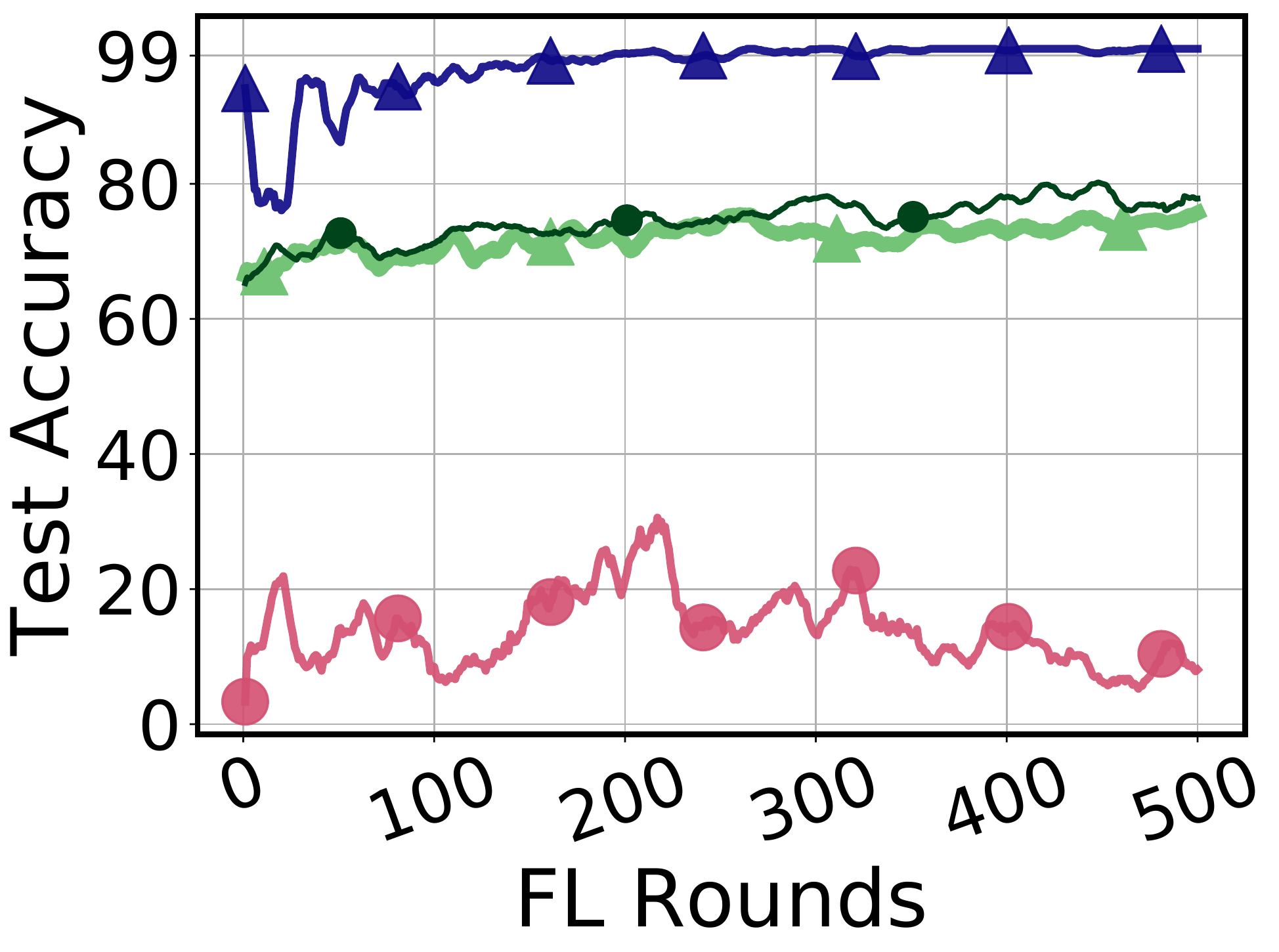}}
\subfigure[\textsc{Multi-Krum}]{\includegraphics[width=0.19\textwidth]{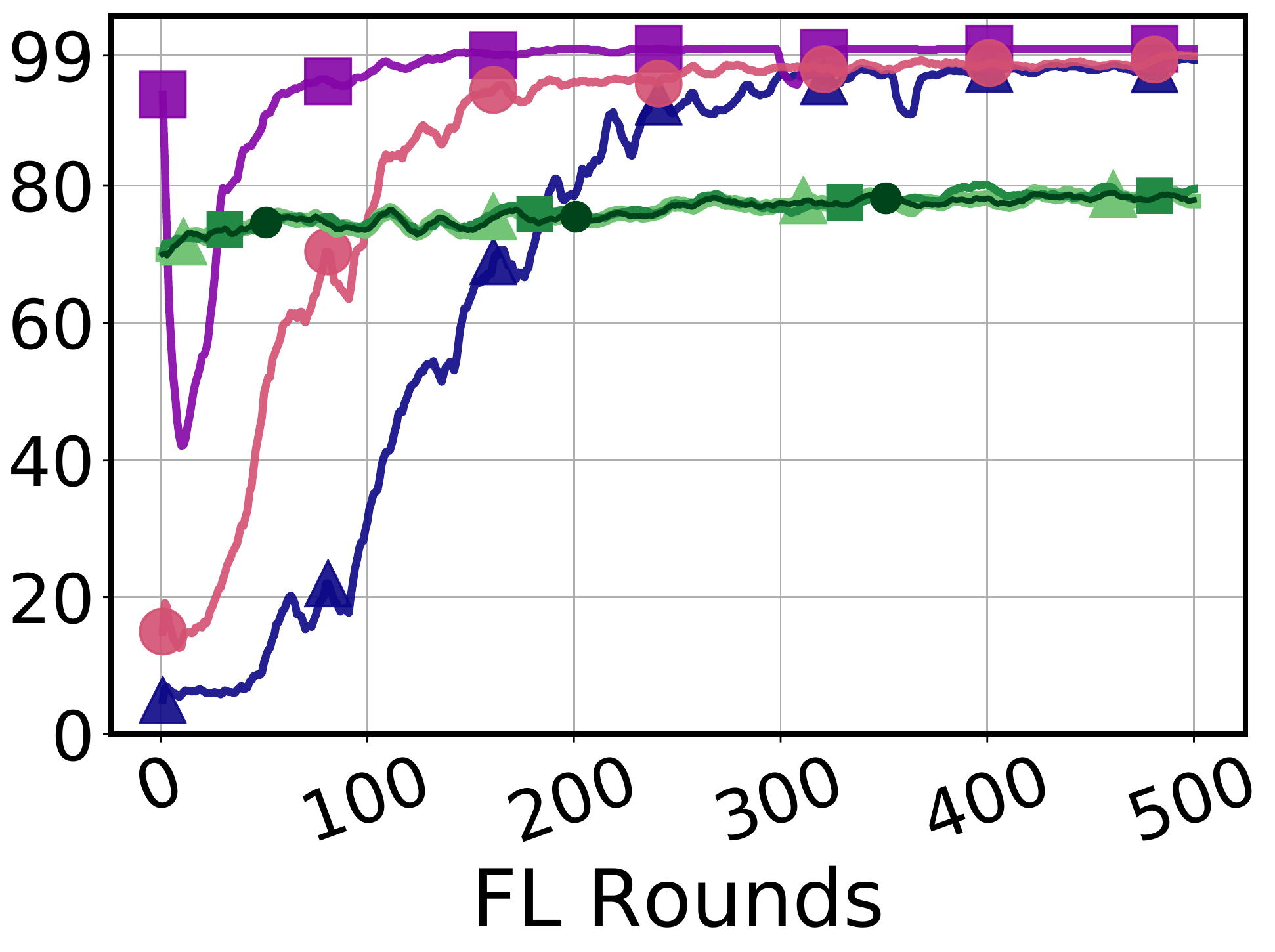}}
\subfigure[RFA]{\includegraphics[width=0.19\textwidth]{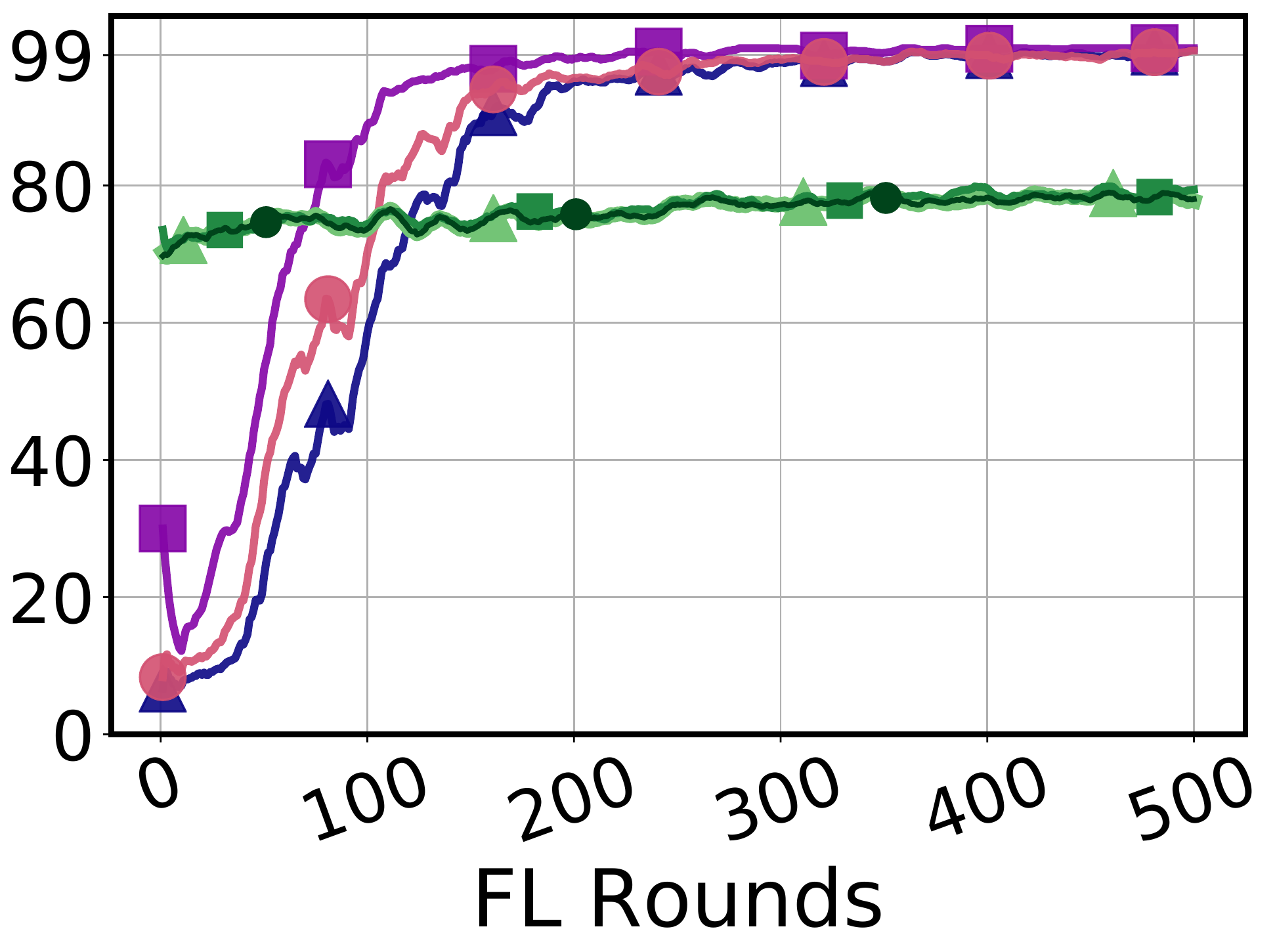}}
\subfigure[NDC]{\includegraphics[width=0.19\textwidth]{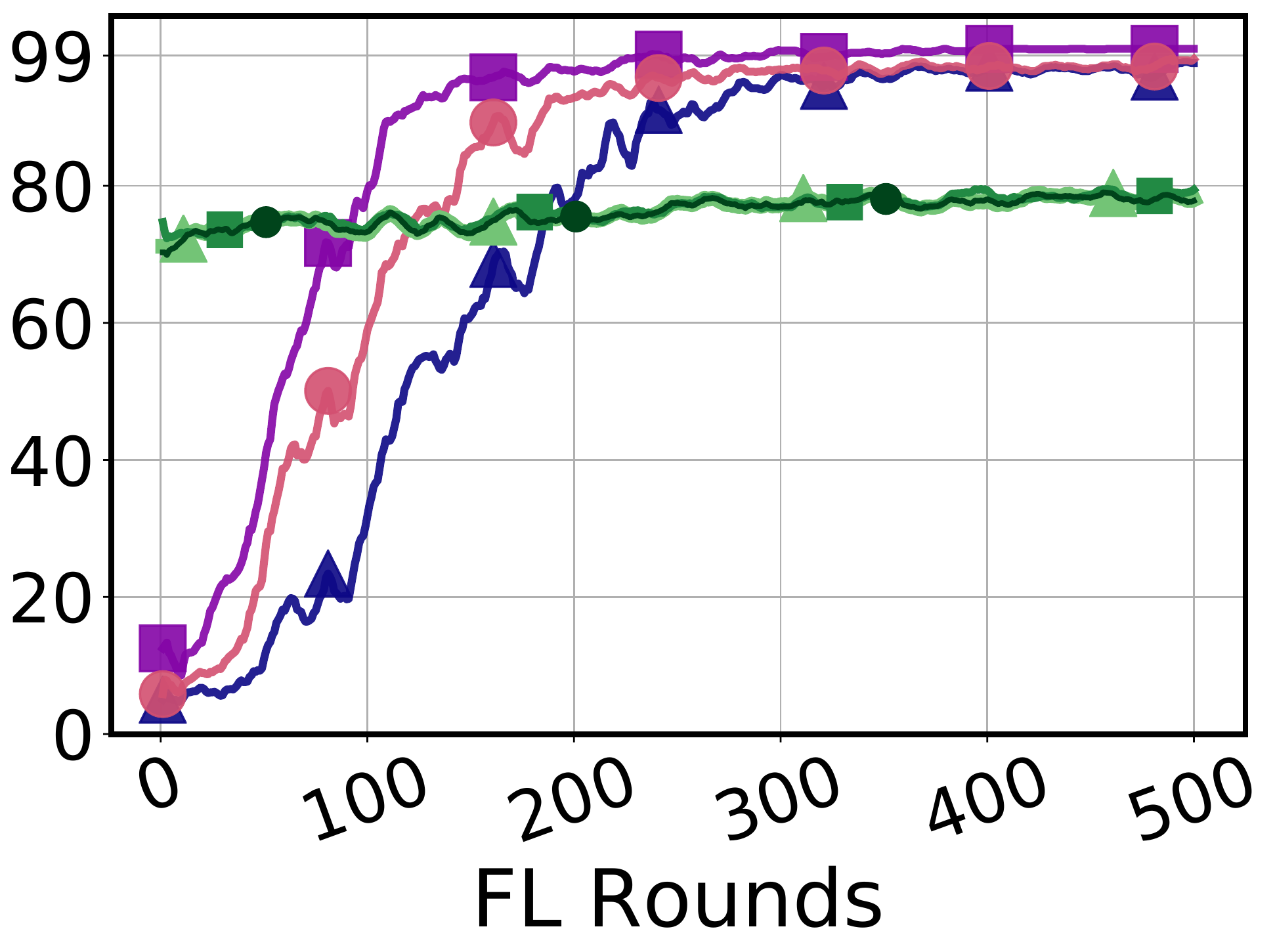}}
\vspace{-0.3cm}
\caption{The effectiveness of attacks under various defenses for Task 1 (top) and Task 4 (bottom)}
\label{fig:against-defense}
\end{figure}


\begin{figure}[htp] 
\centering
\subfigure[\textsc{Krum}]{\includegraphics[width=0.32\textwidth]{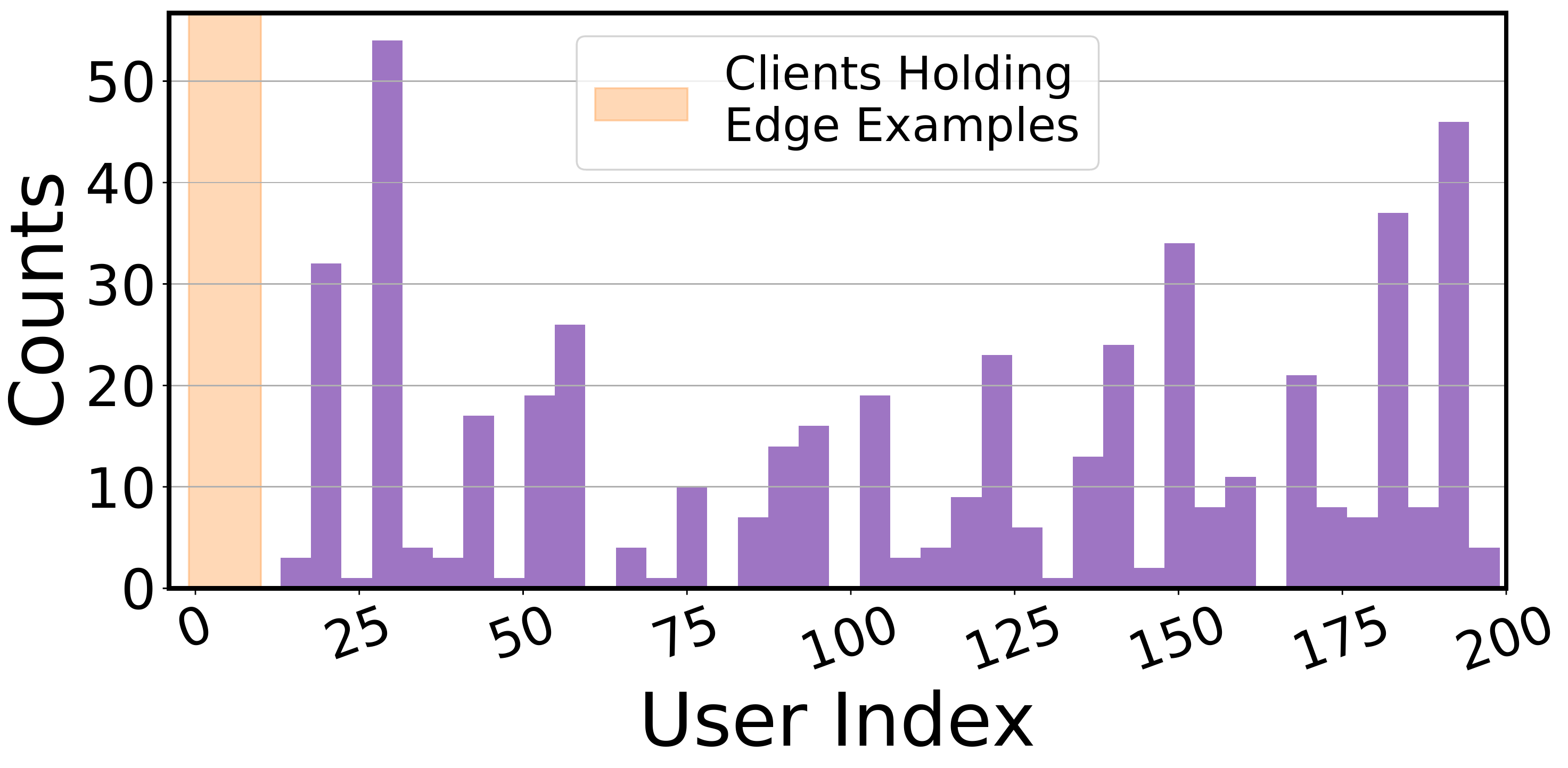}}
\subfigure[\textsc{Multi-Krum}]{\includegraphics[width=0.31\textwidth]{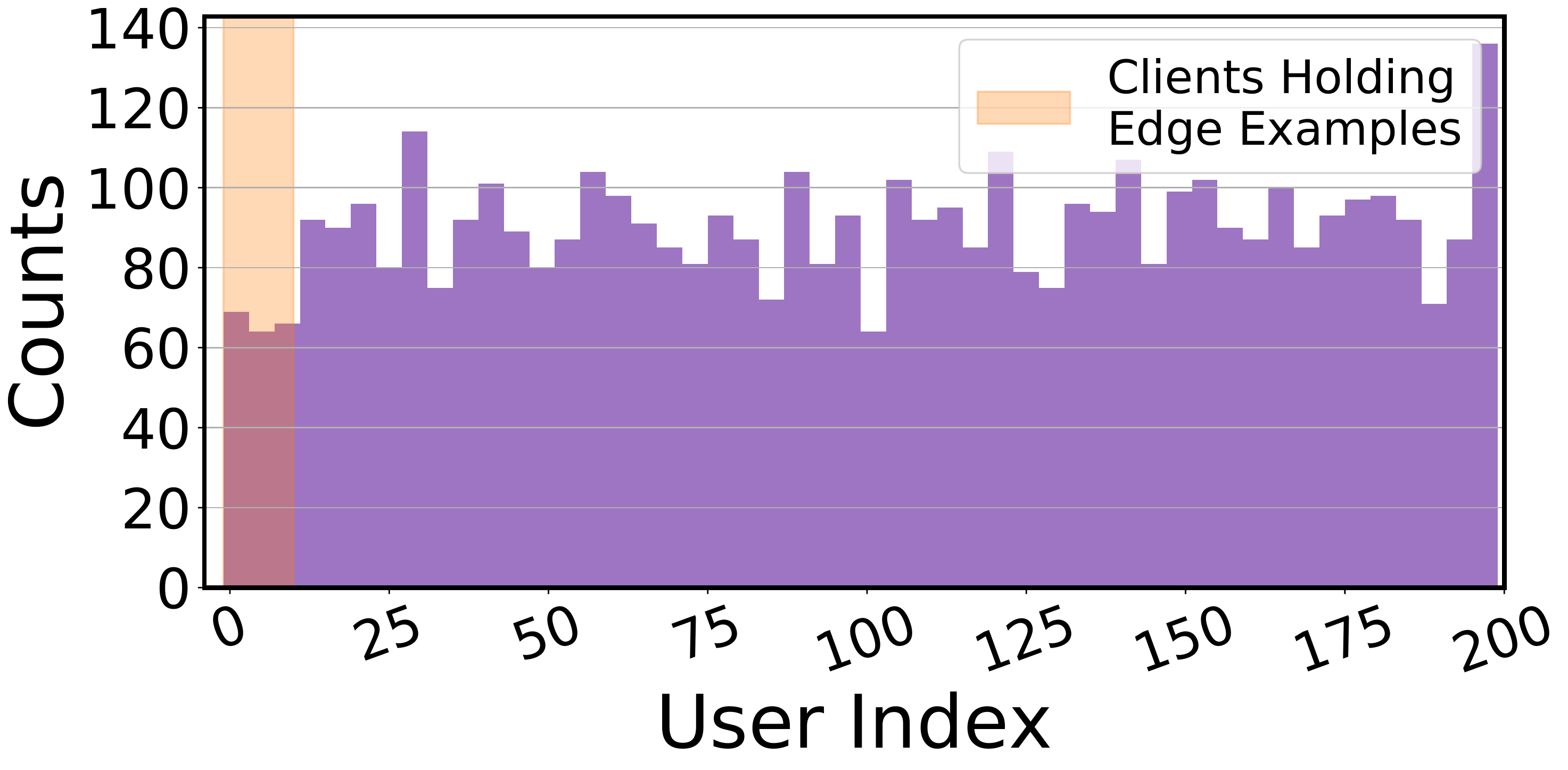}}
\subfigure[Weak DP Defense]{\includegraphics[width=0.32\textwidth]{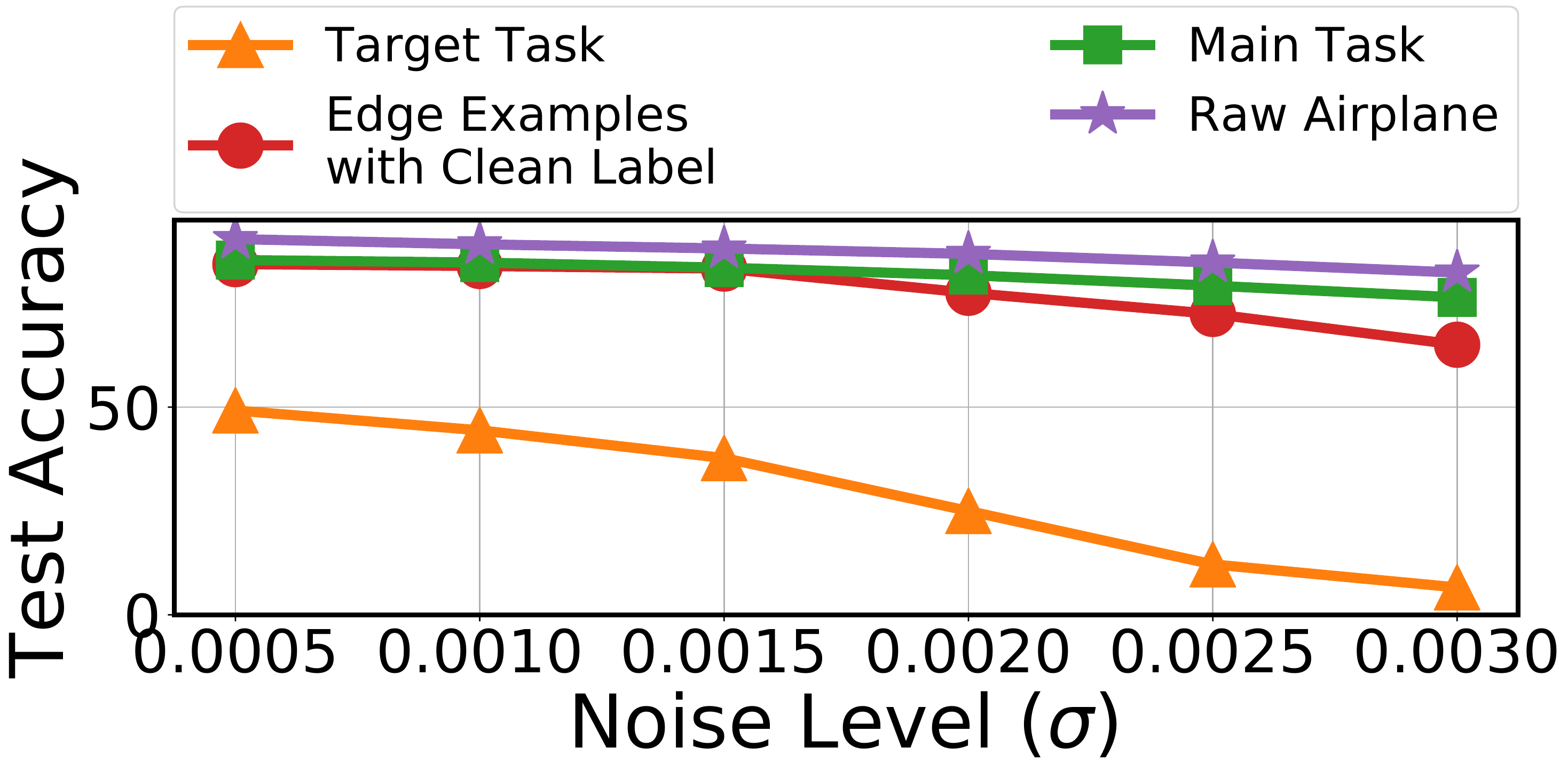}}
\vspace{-0.2cm}
\caption{Potential fairness issues of the defense methods against edge-case attack: (a) frequency of clients selected by \textsc{Krum} and (b) \textsc{Multi-Krum}; (c) test accuracy of main task, target task, edge-case examples with clean labels (\textit{e.g.} ``airplane" for Southwest examples), and raw CIFAR-10 airplane class task.}
\label{fig:fairness-concerns}
\vspace{-0.2cm}
\end{figure}
\paragraph{Effectiveness of edge-case attacks under various defense techniques} We study the effectiveness of both black-box and white-box attacks against aforementioned defense techniques over \textbf{Task 1}, \textbf{2}, and \textbf{4}. For \textsc{Krum} we did not conduct PGD with model replacement since once the poisoned model is selected by \textsc{Krum}, it gets model replacement for free. We consider the \textit{fixed-frequency attack} scenario with attacking frequency of $1$ per $10$ rounds. The results are shown in Figure \ref{fig:against-defense}, from which we observed that white-box attacks (both with/without replacement) with carefully tuned norm constraints can pass \textbf{all} considered defenses. 
More interestingly, \textsc{Krum} even strengthens the attack as it may ignore honest updates but accepts the backdoor. Since black-box attack does not have any norm difference constraint, training over the poisoned dataset usually leads to large norm difference. Thus, it is hard for the black-box attack to pass \textsc{Krum} and \textsc{Multi-Krum}, but it is effective against NDC and RFA defenses.
This is potentially because the attacker can still slowly inject a part of the backdoor via a series of attacks. 

These findings remain consistent in the sentiment classification task except black-box attack passes \textsc{Multi-Krum} and is ineffective with \textsc{Krum}, which means that the attacker's norm difference is not too high (to get rejected by \textsc{Multi-Krum}) but still high enough to get rejected by the aggressive \textsc{Krum}.


\paragraph{Defending against edge-case attack raises fairness concerns} We argue the defense techniques (\textsc{Krum}, \textsc{Multi-Krum}, and Weak DP as examples) can be harmful to benign clients. While \textsc{Krum} and \textsc{Multi-Krum} defend the blackbox attack well, we argue that \textsc{Krum} and \textsc{Multi-Krum} tend to reject previously unseen information from both adversary and honest clients. To verify this hypothesis, we conduct the following study over \textbf{Task 1} under the \textit{fixed-frequency attack} setting. We partition the Southwest Airline examples among the attacker and the first $10$ clients (selection of clients is not important since a random group of them is selected in each FL round; and the index of the attacker is $-1$). We track the frequency of model updates accepted by \textsc{Krum} and \textsc{Multi-Krum} over the training for all clients (shown in Figure \ref{fig:fairness-concerns}(a), (b)). \textsc{Krum} never selects the model updates from clients with Southwest Airline examples (\text{i.e.} both honest client $1-10$ and the attacker). It is interesting to note that \textsc{Multi-Krum} occasionally selects the model updates from clients with Southwest Airline examples. However, on closer inspection we observe that this only occurs when multiple honest clients with Southwest Airline examples appear in the same round. Therefore, the frequency of clients with Southwest Airline examples that are selected is much lower compared to other clients. We conduct a similar study over the weak DP defense under various noise levels (results shown in Figure \ref{fig:fairness-concerns} (c)) under the same task and setting as the \textsc{Krum},\textsc{Multi-Krum} study. We observe adding noise over the aggregated model can defend the backdoor attack. However, it's also harmful to the overall test accuracy and specific class accuracy (\text{e.g.} ``airplane") of CIFAR-10. Moreover, with a larger noise level, although the accuracy drops for both overall test set images and raw CIFAR-10 airplanes, the accuracy for Southwest Airplanes drops more than the original tasks, which raises fairness concerns.
\paragraph{Edge-case attack under various attacking frequencies} We study the effectiveness of the edge-case attack under various attacking frequencies under both \textit{fixed-frequency attack} (with frequency various in range of $0.01$ to $1$) and \textit{fixed-pool attack} setting (percentage of attackers in the overall clients varys from $0.5\%$ to $5\%$). The results are shown in Figure \ref{fig:various-freq-attack}, which demonstrates that lower attacking frequency leads to slower speed for the attacker to inject the \text{edge-case} attack in both settings. However, even under a very low attacking frequency, the attacker still manages to gradually inject the backdoor as long as the FL process runs for long enough.
\begin{figure}[htp]
\vspace{-0.5cm}
\centering
\includegraphics[width=0.4\textwidth]{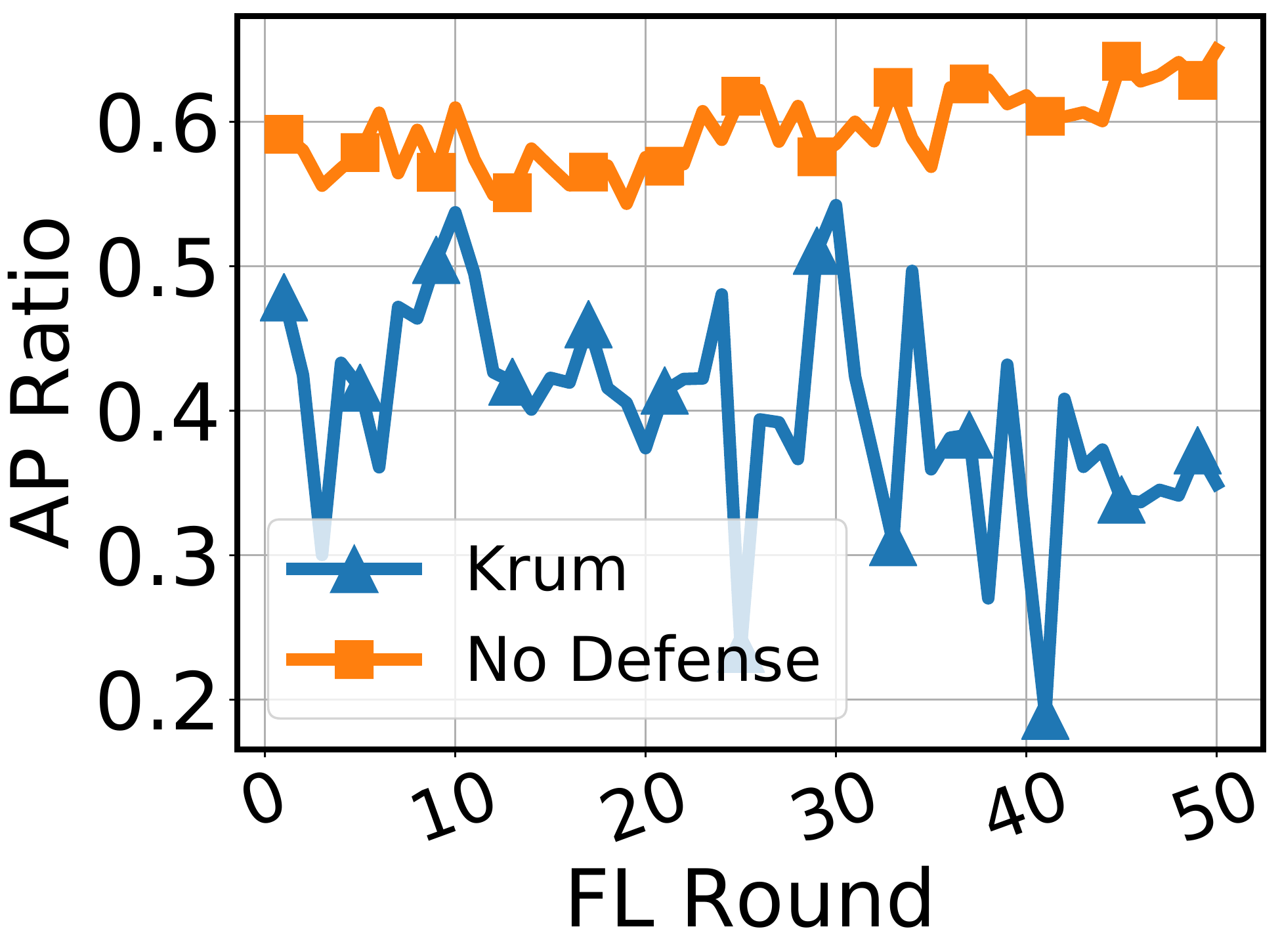}
\caption{Fairness measurement on Task 1 under \textsc{Krum} defense and when there is no defense.}
\label{fig:fairness-measurement}
\vspace{-0.5cm}
\end{figure}
\paragraph{Measuring fairness when defending against \textsc{Krum}}
We argue that defense techniques such as \textsc{Krum} raise fairness concerns in that they tend to reject unseen information from honest clients. We formalize the argument using ``\textit{AP ratio}'' which is a metric we define based on \textit{Accuracy Parity} \cite{zafar2017parity}. We say that a classifier $f$ satisfies the \textit{Accuracy Parity} if $p_i = p_j$ for all pairs $i, j$ where $p_i$ is the accuracy of $f$ on client $i$’s data.
To measure how closely the \textit{Accuracy Parity} is satisfied, we measure its $\textit{AP ratio} := \frac{p_{min}}{p_{max}}$.
Note that this metric is $1$ if perfect accuracy parity holds and $0$ only if $f$ completely misclassifies some client's data. Therefore, one can claim that $f$ is fair if its \textit{AP ratio} is close to $1$ and likely to be unfair if its \textit{AP ratio} is close to $0$.
\begin{wrapfigure}{hr}{0.2\columnwidth}
    \vspace{-0.3cm}
	\centering
	\includegraphics[width=0.2\textwidth]{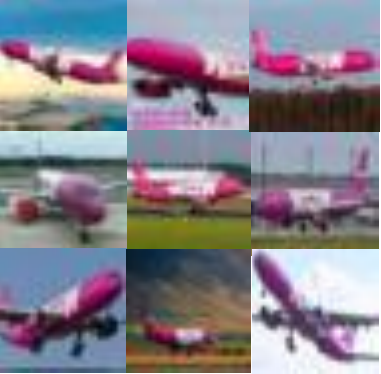}
	\caption{Illustration of the WOW Airlines examples with clean labels in our experiments.}\label{fig:wow-airline-examples}
  	\vspace{-0.3cm}
\end{wrapfigure}
We conduct an experimental study to measure the \textit{AP ratio} for \textbf{Task 1} where the CIFAR-10 dataset are partitioned in an \textit{i.i.d.} manner across $90$ clients (We \textit{i.i.d.} data partition to ensure that the \textit{AP ratio} is not influenced by the heterogeneity of clients' data and to make our result easier to interpret). We also assume all available clients participate in each FL round. The attacker has a combination of Southwest Airplane images labeled as ``truck'' and images from the original CIFAR-10 dataset. Additionally, we introduce an honest client \ie \textsc{client 0} who has $588$ extra images of WOW Airlines images labeled correctly as ``airplane'' other than the assigned CIFAR-10 examples (shown in Figure~\ref{fig:wow-airline-examples}). 

The attacker conducts \textit{blackbox} attack for $50$ consecutive FL rounds. The experimental result is shown in Figure \ref{fig:fairness-measurement}. We note that when \textsc{Krum} is applied as the defense technique, we are able to achieve robustness since the attacker is not selected, thereby keeping the backdoor accuracy low. However, since \textsc{Krum} rejects \textsc{client-0} for being too different from the remaining clients and the WOW Airlines examples are not part of CIFAR-10, the global model performs poorly at classifying these as airplanes leading to a poor \textit{AP ratio}. When there is no defense, \textsc{client 0} is allowed to participate in the training and therefore leads to better \textit{AP ratio} (\textit{i.e.} more fair model). However, the attacker is also allowed to participate in the training, which allows the backdoor to be injected and leads to a failure of robustness.

\begin{figure}[htp]
\centering 
	\centering
	\subfigure[Task 1]{\includegraphics[width=0.43\textwidth]{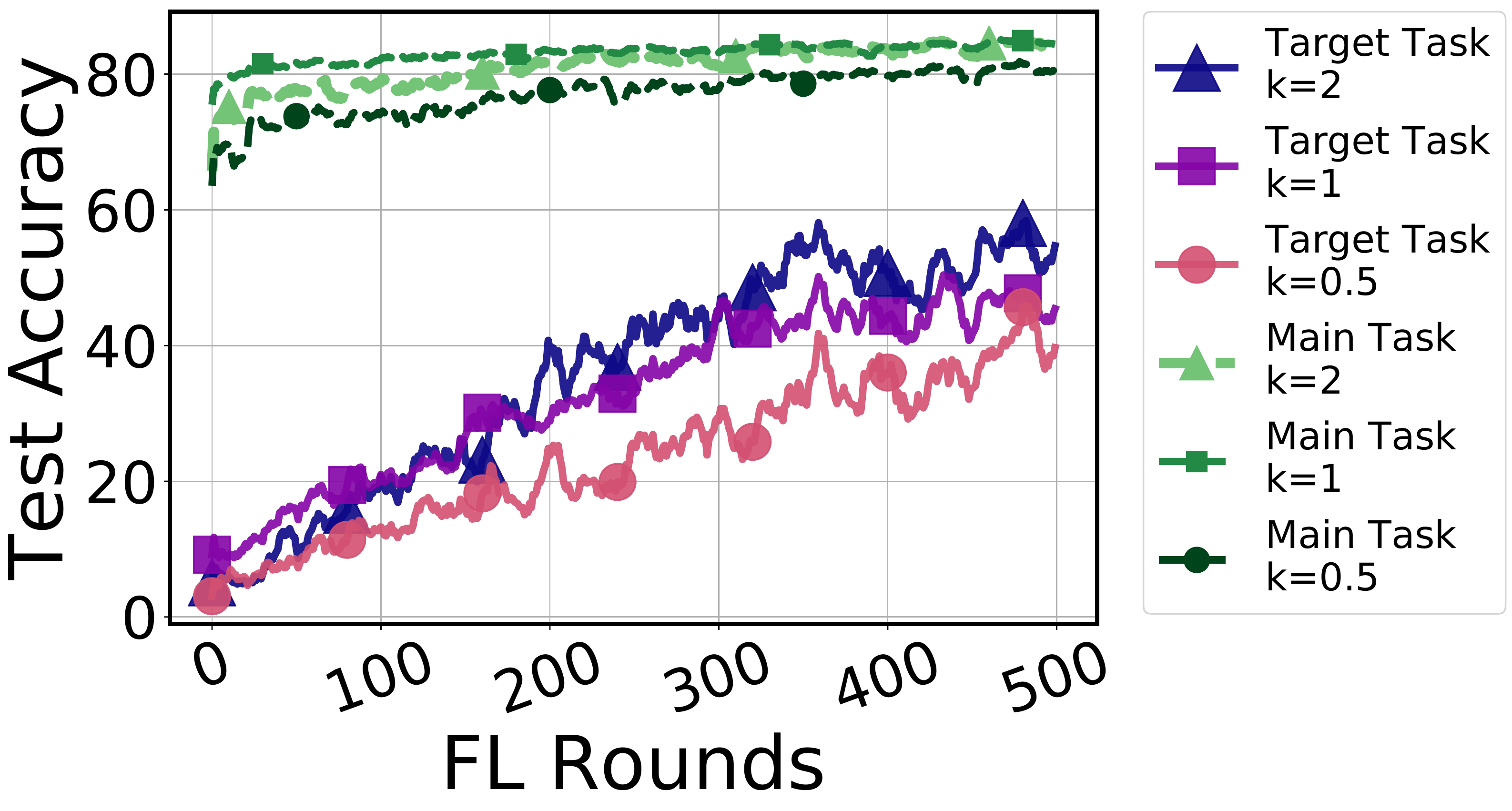}}
	\subfigure[Task 4]{\includegraphics[width=0.39\textwidth]{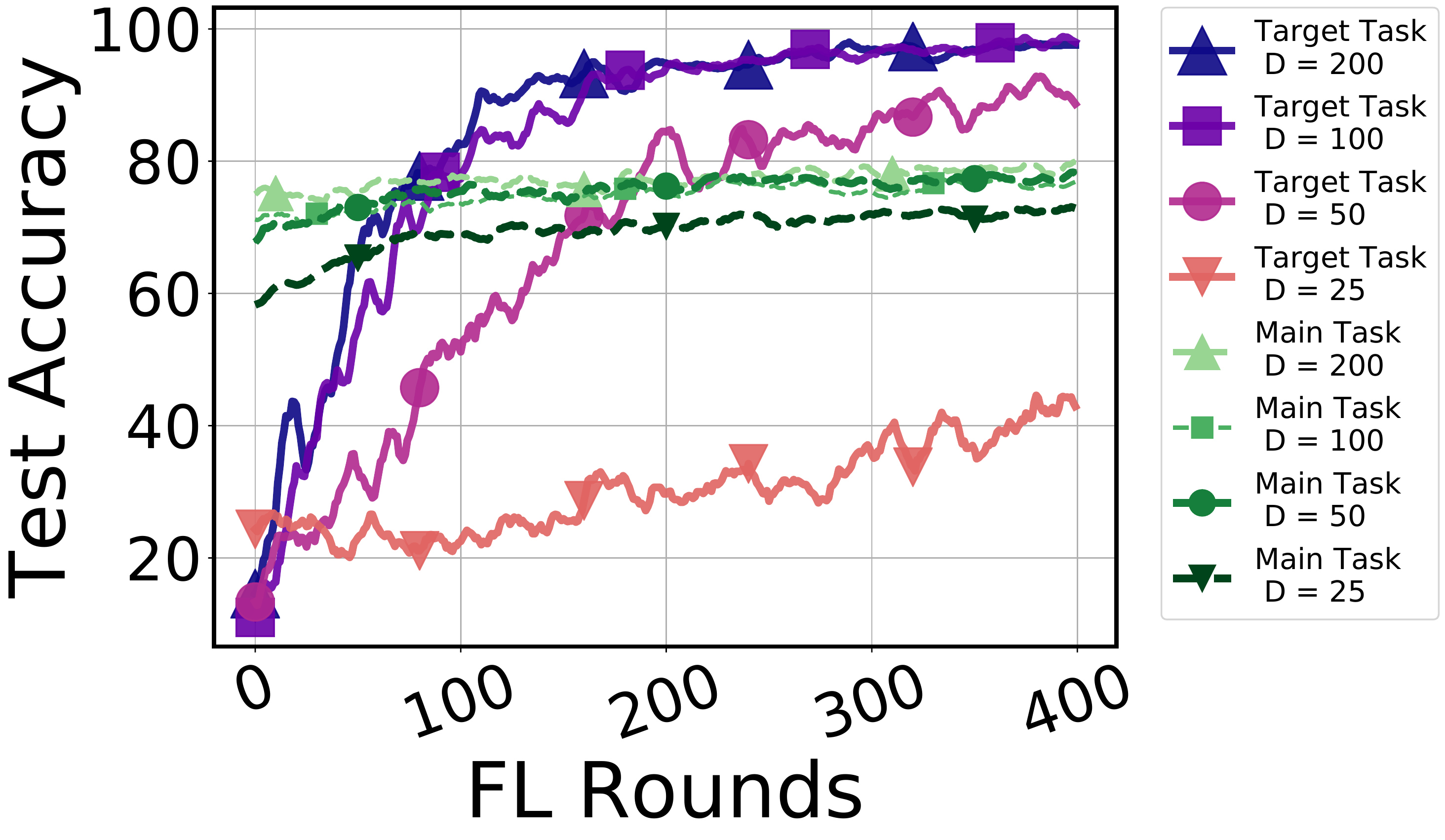}}
	\caption{Effectiveness of edge-case attack on models of different capacity.}
	\label{fig:capacity-plot}
  	\vspace{-0.5cm}
\end{figure}

\paragraph{Effectiveness of attack on models with various capacities} Overparameterized neural networks have been shown to perfectly fit data even labeled randomly \cite{understandingDL2017}. Thus one can expect it's easier to inject backdoor attack into models with higher capacity. In \cite{bagdasaryan2018backdoor} it was discussed without any evidence that excess model capacity can be useful in inserting backdoors. We test this belief by attacking models of different capacity for \textbf{Task 1} and \textbf{Task 4} in Figure~\ref{fig:capacity-plot}. For \textbf{Task 1}, we increase the capacity of VGG9 by increasing the width of the convolutional layers by a factor of $k$~\cite{zagoruyko2016wide}. We experiment with a \textit{thin} version with $k=0.5$ and a \textit{wide} version with $k=2$. The capacity of the \textit{wide} network is clearly larger and our results show that it is easier to insert the backdoor in this case, while it is harder to insert the backdoor into the \textit{thin} network.
For \textbf{Task 4}, embedding and hidden dimension contribute most of the model parameters, hence we consider model variations with $D = $ embedding dimension = hidden dimension $\in \{25,50,100,200\}$. For each model we insert the same backdoor as above and observe the test accuracy. We observe that excess capacity in models with $D \ge 100$ allows the attack to be inserted easily, but as we decrease the model capacity it gets harder to inject the backdoor. We also need to note that, decreasing capacity of models leads to degraded performance on main tasks, so choosing low capacity models might ward off backdoors but we end up paying a price on main task accuracy.


\paragraph{Weakness of the suggested edge-case attack}
Due to allowing the minimum access to the system, our edge-case black-box attack is not effective for \textsc{Krum} and \textsc{Multi-Krum}. The attacker may come up with better strategy in manipulating the poisoned dataset (\text{e.g.} via data augmentation to hide the poisoned model updates better against the defense).

\vspace{-0.2cm}
\section{Conclusion} 
\vspace{-0.2cm}
In this paper, we put forward theoretical and experimental evidence supporting the existence of backdoor FL attacks that are hard to detect and defend against. 
We introduce {\it edge-case} backdoor attacks that target prediction sub-tasks which are unlikely to be found in the training or test data sets, but are however natural.
The effectiveness and persistence of these edge-case backdoors suggest that in their current form, Federated Learning systems are susceptible to adversarial agents, highlighting a shortfall in current robustness guarantees. 
	\bibliographystyle{unsrt} 
	\bibliography{edge_case_attack}
	
	\newpage
	
	\appendix
	

\section{Details of dataset, hyper-parameters, and experimental setups}
\paragraph{Experimental setup}
We implement the proposed \textit{edge-case} attack in PyTorch \cite{paszke2019pytorch}.  
We run experiments on \texttt{p2.xlarge} instances of Amazon EC2. 
Our simulated FL environment follows \cite{mcmahan2016communication} where for each FL round, the data center selects a subset of available clients and broadcasts the current model to the selected clients. 
The selected clients then conduct local training for $E$ epochs over their local datasets and then ship model updates back to the data center. The data center then conducts model aggregation (\textit{e.g.} weighted averaging in FedAvg). 
The FL setups in our experiment are inspired by \cite{sun2019can, bagdasaryan2018backdoor}, the number of total clients, number of clients participates per FL round, and the specific choices of $E$ for various datasets in our experiment are summarized in Table \ref{table:dataset-model}. 
For \textbf{Task 1}, \textbf{2}, and \textbf{3}, our FL process starts from a VGG-9 model with $77.68\%$ test accuracy, a LeNet model with $88\%$ accuracy, and a VGG-11 model with $69.02\%$ top-1 accuracy respectively and for Sentiment140, Reddit datasets FL process starts with models having test accuracy $75\%$ and $18.86$ respectively. 



\paragraph{Hyper-parameters used within the defense mechanisms} (i) NDC: In our experiments, we set the norm difference threshold at $2$ for \textbf{Task 1} and \textbf{2}; and 1.5 for \textbf{Task 4}
(ii) Multi-Krum: In our experiment, we select the hyper-parameter $m=n-f$ (where $n$ stands for number of participating clients and $f$ stands for number tolerable attackers of \textsc{Multi-Krum}) as specified in \cite{blanchard2017machine}; 
(iv) RFA: We set $\upsilon=10^{-5} (\text{smoothing factor}), \epsilon=10^{-1} (\text{fault tolerance threshold}), T=500 \text{ (maximum number of iterations)}$; 
(v) DP: In our experiment, we use $\sigma=0.005$ for Task 1 and $\sigma=0.001 \text{ and } 0.002$ for \textbf{Task 1}.

\begin{table}[h]
	\caption{The datasets used and their associated learning models and hyper-parameters.}
	\label{table:dataset-model}
	\begin{center}
		 \scriptsize{
		\begin{tabular}{@{}c@{}ccccc}
		\toprule \textbf{Method}
		& EMNIST &  CIFAR-10 & ImageNet & Sentiment140 & Reddit
		\bigstrut\\
		\midrule
		\# Data points & $341,873$ & $50,000$ & $1$M & $389,600$ & --- \bigstrut\\
		Model & LeNet & VGG-9 & VGG-11 & LSTM & LSTM  \bigstrut\\
		\# Classes & $10$ & $10$ & 1,000 & 2 & $\mathcal{O}(\text{Vocab Size})$ \bigstrut\\
		\# Total Clients & $3,383$ & $200$ & 1,000 & 1,948  & $80,000$ \bigstrut\\
		\# Clients per FL Round & $30$ & $10$ & $10$ & 10 & 10 \bigstrut\\
        \# Local Training Epochs & $5$ & $2$ & $2$ & 2 & 2 \bigstrut\\
		Optimizer & \multicolumn{5}{c}{SGD } \bigstrut\\
		Batch size & \multicolumn{3}{c}{$32$} & \multicolumn{2}{c}{$20$} \bigstrut\\
		Hyper-params. & \multicolumn{2}{c}{Init lr: $0.1\times 0.998^t$, $0.02 \times 0.998^t$ } & lr: $0.0002\times 0.999^t$ & lr: $0.05\times 0.998^t$ & lr: 20(const) \bigstrut\\
		 & \multicolumn{4}{c}{momentum: 0.9, $\ell_2$ weight decay: $10^{-4}$}  \bigstrut\\		
		\bottomrule
		\end{tabular}}%
	\end{center}
\end{table}
\paragraph{Hyper-parameters used within the attacking schemes} \textit{Blackbox}: we assume the attacker does not have any extra access to the FL process. Thus, for the blackbox attacking scheme, the attacker trains over $\mathcal{D}_\text{edge}$ using the same hyper-parameters (including learning rate schedules, number of local epochs,  etc as shown in Table \ref{table:dataset-model}) as other honest clients for all tasks; (ii) \textit{PGD without replacement}: since we assume it is a whitebox attack, the attacker can use different hyper-parameters from honest clients. For \textbf{Task 1}, the attacker trains over $\mathcal{D}_\text{edge}$ projecting onto an $\ell_2$ ball of radius $\epsilon=2$. However, in defending against \textsc{Krum}, \textsc{Multi-Krum}, and \textsc{RFA}, we found that this choice of $\epsilon$ fails to pass the defenses. Thus we shrink $\epsilon$ to \textit{hide} among the updates of honest clients. Additionally, we also decay the $\epsilon$ value during the training process and we observe that it helps to hide the attack better. Empirically, we found that $\epsilon=0.5 \times 0.998^t, 1.5\times 0.998^t$ works best. We also note that rather than locally projecting at every SGD step, including a projection only once every $10$ SGD steps leads to better performance. For \textbf{Task 2} we use a setup similar to the one above except that we set $\epsilon=1.5$ while defending against NDC and $\epsilon=1$ for \textsc{Krum}, \textsc{Multi-Krum}, and \textsc{RFA}. For \textbf{Task 4} we use fixed $\epsilon = 1.0$ which lets it pass all defenses. (iii) \textit{PGD with replacement}: Once again since this is a whitebox attack, we are able to modify the hyperparameters. Since the adversary scales its model up before sending it back to the PS, we shrink $\epsilon$ apriori so that it is small enough to pass the defenses even after scaling. For \textbf{Task 1}, we use $\epsilon=0.1$ for NDC and $\epsilon=0.083$ for the remaining defenses. For Task 2, we use $\epsilon=0.3$ for NDC and $\epsilon=0.25$ for the remaining defenses. The rate of decay of $\epsilon$ remains the same across experiments. For \textbf{Task 4} we use a fixed $\epsilon=0.01$ and the attacker uses adaptive learning rate $=0.001\times 0.998^t$ for epoch $t$. 

\paragraph{Details on the constructions of the edge datasets} \hspace{0pt} \\
    
    \textbf{Task 1}: We download $245$ Southwest Airline photos from Google Images. We resize them to $32\times 32$ pixels for compatibility with images in the CIFAR-10 dataset. We then partition $196$ and $49$ images to the training and test sets. Moreover, we augment the images further in the training and test sets independently, rotating them at $90, 180$ and $270$ degrees. Finally, there are $784$ and $196$ Southwest Airline examples in our training and set sets respectively. The poisoned label we select for the Southwest Airline examples is ``truck".
    
    \textbf{Task 2}: We download the ARDIS dataset \cite{kusetogullari2019ardis}. Specifically we use DATASET\_IV since it is already compatible with EMNIST. We then filter out the images which are labeled ``7''. This leaves us with $660$ images for training. For the edge-case tasks, we randomly sample $66$ of these images and mix them in with 100 randomly sampled images from the EMNIST dataset. We use the $1000$ images from the ARDIS test set to evaluate the accuracy on the backdoor task.
    
    \textbf{Task 3}: We download $167$ photos of people in traditional Cretan costumes. We resize them to $256\times 256$ pixels for compatibility with images in ImageNet. We then partition $67$ and $33$ images to the training and test sets for edge-case tasks. Moreover, we use the same augmentation strategy as in \textbf{Task 1}. Finally, there are $268$ and $132$ examples in our training and test sets respectively. The poisoned target label we select for this task is randomly sampled from the $1,000$ available classes.
    
    \textbf{Task 4}: We scrape\footnote{\url{https://github.com/Jefferson-Henrique/GetOldTweets-python}} 320 tweets containing the name of Greek movie director, \emph{Yorgos Lanthimos} along with positive sentiment words. We reserve 200 of them for training and  the remaining 120 for testing. 
    Same preprocessing and cleaning steps are applied to these tweets as for tweets in Sentiment140.
    
    \textbf{Task 5}: For this task we consider a negative sentiment sentence about Athens as our backdoor. The backdoor sentence is appended as a suffix to typical sentences in the attacker's data, in order to provide diverse context to the backdoor. Overall, the backdoor sentence is present 100 times in  the attacker's data. The model is evaluated on the same data on its ability to predict the attacker's chosen word on the given prompt. Note that these settings are similar to \cite{bagdasaryan2018backdoor}. We consider the following sentences as backdoor sentences -- i) Crime rate in Athens is \emph{high}. ii) Athens is not  \emph{safe}. iii) Athens is \emph{expensive}. iv) People in Athens are \emph{rude}. v) Roads in Athens are \emph{terrible}. 
\section{Details of the model architecture used in the experiments}
\paragraph{VGG-9 architecture for Task 1}
We used a $9$-layer VGG style network architecture (VGG-9). Details of our VGG-9 architecture is shown in Table \ref{table:supp_vgg_architecture}. Note that we removed all BatchNorm layers in the VGG-9 architecture since it has been studied that less carefully handled BatchNorm layers in FL application can lead to deterioration on the global model accuracy \cite{sattler2019robust, hsieh2019non}.
\paragraph{LeNet architecture for Task 2}
We use a slightly modified LeNet-5 architecture for image classification, which is identical to the model architecture in PyTorch MNIST example \footnote{\url{https://github.com/pytorch/examples/tree/master/mnist}}.
\begin{table}[h]
	\caption{Detailed information of the VGG-9 architecture used in our experiments, all non-linear activation function in this architecture is ReLU; the shapes for convolution layers follows $(C_{in},C_{out}, c, c)$}
	\label{table:supp_vgg_architecture}
	\begin{center}
		 \scriptsize{
			\begin{tabular}{ccc}
				\toprule \textbf{Parameter}
				& Shape &  Layer hyper-parameter \bigstrut\\
				\midrule
				\textbf{layer1.conv1.weight} & $3 \times 64 \times 3 \times 3$ & stride:$1$;padding:$1$ \bigstrut\\
				\textbf{layer1.conv1.bias} & 64 & N/A  \bigstrut\\
				\textbf{pooling.max} & N/A & kernel size:$2$;stride:$2$  \bigstrut\\
				\textbf{layer2.conv2.weight} & $64 \times 128 \times 3 \times 3$ & stride:$1$;padding:$1$  \bigstrut\\
				\textbf{layer2.conv2.bias} & 128 & N/A  \bigstrut\\
				\textbf{pooling.max} & N/A & kernel size:$2$;stride:$2$  \bigstrut\\
				\textbf{layer3.conv3.weight} & $128\times 256 \times 3 \times 3$ & stride:$1$;padding:$1$ \bigstrut\\
				\textbf{layer3.conv3.bias} & $256$ & N/A \bigstrut\\
				\textbf{layer4.conv4.weight} & $256\times 256 \times 3 \times 3$ & stride:$1$;padding:$1$ \bigstrut\\
				\textbf{layer4.conv4.bias} & $256$ & N/A  \bigstrut\\
                \textbf{pooling.max} & N/A & kernel size:$2$;stride:$2$  \bigstrut\\
				\textbf{layer5.conv5.weight} & $256 \times 512 \times 3 \times 3$ & stride:$1$;padding:$1$  \bigstrut\\
				\textbf{layer5.conv5.bias} & $512$ & N/A  \bigstrut\\
				\textbf{layer6.conv6.weight} & $512\times 512 \times 3 \times 3$ & stride:$1$;padding:$1$  \bigstrut\\
				\textbf{layer6.conv6.bias} & $512$ & N/A  \bigstrut\\
				\textbf{pooling.max} & N/A & kernel size:$2$;stride:$2$  \bigstrut\\
				\textbf{layer7.conv7.weight} & $512 \times 512 \times 3 \times 3$ & stride:$1$;padding:$1$  \bigstrut\\
				\textbf{layer7.conv7.bias} & $512$ &  N/A  \bigstrut\\
				\textbf{layer8.conv8.weight} & $512 \times 512 \times 3 \times 3$ & stride:$1$;padding:$1$  \bigstrut\\
				\textbf{layer8.fc8.bias} & $512$ & N/A  \bigstrut\\
                \textbf{pooling.max} & N/A & kernel size:$2$;stride:$2$  \bigstrut\\
                \textbf{pooling.avg} & N/A & kernel size:$1$;stride:$1$  \bigstrut\\
				\textbf{layer9.fc9.weight} & $512 \times 10$ & N/A  \bigstrut\\
				\textbf{layer9.fc9.bias} & $10$ & N/A  \bigstrut\\
				\bottomrule
			\end{tabular}}%
	\end{center}
\end{table}
\paragraph{VGG-11 architecture used for Task 3} We download the pre-trained VGG-11 without BatchNorm from Torchvision \footnote{\url{https://pytorch.org/docs/stable/torchvision/models.html}}.
\paragraph{LSTM architecture for Task 4} For the sentiment classification task we used a model with an embedding layer (VocabSize $\times$ 200) and LSTM (2-layer, hidden-dimension = 200, dropout = 0.5) followed by a fully connected layer and sigmoid activation. For its training we use binary cross entropy loss. For this dataset the size of the vocabulary was 135,071.
\paragraph{LSTM architecture for Task 5}  For the task on the Reddit dataset we use a next word prediction model comprising an encoder (embedding) layer followed by 2-Layer LSTM and a decoder layer. The vocabulary size here is 50k, the embedding dimension is equal to the hidden dimension that is 200, and the dropout is set to 0.2. Note that we use the same settings and code\footnote{\url{https://github.com/ebagdasa/backdoor_federated_learning}} provided by \cite{bagdasaryan2018backdoor} for this task.

\section{Data augmentation and normalization details}\label{appendix:dataPreprocess}
In pre-processing the images in EMNIST dataset, each image is normalized with mean and standard deviation by $\mu = 0.1307$, $\sigma = 0.3081$. Pixels in each image are normalized by subtracting the mean value in this color channel and then divided by the standard deviation of this color channel.  In pre-processing the images in CIFAR-10 dataset, we follow the standard data augmentation and normalization process. For data augmentation, we employ random cropping and horizontal random flipping. Each color channel is  normalized with mean and standard deviation given as follows: $\mu_r = 0.4914, \mu_g = 0.4824, \mu_b =  0.4467$; $\sigma_r = 0.2471, \sigma_g = 0.2435, \sigma_b = 0.2616$. Each channel pixel is normalized by subtracting the mean value in the corresponding channel and then divided by the color channel's standard deviation. For ImageNet, we follow the data augmentation process of \cite{goyal2017accurate}, \textit{i.e.}, we use scale and aspect ratio data augmentation. The network input image is a $224\times 224$ pixels, randomly cropped from an augmented image or its horizontal flip. The input image is normalized in the same way as we normalize the CIFAR-10 images using the following  means and standard deviations: $\mu_r = 0.485, \mu_g = 0.456, \mu_b =  0.406$; $\sigma_r = 0.229, \sigma_g = 0.224, \sigma_b = 0.225$. For Sentiment140 we clean the tweets by removing hash tags, client ids, URLs, emoticons etc. Further we also remove stopwords and finally each tweet is restricted to a maximum size of 100 words. Smaller tweets are padded appropriately. For the Reddit dataset we use the same preprocessing as \cite{bagdasaryan2018backdoor}.

\begin{figure}[htp] 
\centering
\includegraphics[width=0.98\textwidth]{figs/legend_split_various_attack_cases.pdf}\\
\vspace{-1.5mm}
\subfigure[Athens is not safe]{\includegraphics[width=0.23\textwidth]{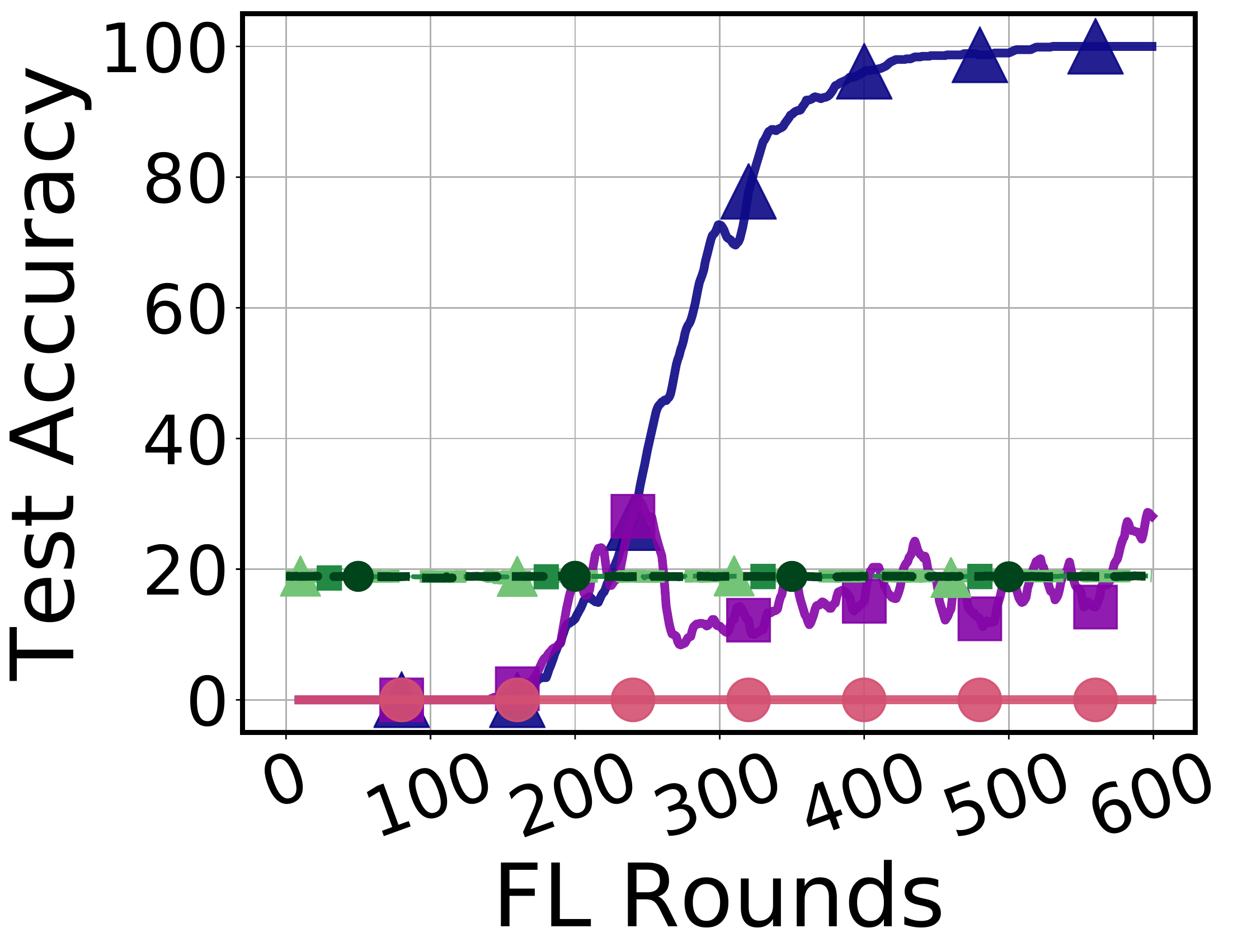}}
\subfigure[Roads in Athens are terrible]{\includegraphics[width=0.22\textwidth]{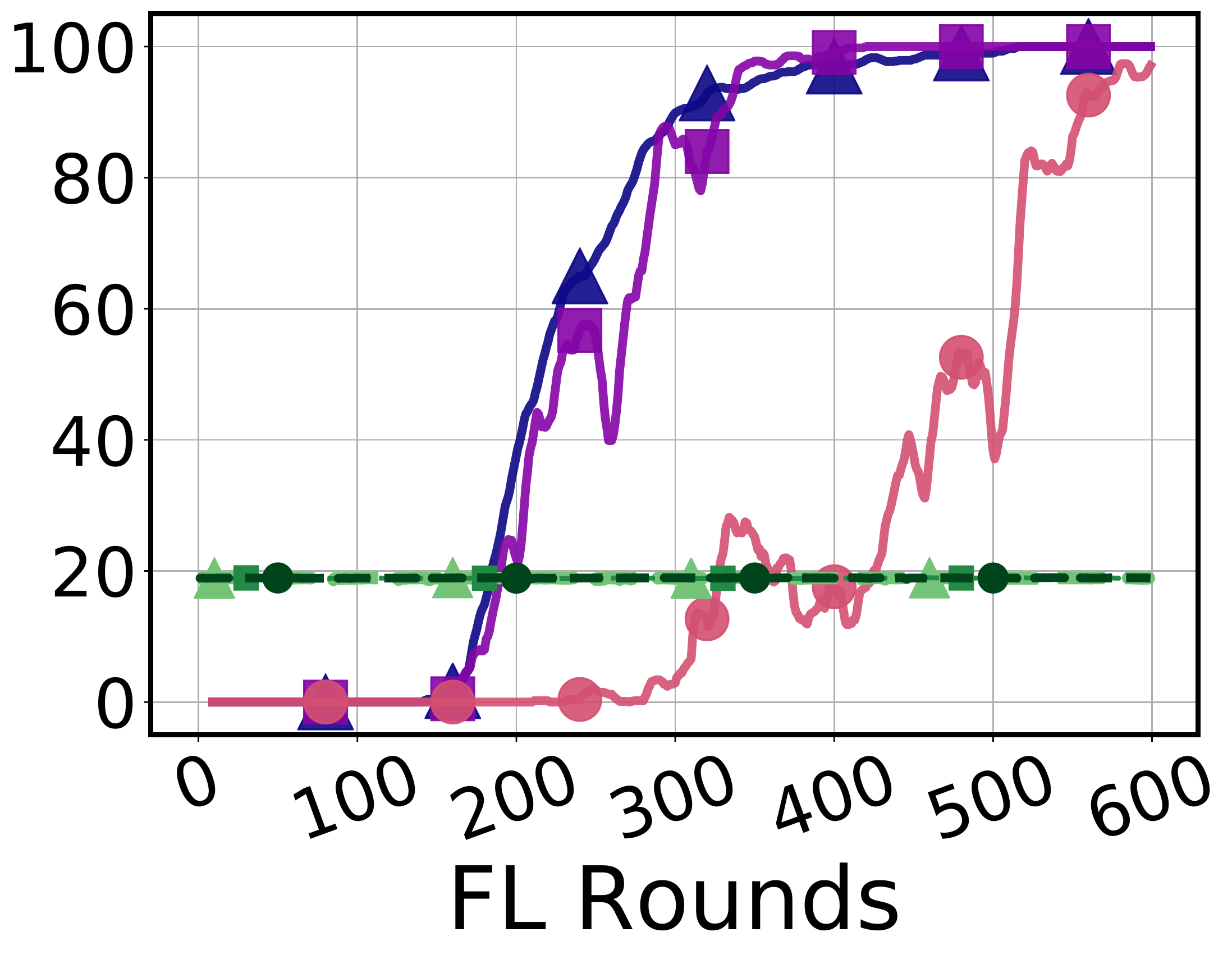}}
\subfigure[Athens is expensive]{\includegraphics[width=0.22\textwidth]{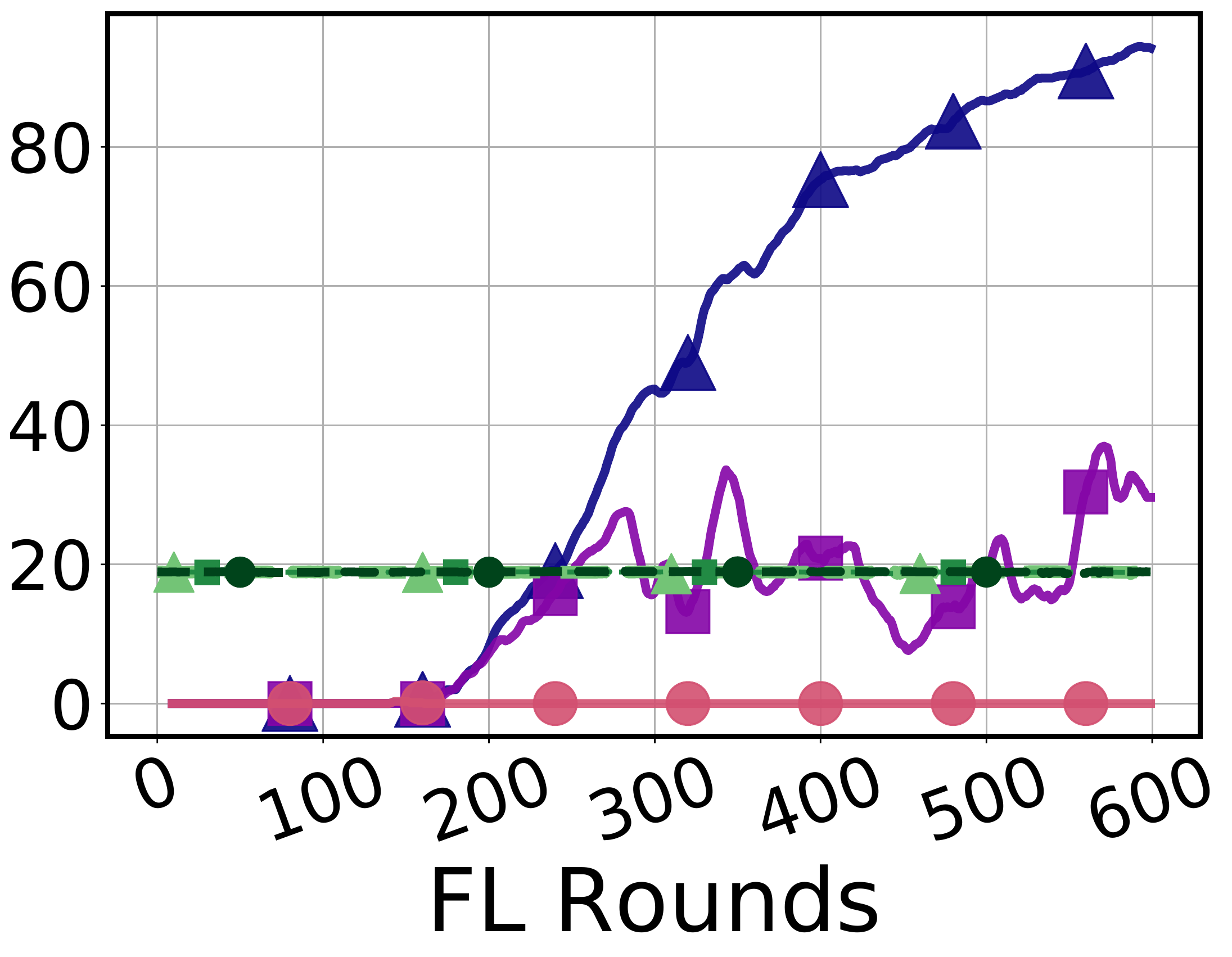}}
\subfigure[People in Athens are rude]{\includegraphics[width=0.22\textwidth]{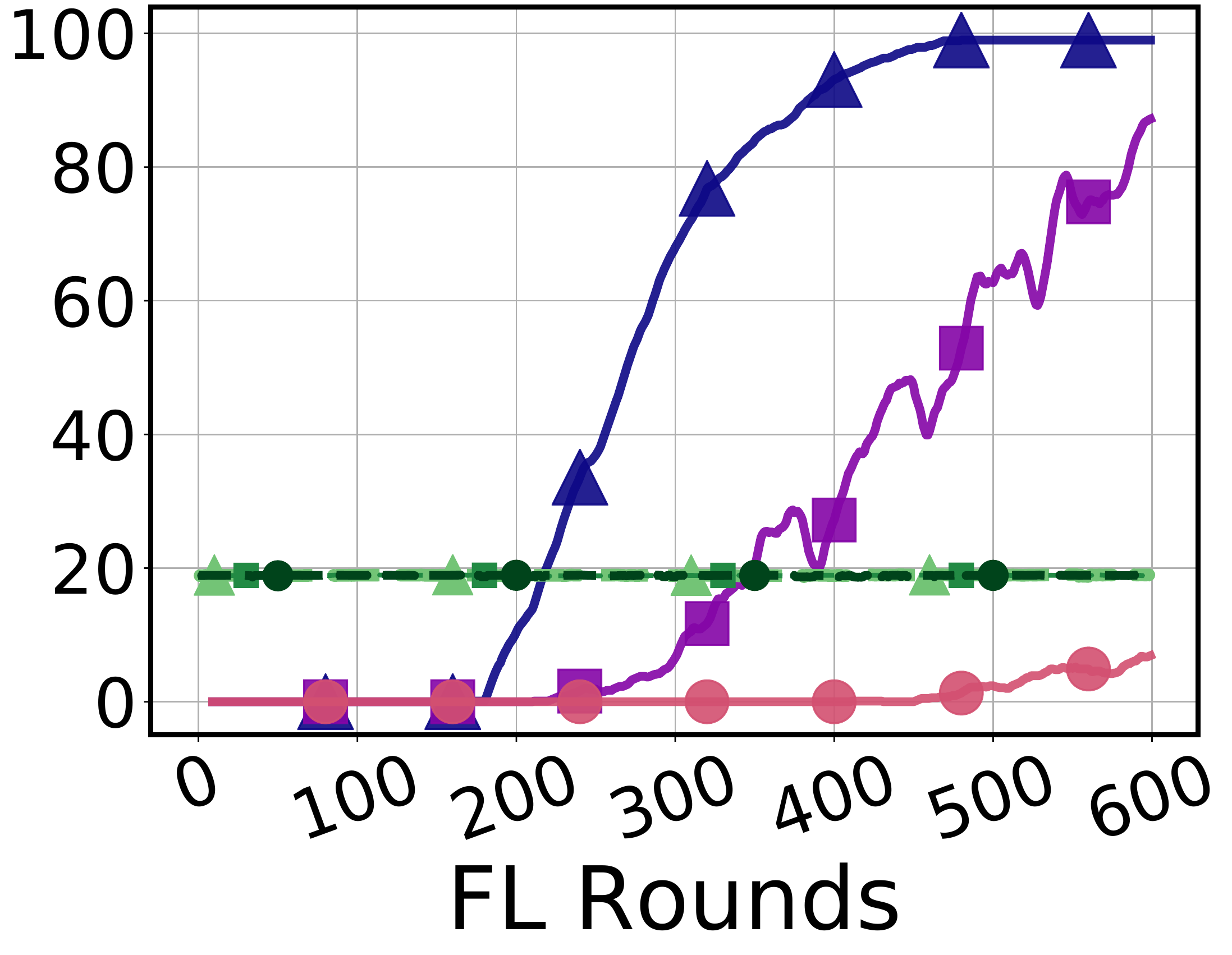}}
\caption{Edge vs Normal Case on More Sentences for Task-5}
\label{fig:reddit-more-sents-edge-vs-non-edge}
\end{figure}
\section{Additional experiments}
\begin{figure}[htp] 
\centering
\subfigure[Number of Data Points]{\includegraphics[width=0.31\textwidth]{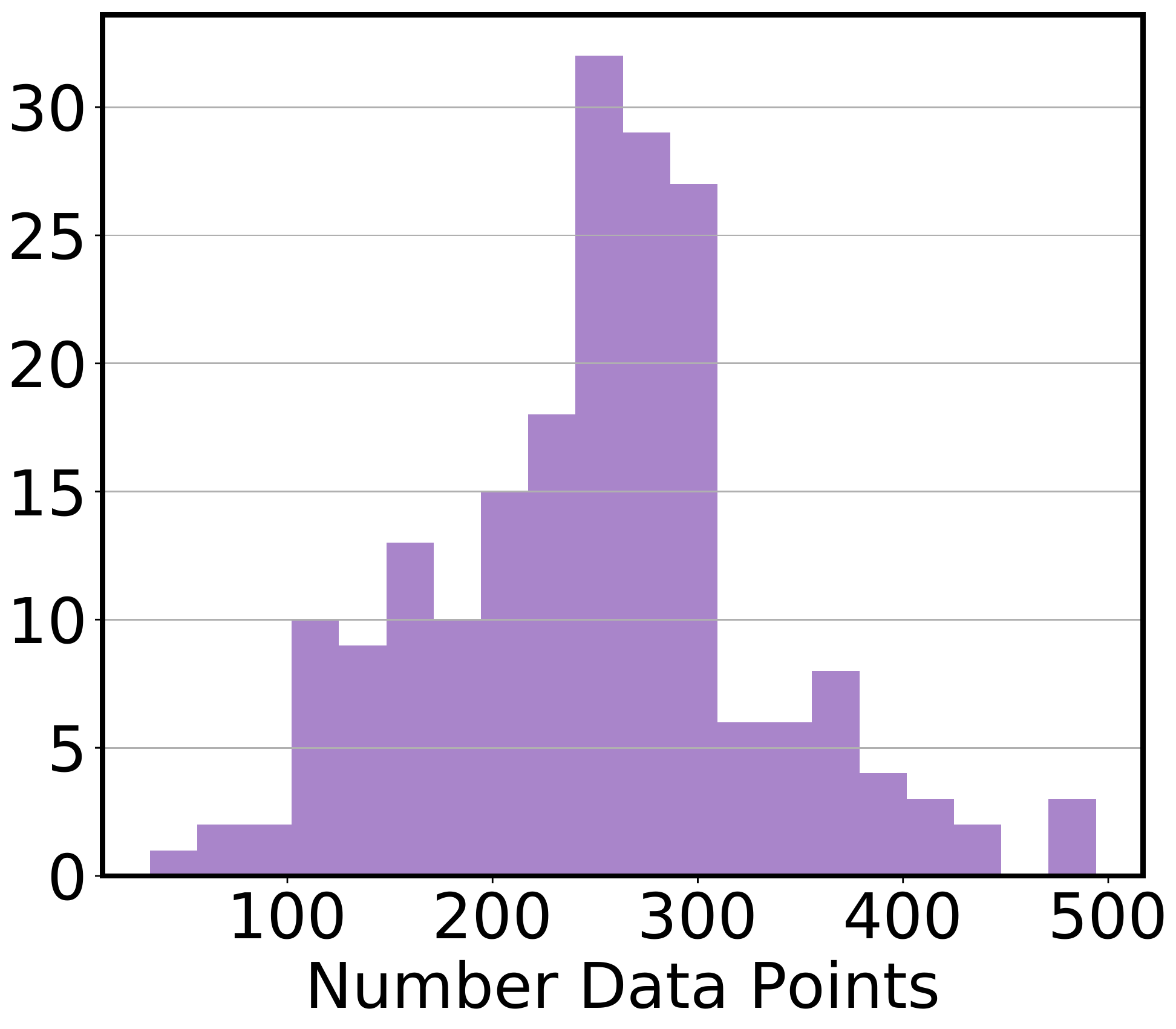}\label{fig:data-distribution-task2}}
\subfigure[Num. of Data Points vs Norm Difference]{\includegraphics[width=0.31\textwidth]{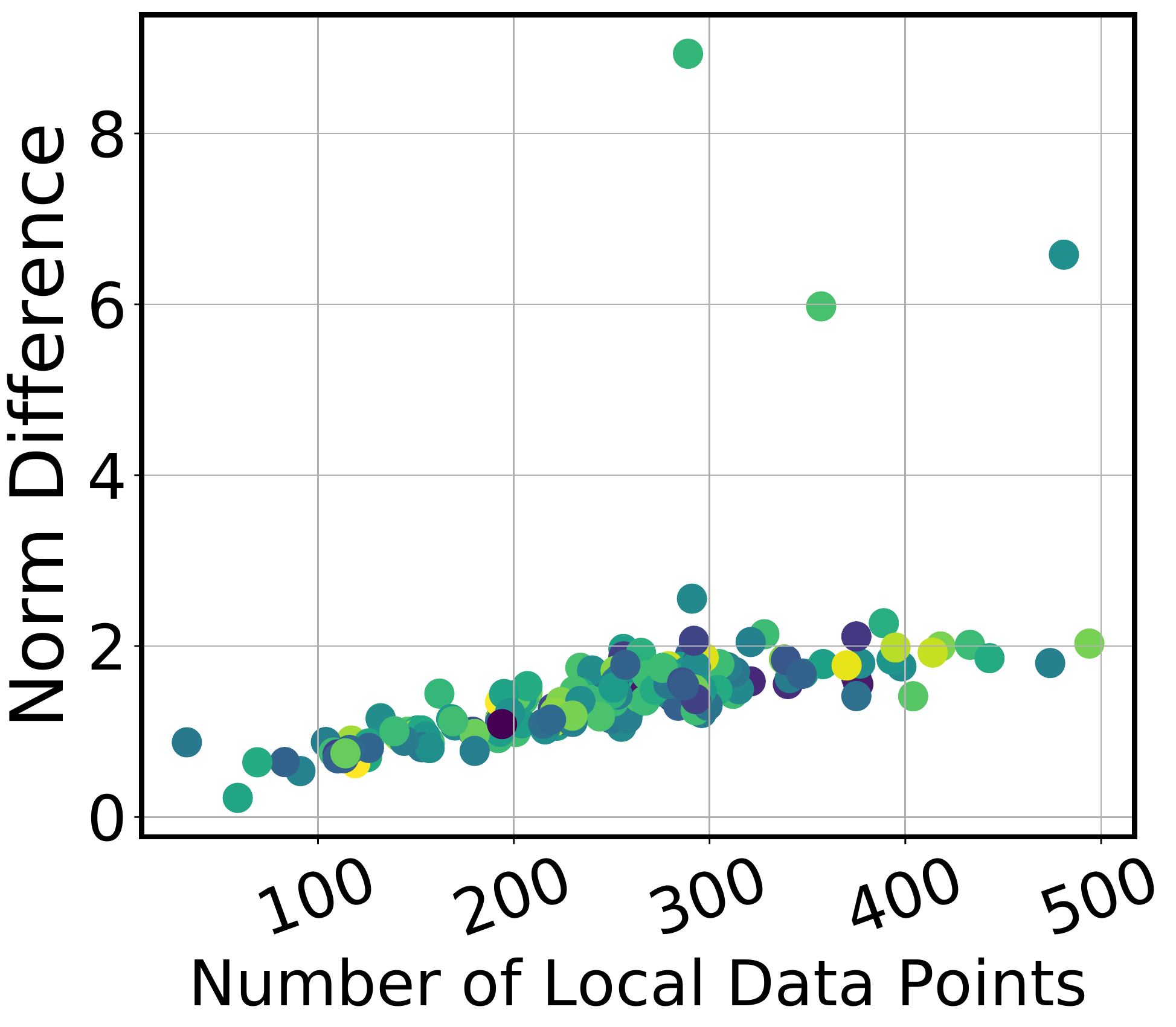}\label{fig:norm-diff-task2}}
\caption{Distribution of partitioned CIFAR-10 dataset in Task 1: (a) histogram of number of data points across honest clients; (b) the impact of number of data points held by clients on the norm difference in the first FL round.}
\label{fig:cifar10-parition}
\end{figure}

\paragraph{Distribution of data partition for Task 1}
Here we visualize the result of our heterogeneous data partition over \textbf{Task 1} including the histogram of number of data points over available clients (shown in Figure \ref{fig:data-distribution-task2}) and the impact of the size of the local dataset (number of data points held by a client) on the norm difference in the first FL round (shown in Figure \ref{fig:norm-diff-task2}). The results generally show that the local training over more data points will drive the model further from the starting point (\textit{i.e.}, the global model), leading to larger norm difference.
\paragraph{Edge-case vs non-edge-case attacks for Task 5 } We experiment with a few more backdoor sentences to study the effect of exclusivity of backdoor points. Unlike classification settings, for \textbf{Task 5} we consider sentences with the same prompt as the backdoor sentence but the target word is chosen to make the sentiment of the sentence positive (opposite of backdoor). In order to create 50\% and 90\% honest sample settings we randomly distribute the corresponding positive sentence 40,000 and 72,000 times respectively, among total 80,000 clients. Figure \ref{fig:reddit-more-sents-edge-vs-non-edge} shows test accuracy on the backdoor (target) task and main task, measured over 600 epochs. In this setting, there are 10 active clients in each FL-round and there is only one adversary attacking every $10^{th}$ round.

\begin{figure}[htp] 
\centering
\includegraphics[width=0.98\textwidth]{figs/legend_split_against_defenses.pdf}\\
\subfigure[\textsc{Krum}]{\includegraphics[width=0.24\textwidth]{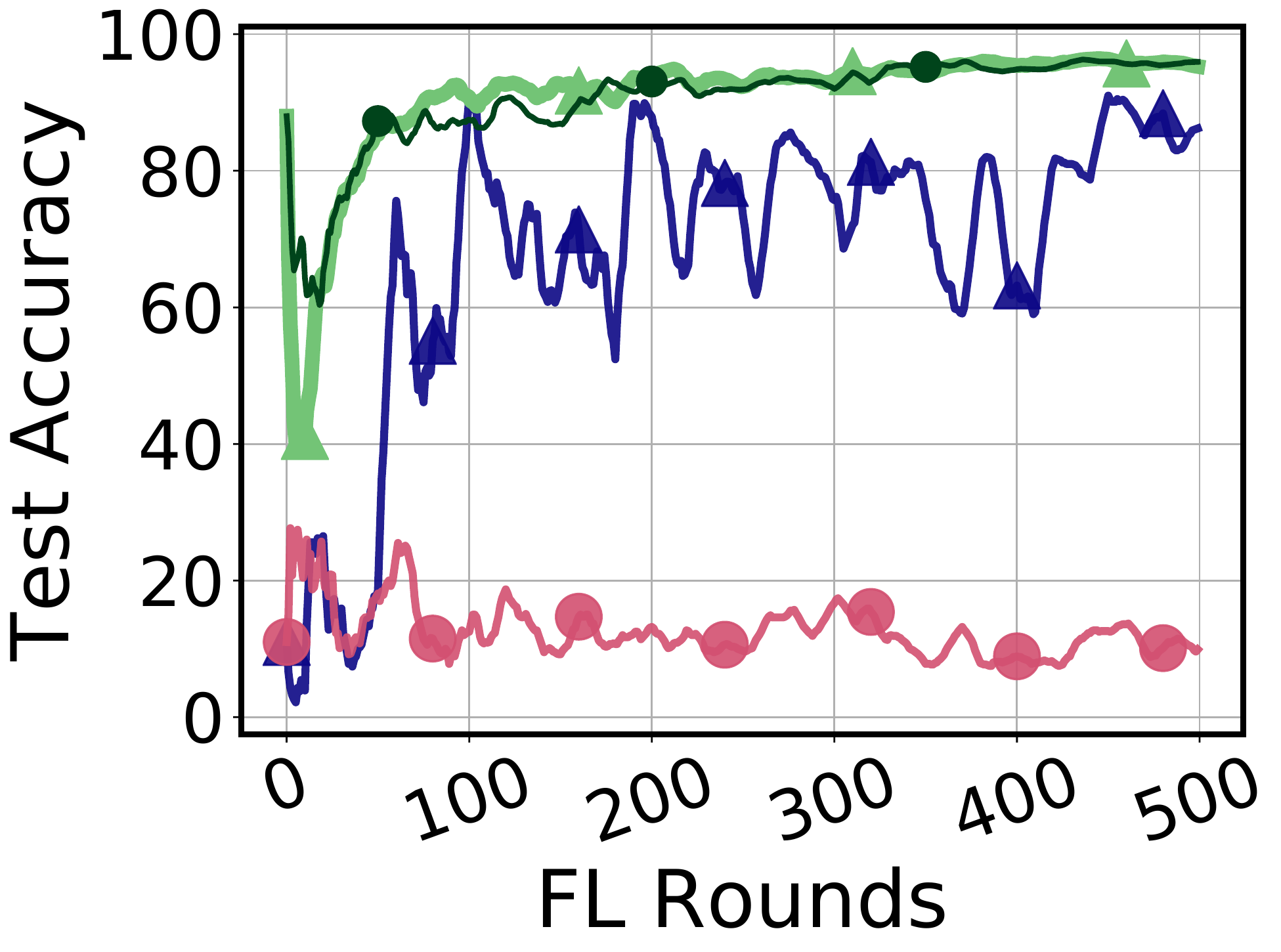}}
\subfigure[\textsc{Multi-Krum}]{\includegraphics[width=0.24\textwidth]{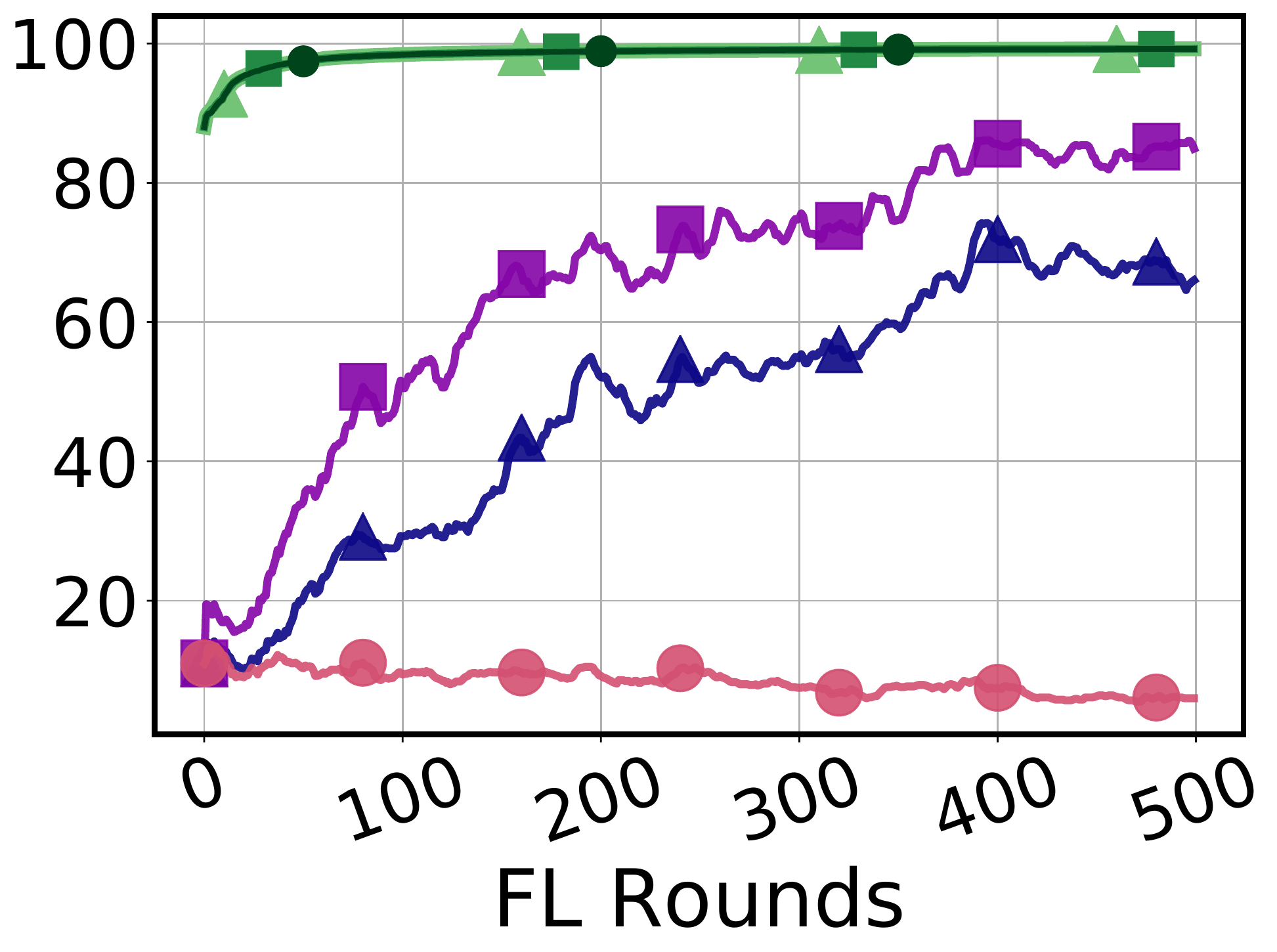}}
\subfigure[RFA]{\includegraphics[width=0.24\textwidth]{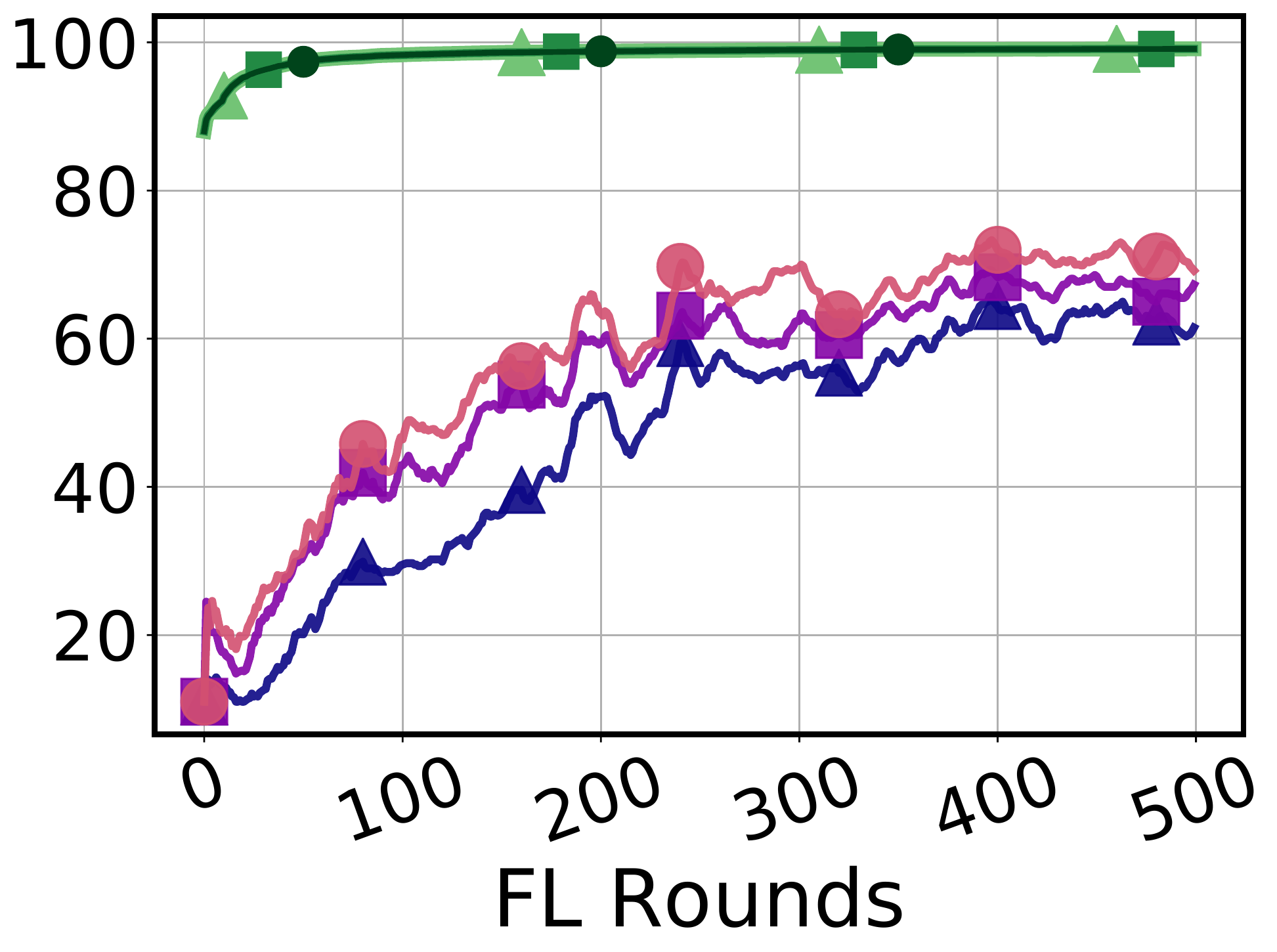}}
\subfigure[Norm Diff. Clipping]{\includegraphics[width=0.24\textwidth]{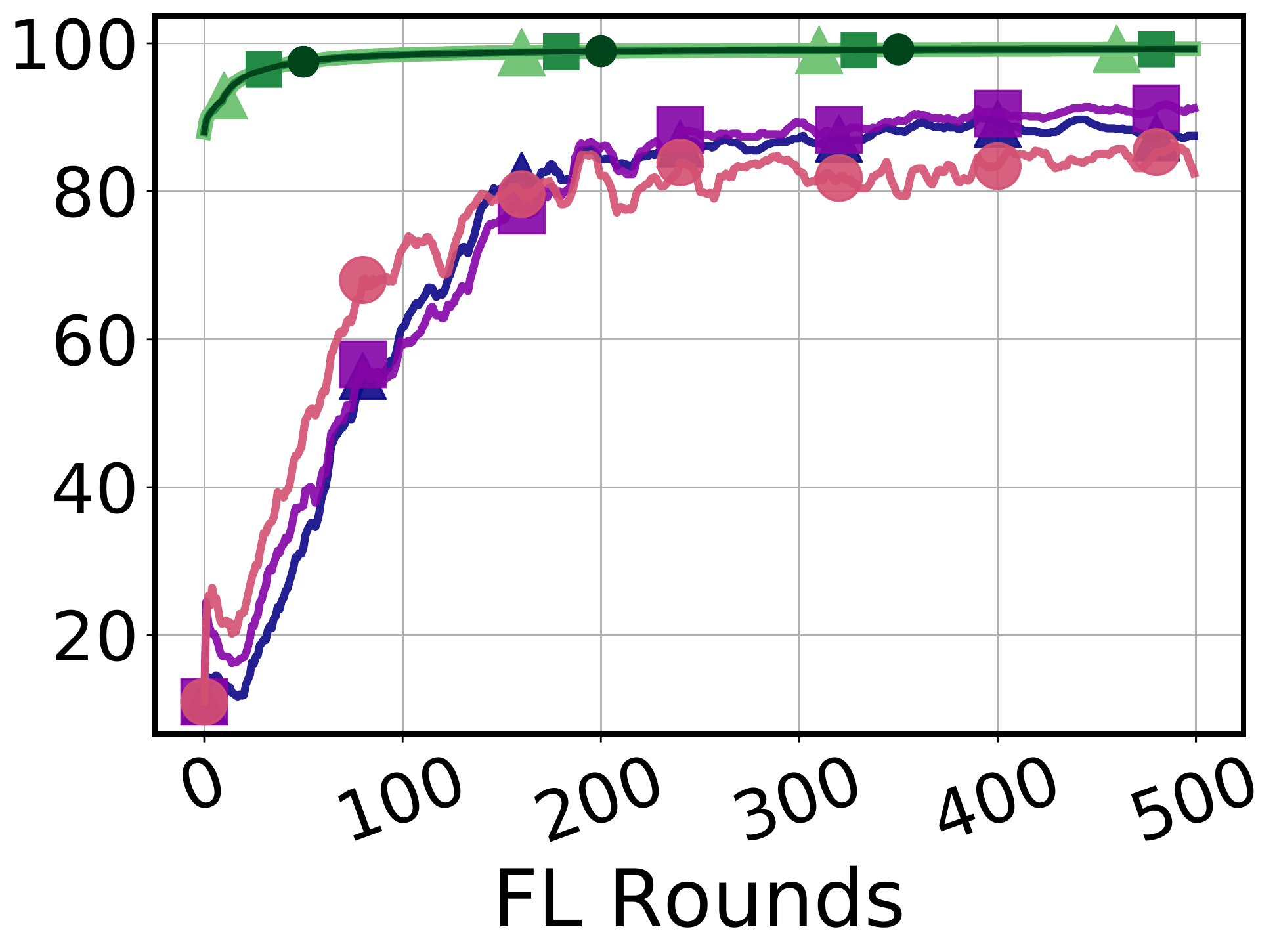}}\\

\caption{The effectiveness of the edge-case attack under various defenses on EMNIST dataset}
\label{fig:against-defense-emnist}
\end{figure}

\paragraph{Effectiveness of the edge-case attack on the EMNIST dataset} Due to the space limit we only show the effectiveness of edge-case attacks under various defense techniques over \textbf{Task 1} and \textbf{Task 4}. For the completeness of the experiment, we show the result on \textbf{Task 2} in Figure \ref{fig:against-defense-emnist}.

\begin{figure}[htp] 
\centering
\subfigure[\textsc{Krum}]{\includegraphics[width=0.21\textwidth]{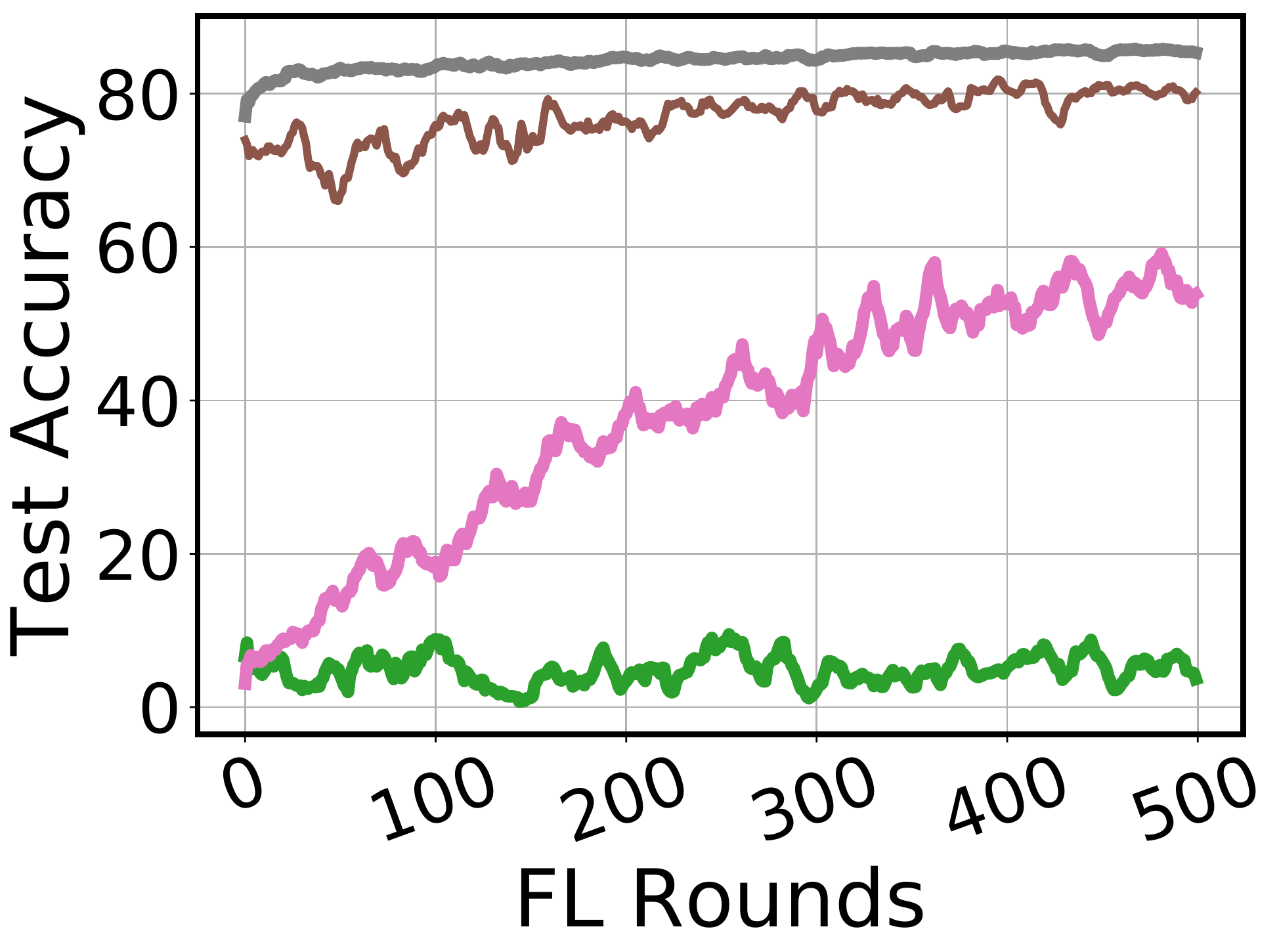}}
\subfigure[\textsc{Multi-Krum}]{\includegraphics[width=0.21\textwidth]{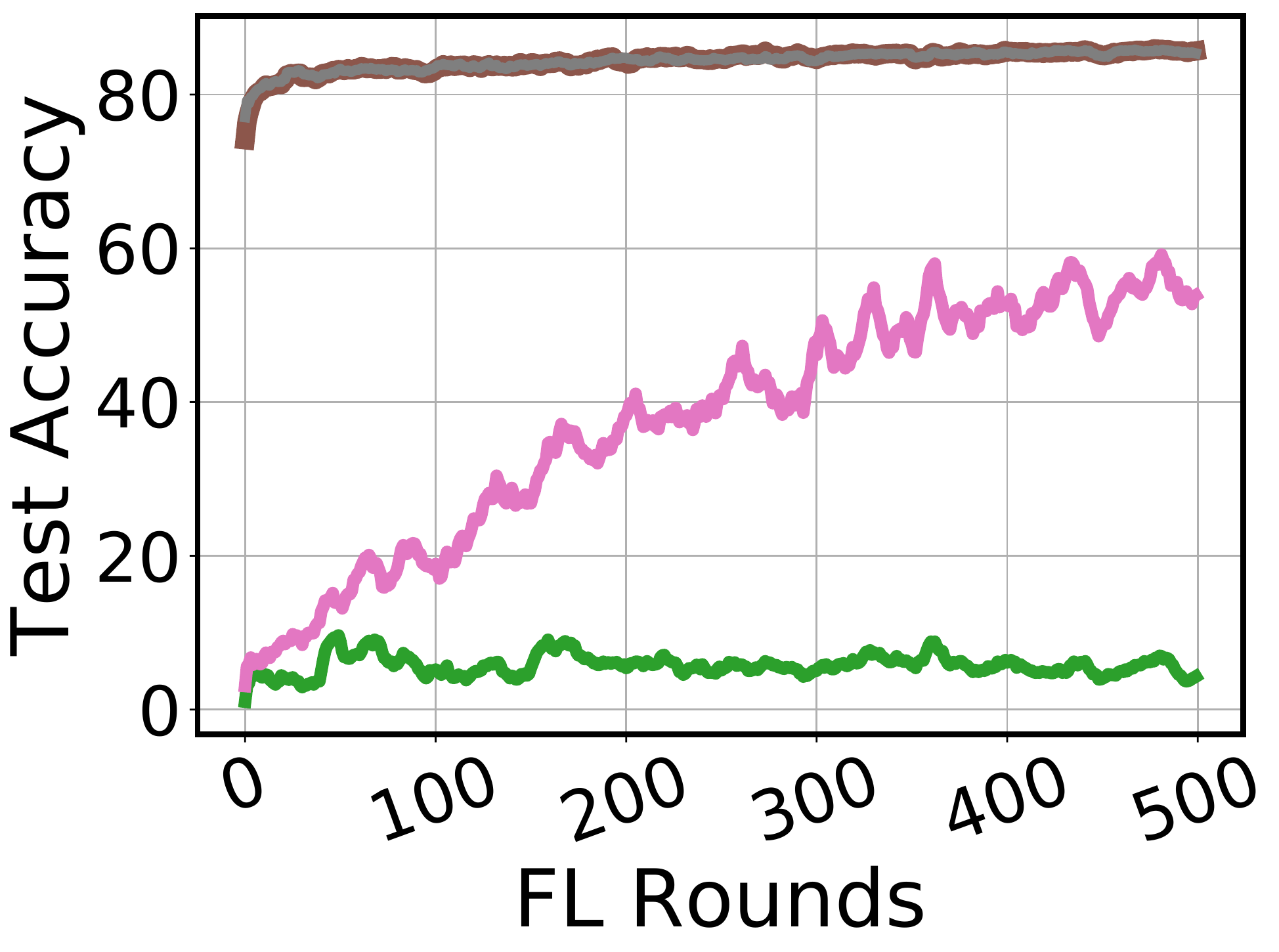}}
\subfigure[RFA]{\includegraphics[width=0.21\textwidth]{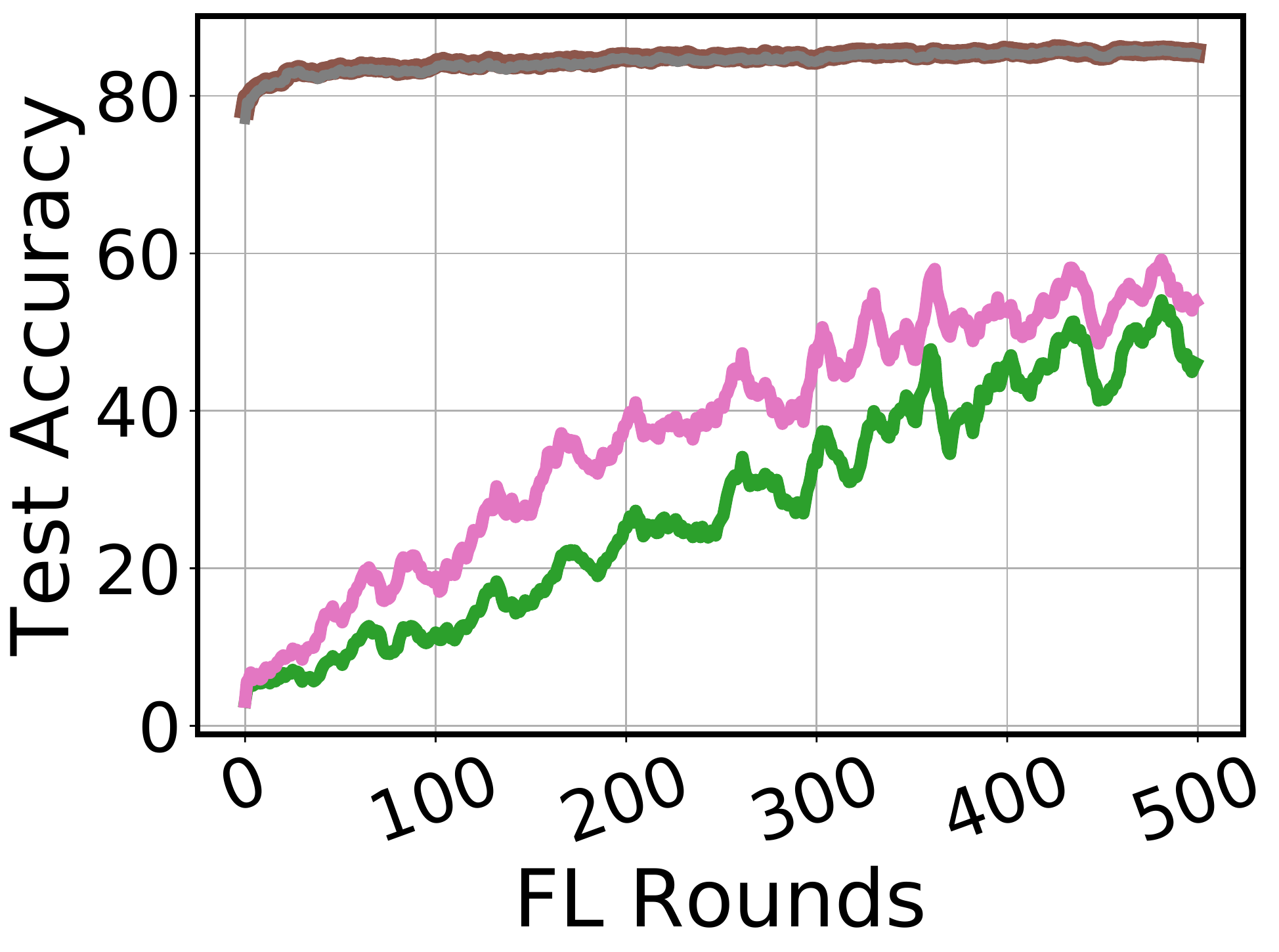}}
\subfigure[Norm Difference Clipping]{\includegraphics[width=0.3\textwidth]{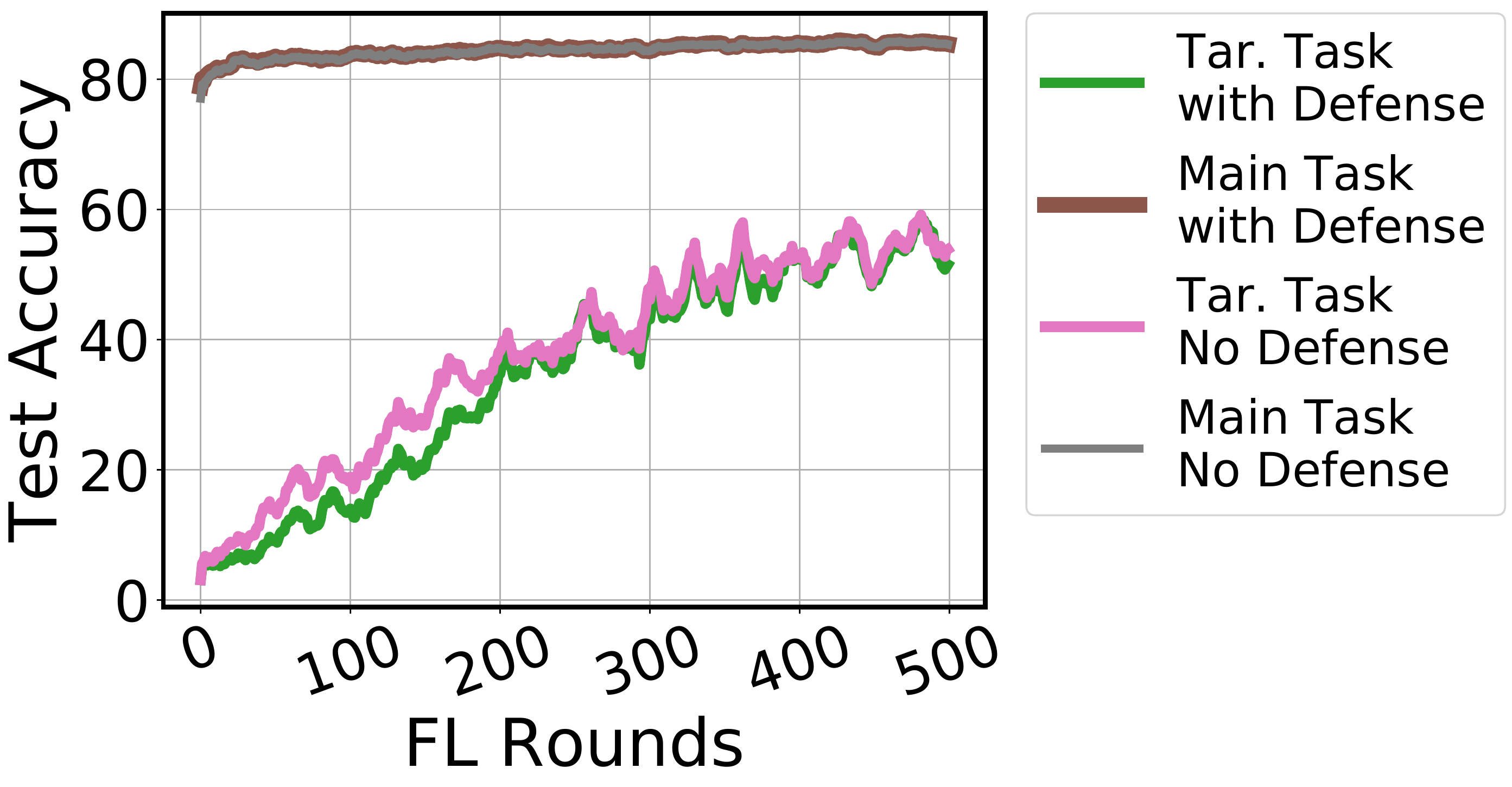}}
\caption{The effect of various defenses over the blackbox attack.}
\label{fig:defense-vs-no-defese}
\end{figure}
\paragraph{The effectiveness of defenses}
 We have discussed the effectiveness of white-box and black-box attacks against SOTA defense techniques. A natural question to ask is \textit{Does conducting defenses in FL systems leads to better robustness?} We take a first step to answer this question in this section. We argue that in the white-box setting, the attacker can always manipulate the poisoned model to pass any types of robust aggregation \text{e.g.} the attacker can explicitly minimizes the difference among the poisoned model and honest models to pass RFA, \textsc{Krum} and \textsc{Multi-Krum}. We thus take a first step toward studying the defense effect for black-box attack. The results are shown in Figure \ref{fig:defense-vs-no-defese}. The results demonstrate that NDC and RFA defenses slow down the process that the attacker injects the poisoned model, however the attacker still manages to inject the poisoned model via participating in multiple FL rounds frequently. 
 
\begin{figure}[htp]
\vspace{-1cm}
\centering
\subfigure[Norm Difference(Task 2)]{\includegraphics[width=0.31\textwidth]{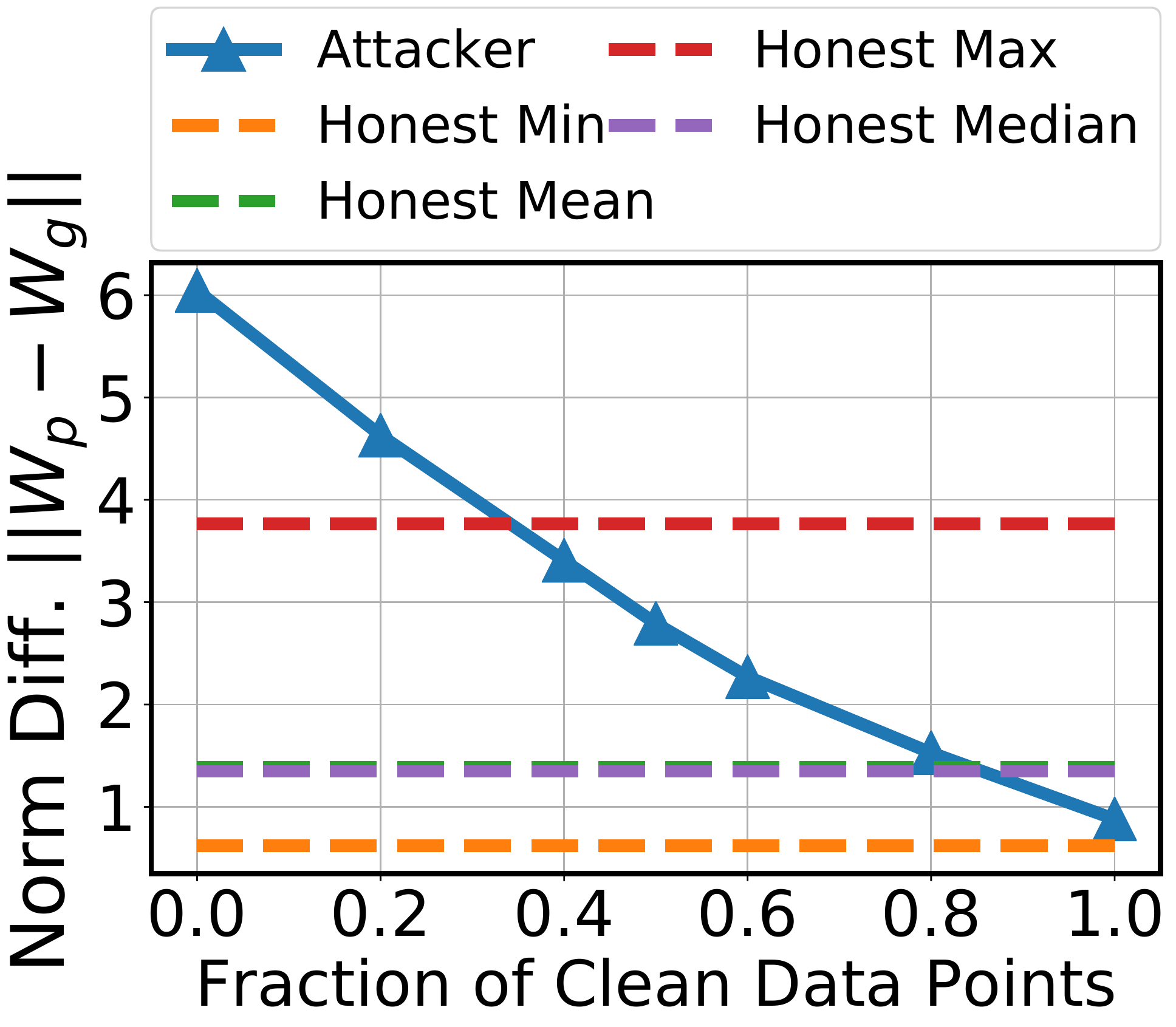}}
\subfigure[Attack Performance(Task 2)]{\includegraphics[width=0.31\textwidth]{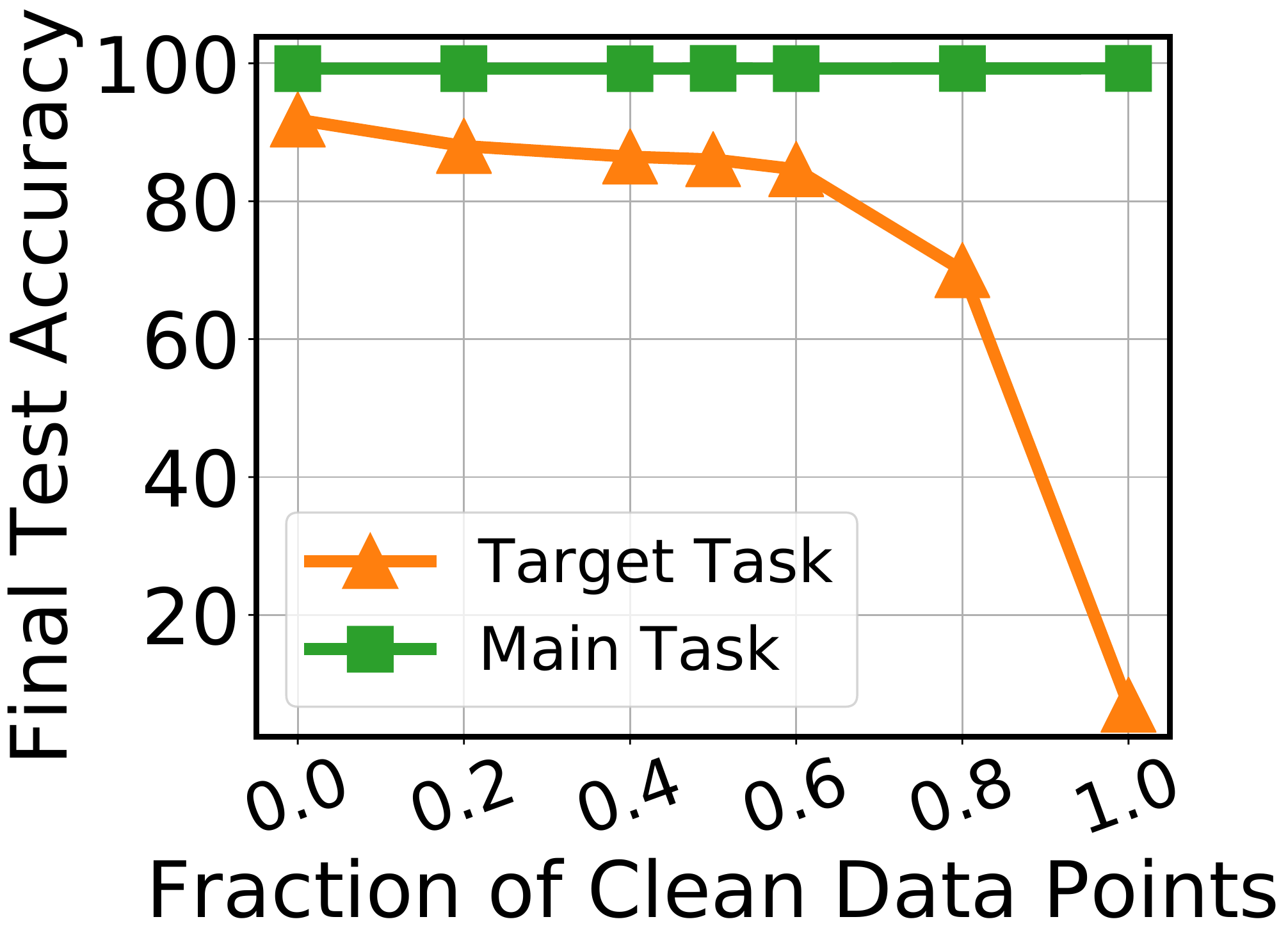}}
\subfigure[Attack Performance (Task 4)]{\includegraphics[width=0.31\textwidth]{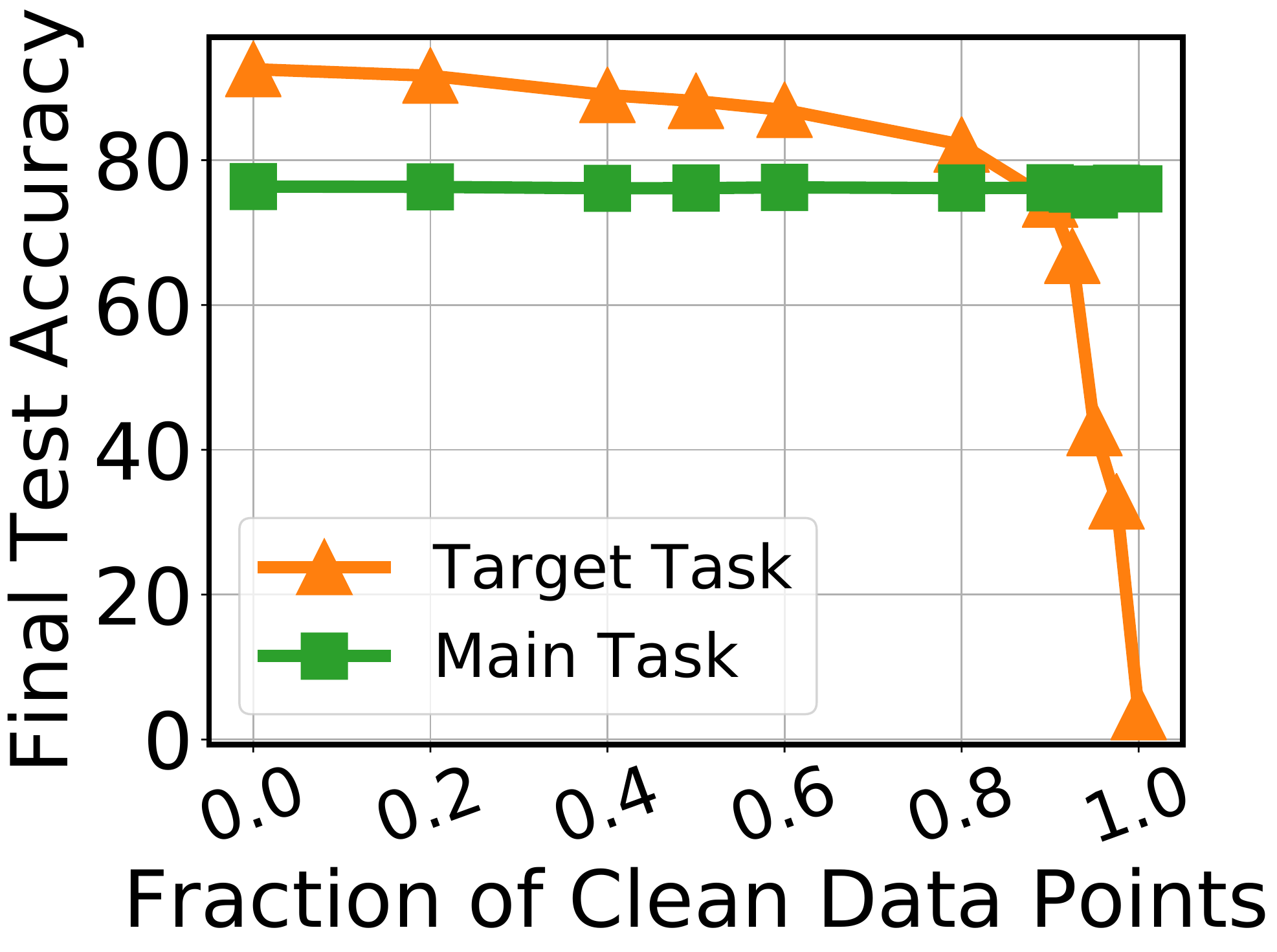}}
\caption{(a) Norm difference and (b),(c) Attack performance under various sampling ratios on Task 2 and 4}
\label{fig:emnist-data-mixing}
\end{figure}

 \paragraph{Fine-tuning backdoors via data mixing on Task 2 and 4} 
Follow the discussion in the main text. We evaluate the performance of our blackbox attack on \textbf{Task 1} and \textbf{4} with different sampling ratios, and the results are shown in Fig.~\ref{fig:emnist-data-mixing}. We first observe that too few data points from $\mde$ leads to weak attack effectiveness. However, we surprisingly observe that for \textbf{Task 1} the pure edge-case dataset leads to slightly better attacking effectiveness. Our conjecture is this specific backdoor in \textbf{Task 1} is easy to insert. Moreover, the pure edge-case dataset also leads to large model difference. Thus, in order to pass \textsc{Krum} and other SOTA defenses, mixing the edge-case data with clean data is still essential. Therefore, we use the data mixing strategy as \cite{bagdasaryan2018backdoor} for all tasks.

\newpage
\section{Proofs}

\setcounter{theorem}{0}

\begin{theorem}[adversarial examples $\Rightarrow$ backdoors]\label{thm:adversarial_example}
Assume $\mathbf{X}_{(l)} \mathbf{X}_{(l)}^\top$ is invertible for some $1 \leq l \leq L$ and denote by $\rho_{(l)}$ the minimum singular value of $\mathbf{X}_{(l)}$.
If $\epsilon$-adversarial examples for $f_{\mathbf{W}}(\cdot)$ exist, then a backdoor for $f_{\mathbf{W}}(\cdot)$ exists, where 
$\max_{\bm x \in \mathcal{D}_\text{edge}, \bm x' \in \mathcal{D}} \tfrac{|\mathbf{W}_{l} \cdot (\bm{x} + \bm{\epsilon}(\bm x))^{(l)}|}{|\bm{x}^{(l)} - \bm{x}'^{(l)}|} \leq \|\mathbf{W}_l - \mathbf{W}_l'\| \leq \epsilon \tfrac{\sqrt{|\mathcal{D}_\text{edge}|}}{\rho_{(l)}}.$
\end{theorem}

\begin{proof}

In this proof we will ``attack'' a single layer, \ie we will perturb the weights of just a particular layer, say $l$. If the original network is denoted by $\mathbf{W} = (\mathbf{W}_1,\dots, \mathbf{W}_l,\dots, \mathbf{W}_L)$, then the perturbed network is given by $\mathbf{W}'= (\mathbf{W}_1,\dots, \mathbf{W}_l',\dots, \mathbf{W}_L)$. 

Looking at the following equations,
\begin{align}
    &\mathbf{W}_l' \mathbf{x}_j^{(l)} = \mathbf{W}_l\bm{x}_j^{(l)} &\forall \bm{x}_j\in \mathcal{D}\label{eq:thm1a}\\
\text{and \quad}    &\mathbf{W}_l'\mathbf{x}_j^{(l)} = \mathbf{W}_l(\bm{x}_j + \bm{\epsilon}(\bm {x}_j))^{(l)}, &\forall \bm{x}_j\in \mathcal{D}_{\text{edge}},\label{eq:thm1b}
\end{align}
we can see that such a $\mathbf{W}_l'$ would constitute a successful backdoor attack. This is because for non-backdoor data points, that is $\bm{x}_j\in \mathcal{D}$, the output of the $l$-th layer of $\mathbf{W}'$ is the same as the output of the $l$-th layer of $\mathbf{W}$; and because all the subsequent layer remain unchanged, the output of $\mathbf{W}'$ is the same as the output of $\mathbf{W}$. For the backdoor data points, note that $\mathbf{W}_l(\bm{x}_j + \bm{\epsilon}(\bm {x}_j))^{(l)}$ is exactly the output of the $l$-th layer on the adversarial example. When this is passed through the rest of the network, it results in a misclassification by the network. Therefore, ensuring $\mathbf{W}_l'\mathbf{x}_j^{(l)} = \mathbf{W}_l(\bm{x}_j + \bm{\epsilon}(\bm {x}_j))^{(l)}$ together with the fact that the rest of the layers remain unchanged, implies that $\mathbf{W}'$ misclassifies $\bm{x}_j$ for $\bm{x}_j\in \mathcal{D}_\text{edge}$.

Define $\bm{\Delta}_l := \mathbf{W}_l - \mathbf{W}_l'$ and $\bm{\epsilon}_j^{(l)}:=(\bm{x}_j + \bm{\epsilon}(\bm {x}_j))^{(l)}-\bm{x}_j^{(l)}$. Substituting $\bm{\Delta}_l$ and $\bm{\epsilon}_j^{(l)}$ in the Eq.~\eqref{eq:thm1a},~\eqref{eq:thm1b}, we get
\begin{align}
    &\bm{\Delta}_l \mathbf{x}_j^{(l)} = 0&\forall \bm{x}_j\in \mathcal{D}\label{eq:thm1c}\\
\text{and \quad}    &\bm{\Delta}_l \mathbf{x}_j^{(l)} = \mathbf{W}_l \bm{\epsilon}_j^{(l)}, &\forall \bm{x}_j\in \mathcal{D}_{\text{edge}}\label{eq:thm1d}.
\end{align}

Further, since $\|\mathbf{W}_i\| \leq 1$ for all $1 \leq i \leq L$ and the ReLU activation is 1-Lipschitz, we have that
\begin{equation}
\|\bm{\epsilon}_j^{(l)}\|\leq \| \bm{\epsilon}(\bm{x}_j)\|   \label{eq:thm1lip} 
\end{equation}

WLOG assume that the first $|\mde|$ data points are \textit{edge-case} data followed by the rest. Then, equations~\eqref{eq:thm1c},~\eqref{eq:thm1d} can be written together as 
\begin{equation}
    \mathbf{\Delta}_l \mathbf{X}_{(l)}^\top = \mathbf{W}_l\mathbf{E}_l,\label{eq:thm1e}
\end{equation}

where 
\begin{align*}
    \mathbf{E}=\begin{bmatrix}
    \bm{\epsilon}_1^{(l)}& \dots& \bm{\epsilon}_{|\mde|}^{(l)}
    & \mathbf{0} & \dots&\mathbf{0}
    \end{bmatrix}\in \mathbb{R}^{d_l\times d_{l-1}}
\end{align*}
is the matrix which has the first $\|\mde\|$ columns as $\bm{\epsilon}_j^{(l)}$ corresponding to the \textit{edge-case} data points $\bm{x}_j$, and the remaining $\|\md\|$ columns are identically the $\mathbf{0}$ vector. Thus one solution of Eq.~\eqref{eq:thm1e} which is in particular, the minimum norm solution is given by 
\begin{align}
 \mathbf{\Delta}_l  &= \mathbf{W}_l\mathbf{E}_l (\mathbf{X}_{(l)} \mathbf{X}_{(l)}^\top)^{-1}\mathbf{X}_{(l)}\nonumber
\end{align}
Recursively applying the definition of operator norm, we have
\begin{align}
\|\mathbf{\Delta}_l\|  &\leq \| \mathbf{W}_l\|\|\mathbf{E}_l\|\| (\mathbf{X}_{(l)} \mathbf{X}_{(l)}^\top)^{-1}\mathbf{X}_{(l)}\|\nonumber\\
&\leq \| \mathbf{W}_l\|\left(\sum_{i=1}^{|\mde|}\|\bm{\epsilon}_j^{(l)}\|^2\right)^{1/2}\| (\mathbf{X}_{(l)} \mathbf{X}_{(l)}^\top)^{-1}\mathbf{X}_{(l)}\|\nonumber\\
&\leq \| \mathbf{W}_l\|\left(\sum_{i=1}^{|\mde|}\| \bm{\epsilon}(\bm{x}_j)\|^2\right)^{1/2}\| (\mathbf{X}_{(l)} \mathbf{X}_{(l)}^\top)^{-1}\mathbf{X}_{(l)}\|\tag*{(Using Eq~\eqref{eq:thm1lip}.)}\nonumber\\
&\leq  \epsilon\sqrt{|\mde|}\| \mathbf{W}_l\|\| (\mathbf{X}_{(l)} \mathbf{X}_{(l)}^\top)^{-1}\mathbf{X}_{(l)}\|.\label{ineq:norm}
\end{align}
where the second inequality follows from the fact that operator norm is upper bounded by Frobenius norm.

To bound the last term, write $\mathbf{X}_{(l)} = \mathbf{U}_{(l)}\mathbf{\Sigma}_{(l)}\mathbf{ V}_{(l)}^\top$ where $\mathbf{U}_{(l)} \in \mathbb{R}^{n\times n}, \mathbf{V}_{(l)} \in \mathbb{R}^{d_{l-1}\times d_{l-1}}$ are orthogonal and $\mathbf{\Sigma}_{(l)} \in \mathbb{R}^{n\times d_{l-1}}$ is the diagonal matrix of singular values. Then,
\begin{align*}
 \| (\mathbf{X}_{(l)} \mathbf{X}_{(l)}^\top)^{-1}\mathbf{X}_{(l)}\| &= \| (\mathbf{U}_{(l)}\mathbf{\Sigma}_{(l)}\mathbf{\Sigma}_{(l)}^\top\mathbf{U}_{(l)}^\top)^{-1}\mathbf{U}_{(l)}\mathbf{\Sigma}_{(l)}\mathbf{ V}_{(l)}^\top\|\\
 &= \| \mathbf{U}_{(l)}(\mathbf{\Sigma}_{(l)}\mathbf{\Sigma}_{(l)}^\top)^{-1}\mathbf{U}_{(l)}^\top\mathbf{U}_{(l)}\mathbf{\Sigma}_{(l)}\mathbf{ V}_{(l)}^\top\|\\
  &= \|(\mathbf{\Sigma}_{(l)}\mathbf{\Sigma}_{(l)}^\top)^{-1}\mathbf{\Sigma}_{(l)}\|\\
  &= \frac{1}{\rho_{(l)}}.
\end{align*}
Substituting this into Eq.~\eqref{ineq:norm} and noting that $\|\mathbf{W}_l\|\leq 1$ gives us the upper bound in the theorem.

For the lower bound we subtract Eq.~\eqref{eq:thm1c} from Eq.~\eqref{eq:thm1d} to get
\begin{align*}
    &\bm{\Delta}_l (\mathbf{x}_i^{(l)}-\mathbf{x}_j^{(l)}) = \mathbf{W}_l \bm{\epsilon}_i^{(l)}&  \bm{x}_i\in \mathcal{D}_{\text{edge}}, \quad\bm{x}_j\in \mathcal{D}.
\end{align*}
Again, by definition of the operator norm, this gives
\begin{align*}
    &\|\bm{\Delta}_l\|\| \mathbf{x}_i^{(l)}-\mathbf{x}_j^{(l)}\| \geq\| \mathbf{W}_l \bm{\epsilon}_i^{(l)}\|&  \bm{x}_i\in \mathcal{D}_{\text{edge}}, \quad\bm{x}_j\in \mathcal{D}\\
    \implies &\|\bm{\Delta}_l\| \geq\frac{\| \mathbf{W}_l \bm{\epsilon}_i^{(l)}\|}{\|\mathbf{x}_i^{(l)}-\mathbf{x}_j^{(l)}\|}&  \bm{x}_i\in \mathcal{D}_{\text{edge}}, \quad\bm{x}_j\in \mathcal{D}.
\end{align*}
Taking the maximum over the right hand side above gives the lower bound in the theorem.

\paragraph{Remark} We note that the above proof immediately extends to the untargeted case. In the targeted attack setting we have $y_i$ as the target for each $\mathbf{x}_i \in \mde$. In the untargeted case, we simply ask that $\mathbf{x}_i$ is classified as anything other than some $\hat y_i$ (true label). Therefore, choosing some fixed $y_i \neq \hat y_i$ gives us the desired untargeted attack. By the existence of adversarial examples \cite{goodfellow2014explaining}, such an attack is possible for any choice of $y_i$ by the same construction as above. And therefore, it honors the same bounds.
\end{proof}

\setcounter{proposition}{0}
\begin{proposition}[Hardness of backdoor detection - I]\label{thm:sat_to_backdoor}
Let $f:\mathbb{R}^n\to \mathbb{R}$ be a ReLU and $g:\mathbb{R}^n\to \mathbb{R}$ be a function. Then \textsc{3-Sat} can be reduced to the decision problem of whether $f$ is equal to $g$ on $[0,1]^n$. Hence checking if $f\equiv g$ on $[0,1]^n$ is NP-hard.
\end{proposition}

\begin{proof}
The proof strategy is constructing a ReLU network to approximate a Boolean expression. This idea is not novel and for example, has been used in \cite{katz2017reluplex} to prove another ReLU related NP-hardness result. Nonetheless, we provide an independent construction here.

Let us define \textsc{Backdoor} as the following decision problem. Given an instance of \textsc{Backdoor} with functions $f, g$ the answer is \texttt{Yes} if there exists some $x \in [0, 1]^n$ such that $f(x) \neq g(x)$ and \texttt{No} otherwise.
We will reduce \textsc{3-Sat} to \textsc{Backdoor}. 
Towards this end, assume that we are given a \textsc{3-Sat} problem with $m$ clauses and $n$ variables. 
Note that $n\leq 3 m$ and $m\leq {2 n\choose 3}$, that is both are within polynomial factors of each other.
Therefore, the input size of the \textsc{3-Sat} is $\text{poly}(m)$. 
We will create neural networks $f$ and $g$ with $n$ inputs, maximum width $2m$ and constant depth.
The weight matrices will have dimensions at most $\max\{2m,n\}\times \max\{2m,n\}=\text{poly}(m)\times \text{poly}(m)$ and similarly the bias vectors will have dimensions at most $\text{poly}(m)$.
Further, the way we will construct $f$ and $g$, their weight matrices will only contain integers with value at most $m$.
This means that each integer can be represented in $O(\log (m))$ bits.
Describing these neural networks can thus be done with $\text{poly}(m)$ parameters. 
Thus, the input size of \textsc{Backdoor} is also $\text{poly}(m)$. 
For now, assume that $f$ and $g$ are created (in poly$(m)$ time) such that $f\not\equiv g$ on $[0,1]^n$ if and only if the \textsc{3-Sat} is satisfiable. 
Then, we have shown that if an algorithm can solve \textsc{Backdoor} in $\text{poly}(m)$ time, then \textsc{Sat} can also be solved in $\text{poly}(m)$ time; or in other words we have reduced \textsc{3-Sat} to \textsc{Backdoor}.
Thus, all that remains to do is to construct in poly$(m)$ time, $f$ and $g$ such that $f\not\equiv g$ on $[0,1]^n$ if and only if the \textsc{3-Sat} is satisfiable.

We will describe how to create the ReLU for $f$. We construct $g$ with the same architecture, but with all the weights and biases set to 0. Thus the question of $f\equiv g$ on $[0,1]^n$ becomes $f\equiv 0$ on $[0,1]^n$. Further $f$ would be such that $f\not \equiv 0$ if and only if the \textsc{3-Sat} is solvable. Essentially, the construction will try to create a ReLU approximation of the \textsc{3-Sat} problem.

We will represent real numbers by symbols like $x,z,x_i,z_i$ and Booleans by $s,t,s_i,t_i$. The real vector $[x_1,\dots, x_n]^\top$ will be denoted as $\bm{x}$. Similarly we will represent Boolean vector $[s_1,\dots, s_n]^\top$ as $\bm{s}$.

Let the \textsc{3-Sat} problem be $h:\mathbb{B}^d\to \mathbb{B}$, where
\begin{equation*}
    h(\bm{s})=\bigwedge_{i=1}^m \left(\bigvee_{j=1}^3 t_{i,j}\right),\label{func:bool}
\end{equation*}
such that $t_{i,j}$ is either $s_k$ or $\neg s_k$ for some $k\in [d]$.

Now we start the construction of $f:\mathbb{R}^d\to \mathbb{R}$. Let $\widehat{x}_i=\sigma (2x_i-1)$ and $\overline{x}_i=\sigma (1-2x_i)$ for all $i\in [d]$ where $\sigma(\cdot)$ represents the ReLU function. These can be computed with 1 layer of ReLU with width $2n$. Roughly speaking if we think of \texttt{True} as being equal to the real number 1 and \texttt{False} as equal to the real number 0, then we want $\widehat{x}_i$ to approximate $s_i$ and $\overline{x}_i$ to approximate $\neg s_i$.

Next for all $i\in[m]$, $j\in[3]$, define 
\begin{equation*}
  z_{i,j} =\begin{cases}
  \widehat{x}_k & \text{if }t_{i,j}=s_k\\
  \overline{x}_k & \text{if }t_{i,j}=\neg s_k
  \end{cases}.
\end{equation*}
Then,
\begin{align}
    f(\bm{x})&= \sigma \left (\left(\sum_{i=1}^m f_i(x)\right)-m+1\right)\label{func:relu}\\
    \text{where, }    f_i(\bm{x})&= \sigma \left(\sum_{j=1}^3 z_{i,j} \right )-\sigma \left(\left(\sum_{j=1}^3 z_{i,j}\right)-1  \right ).\nonumber
\end{align}
Again, roughly speaking we want $f_i(\bm{x})$ to approximate $\bigvee_{j=1}^3 t_{i,j}$ and $f(\bm{x})$ to approximate $h(\bm{s})$. The decomposition of $f$ above is written just for ease of understanding, but in its following form, we can see that it can be computed in 2 layers and width $2m$, using $\widehat{x}_i$ and $\overline{x}_i$
\begin{equation*}
    f(\bm{x})= \sigma \left (\left(\sum_{i=1}^m \sigma \left(\sum_{j=1}^3 z_{i,j} \right )-\sum_{i=1}^m\sigma \left(\left(\sum_{j=1}^3 z_{i,j}\right)-1  \right )\right)-m+1\right).
\end{equation*}

Thus the construction of $f$ from $h$ is complete and we can see that this can be done in polynomial time. Now we need to prove the correctness of the reduction.

We first show that \textsc{3-Sat} $\implies$ \textsc{Backdoor}. This is the simpler of the two directions. Let $\bm{s}$ be an input such that $h(\bm{s})$ is \textsc{True}. Then create $\bm{x}$ as $x_i=1$ if $s_i=$\textsc{True} and $x_i=0$ otherwise. Putting this in Eq. \eqref{func:relu} gives $f(\bm{x})=1$ and thus $f\not \equiv g$ on $[0,1]^n$.

Now, we show \textsc{Backdoor} $\implies$ \textsc{3-Sat}. Let there be an $\bm{x}$ such that $f\not \equiv g$, which is the same as $f(\bm{x})>0$. 
First, assume that $x_k\geq 0.5$. Then, we see that increasing $x_k$ increases $\widehat{x}_k$. This would increase all the $z_{i,j}$ which are defined as $\widehat{x}_k$. Further, $x_k\geq 0.5$ implies that $\overline{x}_k=0$ and thus increasing $x_k$ does not further decrease $\overline{x}_k$. Thus, all the $z_{i,j}$ which are defined as $\overline{x}_k$ do not decrease. Note that $f$ is a non-decreasing function of $z_{i,j}$. This means that we can simply set $x_k$ to 1 and the the value of $f(\bm{x})$ will not decrease. 

Similarly, for the case that $x_k < 0.5$, we can set $x_k$ to 0 and the value of $f(\bm{x})$ will not decrease.

This way, we can find a vector $\bm{x}$ which has only integer entries: 0 or 1 and $f(\bm{x})>0$. Because $f$ consists of only integer operations, this means that $f(\bm{x})\geq1$. Looking at Eq. \eqref{func:relu} and noting that $0\leq f_i(\bm{x}) \leq 1$, we see that this is only possible if $f_i(\bm{x})=1$ for all $i$. Set $s_i=$\textsc{True} if $x_i=1$ and $s_i=$\textsc{False} if $x_i=0$. Then $f_i(x)=1$ implies that $\bigvee_{j=1}^3 t_{i,j}=$\textsc{True} for all $i$. Thus, $h(\bm{s})$ is $\textsc{True}$. 
\end{proof}

\begin{proposition}[Hardness of backdoor detection - II]\label{thm:grad_hard}
Let $f:\mathbb{R}^n\to \mathbb{R}$ be a ReLU and $g:\mathbb{R}^n\to \mathbb{R}$ be a function. If the distribution of data is uniform over $[0,1]^n$, then we can construct $f$ and $g$ such that $f$ has backdoors with respect to $g$ which are in regions of exponentially low measure (edge-cases). Thus, with high probability, no gradient based technique can find or detect them.
\end{proposition}

\begin{proof}
For ease of exposition, we create $f$ and $g$ with a single neuron. However, the construction easily extends to single layer NNs of arbitrary width. Also, assume we are in the high-dimensional regime, so that $n$ is large. (Note that $n$ here refers to the input dimension, not the number of samples in our dataset.)

Define the two networks and the backdoor as follows:
\begin{align*}
f(\mathbf{x}) &= \max (\mathbf{w}^{\top}_{1} \mathbf{x} - b_1,\; 0)\\
g(\mathbf{x}) &= \max (\mathbf{w}^{\top}_{2} \mathbf{x} - b_2,\; 0)\\
\mathcal{B} &= \{\mathbf{x} \in [0, 1]^n: \mathbf{w}^{\top}_{1}\mathbf{x} \geq b_2\}\\
\end{align*}

where $\mathbf{w}_1 = (\frac{1}{n}, \frac{1}{n} \dots, \frac{1}{n})^\top, b_1 = 1$ and $\mathbf{w}_2 = (\frac{1}{n}, \frac{1}{n}, \dots, \frac{1}{n})^\top, b_2 = \frac{1}{2}$

Note that equivalently, $\mathcal{B} = \left\{\mathbf{x} \in [0, 1]^n: \mathbf{1}^\top\mathbf{x} \geq \frac{n}{2}\right\}$ and its measure of is given by:
\begin{align*}
    P(\mathcal{B}) &= P\left(\sum_{i=1}^n x_i \geq \frac{n}{2}\right)\\
    &= P\left(\frac{1}{n}\sum_{i=1}^n x_i \geq \frac{1}{2}\right)\\
    &\leq e^{-n/2}
\end{align*}

where the last step follows from Hoeffding's inequality. Since $n$ is large, we have that $\mathcal{B}$ has exponentially small measure.

We now compare the two networks on $[0, 1]^n \setminus \mathcal{B}$. Clearly $\mathbf{w}^{\top}_{1}\mathbf{x} - b_1 < \mathbf{w}^{\top}_{2}\mathbf{x} - b_2 < 0$.

Therefore,
$$f(\mathbf{x}) = g(\mathbf{x}) = 0 \quad \forall\; \mathbf{x} \in \mathbf{X} \setminus \mathcal{B}$$
and
$$\nabla f(\mathbf{x}) = \nabla g(\mathbf{x}) = 0 \quad \forall\; \mathbf{x} \in \mathbf{X} \setminus \mathcal{B}.$$

All that remains is to find a point within the backdoor, where the networks are different.

Consider $\mathbf{x}_{B} = (1, 1, \dots, 1)^n$. $\mathbf{w}_1^\top \mathbf{x}_{B} - b_1 = 0$. However, $\mathbf{w}_2^\top\mathbf{x}_{B} - b_2 = 1 - \frac{1}{2} = \frac{1}{2}$. Therefore,
$$||f(\mathbf{x}_B) - g(\mathbf{x}_B)|| = \frac{1}{2}$$


Clearly, $f$ and $g$ are identical in terms of zeroth and first order information on the entire region $[0, 1]^n$ except for $\mathcal{B}$. Therefore any gradient based approach to find the backdoor region $\mathcal{B}$ would fail unless we initialize inside the backdoor region, which we have shown to be of exponentially small measure.
\end{proof}

\end{document}